\documentclass[11pt,twoside]{article} 

\usepackage{eqnarray,amsmath}
\usepackage[utf8]{inputenc} 
\usepackage[T1]{fontenc}    
\usepackage{booktabs}       
\usepackage{amsfonts}       
\usepackage{nicefrac}       
\usepackage{microtype}      

\usepackage{epsf}
\usepackage{epsfig}
\usepackage{fancyhdr}
\usepackage{graphics}
\usepackage{graphicx}
\usepackage{psfrag}
\usepackage{fullpage}
\usepackage{pdfpages}

\usepackage{url}
\usepackage[colorlinks,linkcolor=magenta,citecolor=blue, pagebackref=true]{hyperref}
\renewcommand*{\backrefalt}[4]{%
    \ifcase #1 \footnotesize{(Not cited.)}%
    \or        \footnotesize{(Cited on page~#2.)}%
    \else      \footnotesize{(Cited on pages~#2.)}%
    \fi}

\usepackage{color}

\usepackage{amsthm}
\usepackage{amsmath}
\usepackage{amssymb,bbm}
\usepackage{caption}
\usepackage{algorithmic}
\usepackage{algorithm}
\usepackage{textcomp}
\usepackage{siunitx}
\usepackage{wrapfig}
\usepackage{algorithmic}
\usepackage{algorithm}
\usepackage{multirow}
\usepackage{multicol}

\newtheorem{assumption}{Assumption}

\newtheorem{lemma}{Lemma}

\newtheorem{theorem}{Theorem}
\newtheorem{proposition}{Proposition}
\newtheorem{definition}{Definition}

\usepackage{eqnarray,amsmath}
\usepackage{ragged2e}

\usepackage{epsf}
\usepackage{epsfig}
\usepackage{fancyhdr}
\usepackage{graphics}
\usepackage{graphicx}
\usepackage{psfrag}
\usepackage{fullpage}
\usepackage{pdfpages}
\usepackage{subcaption}

\usepackage{color}

\usepackage{amsthm}
\usepackage{amsmath}
\usepackage{amssymb,bbm}
\usepackage{caption}
\usepackage{algorithmic}
\usepackage{algorithm}
\usepackage{textcomp}
\usepackage{siunitx}
\usepackage{wrapfig}
\usepackage{algorithmic}
\usepackage{algorithm}
 \usepackage{multirow}

\newcommand{\bbE}{\mathbb{E}}

\newcommand{\softmax}{\mathrm{Softmax}}

\newcommand{\dint}{\mathrm{d}}

\newcommand{\dboijn}{\Delta\omega_{ij}^{n}}
\newcommand{\dbojn}{\Delta \omega_{j_1}^{n}}

\newcommand{\dtn}{\Delta \tau^n}

\newcommand{\deijn}{\Delta \eta_{ij}^{n}}
\newcommand{\dkijn}{\Delta \kappa_{i_2j_2|j_1}^{n}}

\newcommand{\boin}{\omega_{i}^n}
\newcommand{\bzin}{\beta_{i}^n}
\newcommand{\ain}{a_i^n}
\newcommand{\bin}{b_i^n}
\newcommand{\ein}{\eta_i^n}
\newcommand{\vin}{\nu^n_{i_2|j_1}}
\newcommand{\kin}{\kappa^n_{i_2|j_1}}
\newcommand{\hein}{\eta^n_{j_1i_2}}
\newcommand{\tn}{\tau^n}

\newcommand{\bojn}{\omega_{j}^n}
\newcommand{\bzjn}{\beta_{j}^n}

\newcommand{\hbojn}{\omega_{j_1}^n}
\newcommand{\hbzjn}{\beta_{j_1}^n}

\newcommand{\boj}{\omega_{j}^*}
\newcommand{\bzj}{\beta_{j}^*}
\newcommand{\aj}{a_j^*}
\newcommand{\bj}{b_j^*}
\newcommand{\ej}{\eta_j^*}
\newcommand{\vj}{\nu^*_{j_2|j_1}}
\newcommand{\kj}{\kappa^*_{j_2|j_1}}
\newcommand{\hej}{\eta^*_{j_1j_2}}
\newcommand{\hboj}{\omega_{j_1}^*}
\newcommand{\hbzj}{\beta_{j_1}^*}
\newcommand{\tj}{\tau^*}

\newcommand{\boi}{\omega_{i}^*}
\newcommand{\bzi}{\beta_{i}^*}

\newcommand{\zerod}{{0}_d}
\newcommand{\zeroq}{{0}_q}

\newcommand{\normf}[1]{\|#1\|_{L^2(\mu)}}

\DeclareMathOperator*{\argmin}{arg\,min}

\begin{document}

\begin{center}

{\bf{\LARGE{Convergence Rates for Softmax Gating Mixture of Experts}}}
  
\vspace*{.2in}
{\large{
\begin{tabular}{ccc}
Huy Nguyen & Nhat Ho$^{\star}$ & Alessandro Rinaldo$^{\star}$
\end{tabular}
}}

\vspace*{.1in}

\begin{tabular}{c}
The University of Texas at Austin$^{\dagger}$
\end{tabular}

\today

\vspace*{.2in}

\begin{abstract}
    Mixture of experts (MoE) has recently emerged as an effective framework to advance the efficiency and scalability of machine learning models by softly dividing complex tasks among multiple specialized sub-models termed experts. Central to the success of MoE is an adaptive softmax gating mechanism which takes responsibility for determining the relevance of each expert to a given input and then dynamically assigning experts their respective weights. Despite its widespread use in practice, a comprehensive study on the effects of the softmax gating on the MoE has been lacking in the literature. To bridge this gap in this paper, we perform a convergence analysis of parameter estimation and expert estimation under the MoE equipped with the standard softmax gating or its variants, including a dense-to-sparse gating and a hierarchical softmax gating, respectively. Furthermore, our theories also provide useful insights into the design of sample-efficient expert structures. In particular, we demonstrate that it requires polynomially many data points to estimate experts satisfying our proposed \emph{strong identifiability} condition, namely a commonly used two-layer feed-forward network. In stark contrast, estimating linear experts, which violate the strong identifiability condition, necessitates exponentially many data points as a result of intrinsic parameter interactions expressed in the language of partial differential equations. All the theoretical results are substantiated with a rigorous guarantee.
\end{abstract}
\end{center}
\let\thefootnote\relax\footnotetext{$\star$ Co-last authors.}

\section{Introduction}
\label{sec:introduction}
\begin{figure*}[!ht]
    \centering
    \begin{subfigure}{.45\textwidth}
        \centering
        \includegraphics[scale = .22]{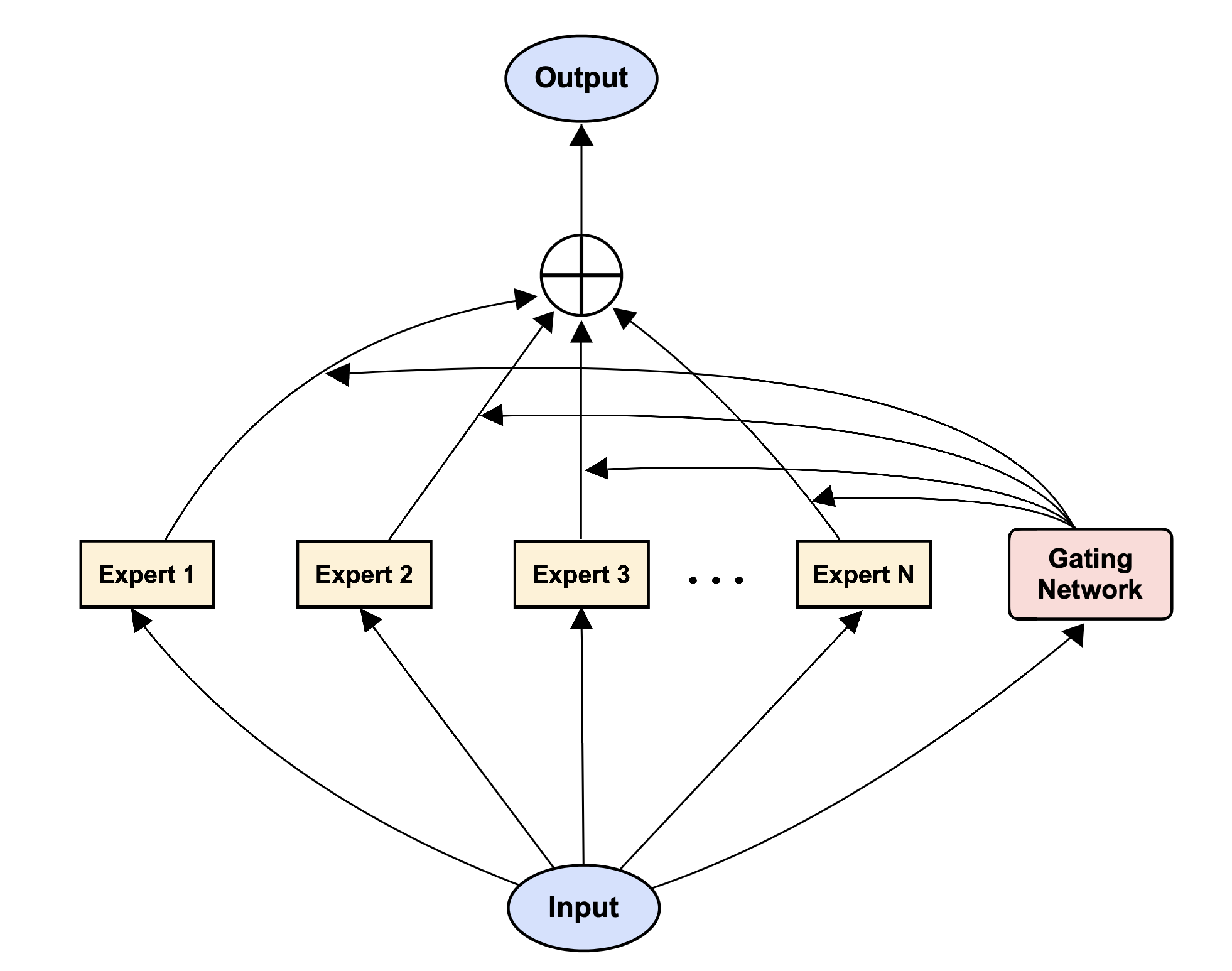}
        \caption{Mixture of Experts.}	
        \label{subfig:moe}
    \end{subfigure}
    \hspace{0.4cm}
    \begin{subfigure}{.45\textwidth}
        \centering
        \includegraphics[scale = .23]{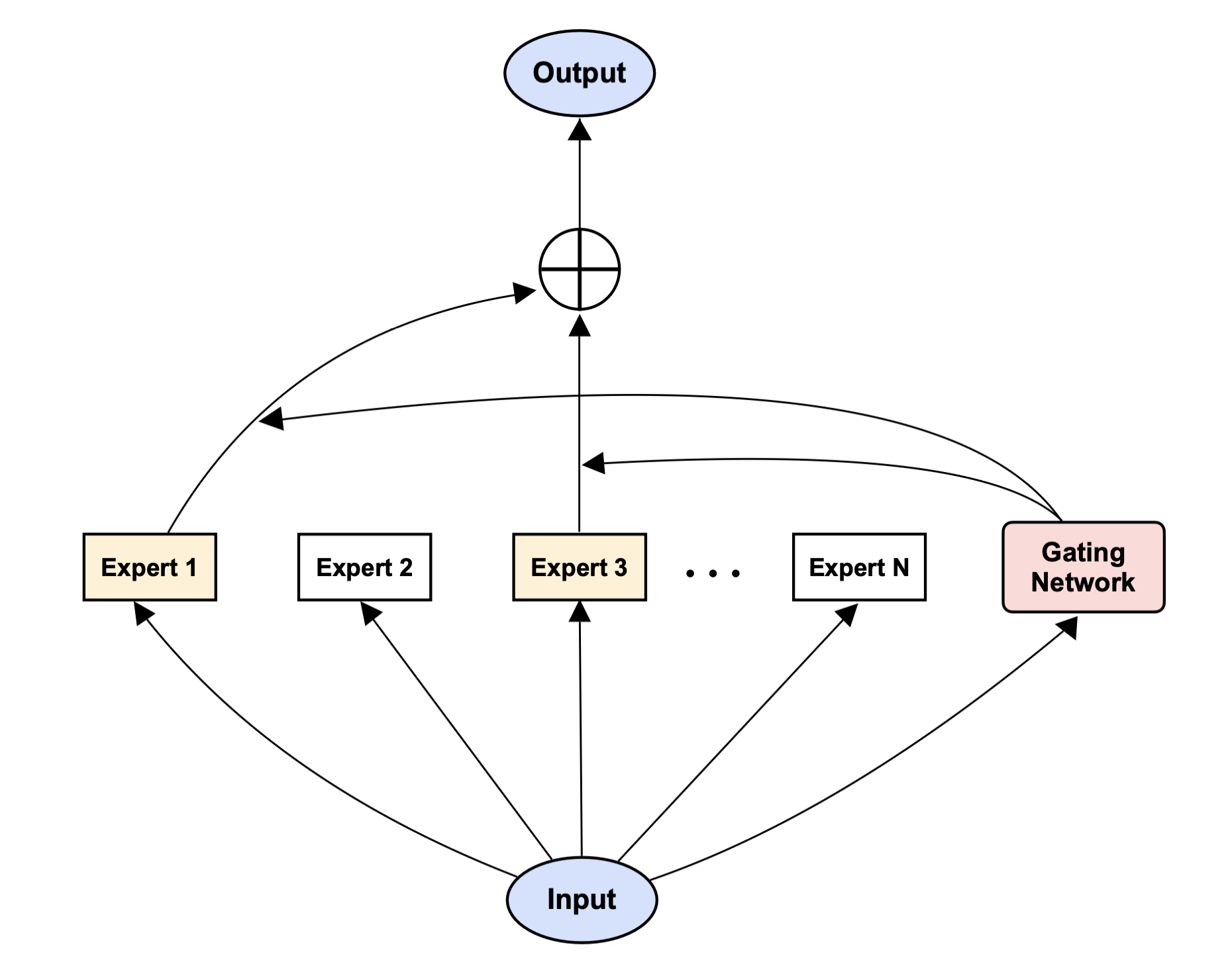}
        \caption{Sparse Mixture of Experts.}	
        \label{subfig:sparse_moe}
    \end{subfigure}%
    \caption{Illustration of mixture of experts.}
\end{figure*}

\noindent
Introduced by Jacob et al. \cite{Jacob_Jordan-1991}, mixture of experts (MoE) has been known as a powerful statistical machine learning framework that generalizes the concept of conventional mixture models \cite{Lindsay-1995} based on the principle of divide and conquer. More specifically, it incorporates the power of multiple sub-models referred to as experts through an adaptive gating mechanism. Here, each expert can be a classifier \cite{chen2022theory,nguyen2024general}, a regression function \cite{faria2010regression} or a feed-forward network (FFN) that specializes in some specific tasks \cite{shazeer2017topk,dai2024deepseekmoe}. Meanwhile, a gating function is formulated as an FFN followed by a softmax normalization, which we refer to as the softmax gating function throughout the paper. Its responsibility is to dynamically assign a corresponding weight for each individual expert in a way that experts which are more relevant to the input will have larger weights than others. Therefore, the weight set of the MoE model varies with the input value rather than remains constant as that of the traditional mixture model (see Figure~\ref{subfig:moe}). This feature accounts for the flexibility and adaptivity of the MoE, leading to a series of MoE applications in classification \cite{waterhouse_classification_1994}, multi-task learning \cite{hazimeh_dselect_k_2021,gupta2022sparsely}, speech recognition \cite{peng1996speech,You_Speech_MoE}, and bioinformatics \cite{Qi2007AMO}, etc. 

\textbf{}

\noindent
In order to enhance the computational efficiency of the MoE, Shazeer et al. \cite{shazeer2017topk} has recently developed a sparse variant of the softmax gating function known as the Top-$K$ sparse softmax gating function. In particular, the gating network is trained to route each input to only the $K$ most relevant expert networks rather than all of them (see Figure~\ref{subfig:sparse_moe}). This mechanism can be viewed as a form of conditional computation \cite{bengio2013conditional,cho2014exponentially}, and it allows the sparse MoE to reduce the computational overhead significantly while maintaining or even improving the model performance. A typical application of the sparse MoE is in the Transformer model \cite{vaswani2017attention}, which has been the state-of-the-art architecture for many natural language processing tasks. More concretely, the FFN in a transformer layer is often replaced with a sparse mixture of smaller FFNs whose total number of parameters matches that of the original FFN layer, which might contain up to billions of parameters \cite{jiang2024mixtral,dai2024deepseekmoe}. Since only a few FFNs are activated per input, it is obvious that the computational cost decreases substantially. Moreover, the usage of multiple FFNs might help learn domain-specific or task-specific tasks better than using a single FFN. For example, when the dataset consists of diverse data modalities such as texts, images, time series, etc., then each FFN can be trained to specialize in processing a specific data modality to make the most out of the dataset \cite{han2024fusemoe,yun2024flexmoe}, thereby enhancing the model performance. As a result, the sparse MoE has been widely utilized in several large-scale models, namely large language models \cite{liu2024deepseek,jiang2024mixtral,grattafiori2024llama3,geminiteam2024gemini15}, computer vision \cite{Riquelme2021scalingvision,liang_m3vit_2022,chen2023adamv}, domain generalization \cite{li2023sparse,nguyen2024cosine}, and reinforcement learning \cite{ceron2024rl,chow_mixture_expert_2023}. \\

\noindent
Despite its widespread use, there are two main disadvantages of
the sparse gating function proposed in \cite{shazeer2017topk}. First, fixing the number of activated experts per input not only hinders the expert exploration but also makes the gating function discontinuous, probably causing some challenges in terms of optimization. Second, since the gating network lacks the experience of expert selection in early training, it may also exhibit undesirable instability. For those reasons, Nie et al. \cite{nie2022evomoe} propose a dense-to-sparse gating function which involves a temperature parameter in the dense softmax gating to control the weight distribution. In particular, when the temperature parameter tends to infinity, the mixture weights are uniformly distributed, meaning that all the experts are activated and assigned identical weights. On the other hand, when the temperature parameter approaches zero, the weight distribution converges to a one-hot vector, indicating that only one expert is activated per input. This strategy helps smooth the expert selection process as well as dynamically adjust the sparsity level of the MoE model. Therefore, the dense-to-sparse gating has been leveraged in several works on MoE, namely \cite{chi_representation_2022,nguyen2024temperature,csordas2023approximating}. \\

\begin{figure*}[!ht]
    \centering
    \includegraphics[scale =.3]{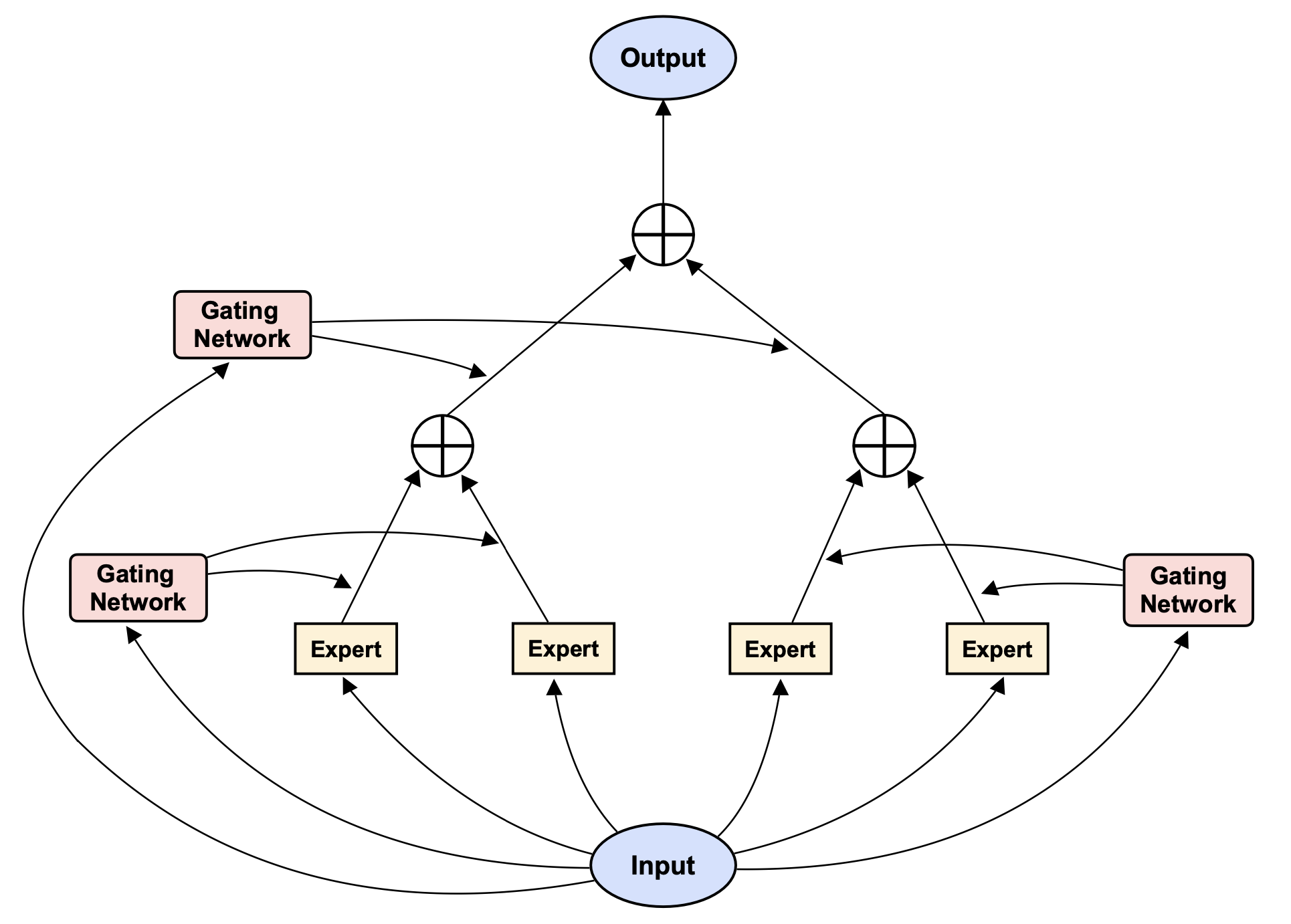}
    \caption{Illustration of two-level hierarchical mixture of experts.}	
    \label{fig:hmoe}
\end{figure*}

\noindent
In addition to the standard MoE and its sparse versions that we have discussed so far, we would also like to bring attention to a hierarchical MoE (HMoE) model introduced by Jordan et al. \cite{Jordan-1994} as a tree-structured architecture for supervised learning. The HMoE encompasses multiple layers of gating networks to organize the experts in a multi-level hierarchy as illustrated in Figure~\ref{fig:hmoe}. Such hierarchical design has been demonstrated to facilitate the expert specialization and enhance the model generalization, particularly in the scenarios involving diverse data distributions \cite{nguyen2024hmoe,irsoy2021dropout} or complex decision-making tasks \cite{jeremiah2013specifying,moges2016hierarchical}.\\ 

\noindent
\textbf{Related works.} Given the success of applying the MoE and HMoE models in practice, it is natural to ask about the theoretical attempts to understand those models. First, Zeevi et al. \cite{zeevi1998approximation} established the error of approximating a target function in the Sobolev class using a mixture of generalized linear experts under the $L^p$ norm. After that, Jiang et al. \cite{jiang1999hmoe} proceeded to studied the approximation power of the HMoE models where exponential family regression models with generalized linear mean functions were aggregated. They demonstrated that such models were able to approximate one-parameter exponential family densities with arbitrary smooth mean functions in a transformed Sobolev space when the number of experts increased. Next, Mendes et al. \cite{mendes2011convergence} investigated the convergence rate of the maximum likelihood estimator (MLE) under the MoE model where each expert was formulated as a polynomial regression model. From the theoretical results, they provided some implications for the optimal number of experts and the complexity of the expert models. The convergence analysis of the MLE was then continued in \cite{ho2022gaussian} and \cite{nguyen2023demystifying} but under the Gaussian MoE models. Those works pointed out that the parameter estimation rates were negatively affected by some intrinsic interaction among the gating and expert parameters expressed in the language of partial differential equations (PDEs). Unlike previous works where the MoE was associated with a probability distribution, Nguyen et al. \cite{nguyen2024squares} considered a regression framework where the regression function took the form of a deterministic softmax gating MoE, and studied the problem of expert estimation using the least square method. In particular, they derived a so-called strong identifiability condition to characterize the types of experts that admitted faster convergence rates than others, namely those formulated as two-layer FFNs with GELU or sigmoid activation function. In this work, we will first revisit those results and then extend them to the settings where the regression function admits the form of a dense-to-sparse gating MoE and a hierarchical MoE, respectively, which have remained elusive in the literature. For the sake of presentation, let us introduce the formal problem statement for the setting of the softmax gating MoE below and then defer those for their variants to Section~\ref{sec:dense-to-sparse_MoE} and Section~\ref{sec:hierarchical_MoE}. \\


\noindent
\textbf{Problem setting.} We assume that an i.i.d. sample of size $n$: $(X_{1}, Y_{1}), (X_{2}, Y_{2}), \ldots, (X_{n}, Y_{n})$ in $\mathcal{X}\times \mathcal{Y}\subseteq\mathbb{R}^d\times\mathbb{R}$ is generated according to the model 
\begin{align}
    Y_{i} = f_{G_{*}}(X_{i}) + \varepsilon_{i}, \quad i=1,2,\ldots,n. \label{eq:moe_regression_model}
\end{align}
Above, we assume that $X_{1}, X_{2}, \ldots, X_{n}$ are i.i.d. samples from some probability distribution $\mu$. Meanwhile, $\varepsilon_{1}, \varepsilon_{2}, \ldots, \varepsilon_{n}$ are independent noise random variables such that their conditional probability distributions given the input are Gaussian, that is, $\varepsilon_{i}|X_i\sim\mathcal{N}(0,\nu)$, for all $1 \leq i \leq n$. We note in passing that the Gaussian assumption is simply for the ease of proof arguments. Next, the regression function $f_{G_{*}}(\cdot)$ is assumed to take the form of the softmax gating mixture of $k_*$ experts, namely
\begin{align}
    \label{eq:softmax_MoE}
    f_{G_{*}}(x) := \sum_{i=1}^{k_*} \softmax((\omega^*_{i})^{\top}x+\beta^*_{i})\cdot \mathcal{E}(x,\eta^*_{i}),
\end{align}
where $(\beta^*_{i},\omega^*_{i},\eta^*_{i})_{i=1}^{k_*}$ are true yet unknown parameters belonging to the parameter space $\Theta\subseteq\mathbb{R}\times\mathbb{R}^d\times\mathbb{R}^q$ and $G_{*} := \sum_{i = 1}^{k_{*}} \exp(\beta^{*}_{i}) \delta_{(\omega^{*}_{i},\eta^*_{i})}$ denotes the associated \emph{mixing measure}, that is, a weighted sum of Dirac delta measures. The corresponding regression functions to the settings of the dense-to-sparse gating MoE and the HMoE are given in equations~\eqref{eq:dense-to-sparse_MoE} and \eqref{eq:hierarchical_MoE}, respectively. Additionally, the terms $\mathcal{E}(x,\eta^*_{i})$, for $1\leq i\leq k_*$, are referred to as experts.\\

\noindent
It is worth noting that expert specialization is an essential problem in the literature of MoE models \cite{dai2024deepseekmoe,oldfield2024specialize}. Therefore, our main objective is to learn how fast we can estimate the parametric ground-truth experts $\mathcal{E}(x,\eta^*_{i})$, which can be obtained by determining the convergence rate of parameter estimation. Since the number of experts $k_*$ is unknown in practice, we over-specify the model in equation~\eqref{eq:softmax_MoE} by a mixture of $k$ experts, where $k>k_*$ is a given threshold. Then, we can estimate the unknown parameters $(\beta^*_{i},\omega^*_{i},\eta^*_{i})_{i=1}^{k_*}$ via estimating the mixing measure $G_*$ using the least square method as follows:
\begin{align}
    \label{eq:least_square_estimator}
    \widehat{G}_n:=\argmin_{G\in\mathcal{G}_{k}(\Theta)}\sum_{i=1}^{n}\Big(Y_i-f_{G}(X_i)\Big)^2,
\end{align}
where $\mathcal{G}_{k}(\Theta):=\{G=\sum_{i=1}^{k'}\exp(\beta_{i})\delta_{(\omega_{i},\eta_{i})}:1\leq k'\leq k, \  (\beta_{i},\omega_{i},\eta_{i})\in\Theta\}$ stands for the set of all mixing measures with no more than $k$ atoms.\\

\noindent
\textbf{Contributions.} In this paper, we analyze the convergence behavior of parameter estimation and expert estimation under the (hierarchical) MoE with the softmax gating and the dense-to-sparse gating. Our ultimate goal is to find the optimal expert structure for each MoE model to provide a useful guide on the model design for its practical applications. Our contributions are three-fold and can be summarized as follows (see also Table~\ref{table:expert_rates}):\\

\noindent
\emph{1. Softmax gating MoE:} We re-state the strong identifiability condition in \cite{nguyen2024squares} for characterizing the structure of experts admitting faster convergence rates than others. The main intuition behind that condition is to eliminate interactions among parameters through some PDEs. Then, we demonstrate that the rates for estimating strongly identifiable experts, including those formulated as FFNs with GELU or sigmoid activation, are parametric on the sample size. On the other hand, linear experts, which fail to satisfy the strong identifiability condition, are shown to have slower estimation rates than any polynomial rates.\\

\noindent
\emph{2. Dense-to-sparse gating MoE:} Recall that we involve the temperature parameter in the dense-to-sparse gating function to control the sparsity of the MoE model. However, our theory reveals that the temperature has an undesirable interaction with the gating parameters expressed via a PDE, leading to a substantial deceleration in the expert convergence regardless of the expert structure. In response to this issue, we generalize the linear router inside the dense-to-sparse gating function to a general router and then establish a condition to survey which combinations of the router and the expert structure will accelerate the expert convergence.\\

\noindent
\emph{3. Hierarchical MoE:} For the softmax gating hierarchical MoE, we discover that experts satisfying the aforementioned strong identifiability condition still enjoy a faster estimation rate than others. At the same time, we show that the rates for estimating linear experts are slower than any polynomial rates due to the interaction between gating parameters and expert parameters.\\

\begin{table*}[t!]
\caption{Summary of estimation rates for strongly identifiable experts and linear experts.}
\centering
\scalebox{0.75}{
\begin{tabular}{ | c | c |c|c|} 
\hline
{\textbf{}} & {\bf Strongly Identifiable Experts} & {\bf Linear Experts}& {\bf Theorems}  \\
\hline 
Softmax gating MoE &$\mathcal{O}_P([\log(n)/n]^{1/4})$ &$\mathcal{O}_P(1/\log^{\lambda}(n))$  &  Thm.\ref{theorem:general_experts}, Thm.\ref{theorem:linear_experts}\\
\hline
Dense-to-sparse gating MoE with linear router&$\mathcal{O}_P(1/\log^{\lambda}(n))$ &$\mathcal{O}_P(1/\log^{\lambda}(n))$ & Thm.\ref{theorem:linear_dense-to-sparse_experts}\\
\hline
Dense-to-sparse gating MoE with general router&$\mathcal{O}_P([\log(n)/n]^{1/4})$ &$\mathcal{O}_P(1/\log^{\lambda}(n))$ & Thm.\ref{theorem:general_dense-to-sparse_experts}\\
\hline
Hierarchical MoE&$\mathcal{O}_P([\log(n)/n]^{1/4})$ &$\mathcal{O}_P(1/\log^{\lambda}(n))$  & Thm.\ref{theorem:hierarchical_general_experts}, Thm.\ref{theorem:hierarchical_linear_experts}\\
\hline
\end{tabular}}
\label{table:expert_rates}
\end{table*}

\noindent
\textbf{Organization.} The paper proceeds as follows. In Section~\ref{sec:softmax_MoE}, we recall the convergence analysis of the softmax gating MoE equipped with strongly identifiable experts and linear experts. Next, we study the effects of involving the temperature parameter in the softmax gating function on the convergence of expert estimation in Section~\ref{sec:dense-to-sparse_MoE}. Subsequently, we generalize the results to the setting of the HMoE model in Section~\ref{sec:hierarchical_MoE} before providing an in-depth discussion on the theoretical results in Section~\ref{sec:conclusion}. Full proofs and additional results can be found in the Appendices.\\

\noindent
\textbf{Notations.} For any $n\in\mathbb{N}$, we denote $[n]$ as the set $=\{1,2,\ldots,n\}$. Additionally, for any set $S$, we refer to $|S|$ as its cardinality. Next, for any vectors $v:=(v_1,v_2,\ldots,v_d) \in \mathbb{R}^{d}$ and $\alpha:=(\alpha_1,\alpha_2,\ldots,\alpha_d)\in\mathbb{N}^d$, we let $v^{\alpha}=v_{1}^{\alpha_{1}}v_{2}^{\alpha_{2}}\ldots v_{d}^{\alpha_{d}}$, $|v|:=v_1+v_2+\ldots+v_d$ and $\alpha!:=\alpha_{1}!\alpha_{2}!\ldots \alpha_{d}!$, while $\|v\|$ stands for its $2$-norm value. Lastly, for any two positive sequences $(a_n)_{n\geq 1}$ and $(b_n)_{n\geq 1}$, we write $a_n = \mathcal{O}(b_n)$ or $a_{n} \lesssim b_{n}$ if $a_n \leq C b_n$ for all $ n\in\mathbb{N}$, where $C > 0$ is some universal constant. The notation $a_{n} = \mathcal{O}_{P}(b_{n})$ indicates that $a_{n}/b_{n}$ is stochastically bounded.

\section{Softmax Gating MoE}
\label{sec:softmax_MoE}
\noindent
In this section, we first establish the convergence rate of estimation for the regression function under some mild assumptions. This lays the foundation for deriving a condition to capture the optimal expert structure in terms of sample efficiency in Section~\ref{sec:general_experts}. Next, we determine the convergence behavior of estimation for linear experts, which fail to satisfy that condition, in Section~\ref{sec:linear_experts}.\\

\noindent
\textbf{Assumptions.} To begin with, let us introduce the essential yet mild assumptions for our subsequent convergence analysis. \\

\noindent
\emph{(A.1)} The parameter space $\Theta$ is a compact subset of $\mathbb{R}\times\mathbb{R}^d\times\mathbb{R}^q$, and its dimension is fixed unless stated otherwise. Meanwhile, the input space $\mathcal{X}$ is a bounded subset of $\mathbb{R}$. \\

\noindent
\emph{(A.2)} 
The ground-truth expert parameters $\eta^*_{1},\eta^*_{2},\ldots,\eta^*_{k_*}$ are distinct. Furthermore, the expert functions $\mathcal{E}(x,\eta)$ is bounded and Lipschitz continuous with respect to $\eta$ for almost every $x$.\\

\noindent
\emph{(A.3)} 
The last pair of gating parameters are zero, that is, $\beta^*_{k_*}=0$ and $\omega^*_{k_*}=\zerod$.\\

\noindent
\emph{(A.4)} 
At least one of the ground-truth gating parameters $\omega^*_{1},\omega^*_{2},\ldots,\omega^*_{k_*}$ is different from zero.\\

\noindent
Above, the first assumption (A.1) is to ensure the convergence of least square estimators, whereas the second (A.2) is necessary for the distinction among experts. Next, the third assumption (A.3) helps prevent the invariance to translation of the softmax gating function, which negatively affects the identifiability of the softmax gating MoE model. The last assumption (A.4) is to guarantee that the gating value is input-dependent as in practice.\\

\noindent
Given the above assumptions, we are ready to study the convergence behavior of regression estimation in the following proposition.

\begin{proposition}
    \label{theorem:regression_rate_softmax}
    For a least square estimator $\widehat{G}_n$ in equation~\eqref{eq:least_square_estimator}, the convergence rate of regression function $f_{\widehat{G}_n}$ is given by
    \begin{align}
        \label{eq:model_bound}
        \normf{f_{\widehat{G}_n}-f_{G_*}}=\mathcal{O}_{P}([\log(n)/n]^{\frac{1}{2}}).
    \end{align}
\end{proposition}
\noindent
The proof of Proposition~\ref{theorem:regression_rate_softmax} can be found in Appendix~\ref{appendix:regression_rate_softmax}. The bound in equation~\eqref{eq:model_bound} indicates that the least square estimator $f_{\widehat{G}_n}$ converges to its true regression function $f_{G_*}$ under the $L^2(\mu)$ norm at the rate of order $\mathcal{O}_{P}([\log(n)/n]^{\frac{1}{2}})$, which is parametric on the sample size $n$. As a consequence, in order to capture the convergence behavior of parameter estimation and expert estimation, it is sufficient to build a loss function among parameters $\mathcal{L}(\widehat{G}_n,G_*)$ such that $\normf{f_{\widehat{G}_n}-f_{G_*}}\gtrsim\mathcal{L}(\widehat{G}_n,G_*)$, which implies that $\mathcal{L}(\widehat{G}_n,G_*)=\mathcal{O}_{P}([\log(n)/n]^{\frac{1}{2}})$. Given these results, we attempt to construct a condition to characterize \emph{strongly identifiable} expert functions that require fewer data points to estimate than others.



\subsection{Strongly identifiable experts}
\label{sec:general_experts}
\noindent
Before presenting the condition for the strongly identifiable expert structure, let us highlight the challenges of establishing the essential $L^2$-lower bound $\normf{f_{\widehat{G}_n}-f_{G_*}}\gtrsim\mathcal{L}(G,G_*)$ to determine the convergence rates of parameter estimation and expert estimation.\\

\noindent
\textbf{Challenges.} In our proof, a key step to derive the aforementioned $L^2$-lower bound is to decompose the discrepancy between the estimation of the regression function and its true counterpart, that is, $f_{\widehat{G}_n}(x)-f_{G_*}(x)$, using Taylor expansions to the function $x\mapsto F(x;\omega,\eta):=\exp(\omega^{\top}x)\mathcal{E}(x,\eta)$ staying implicit in the representation of regression function. Furthermore, it is necessary to guarantee that the function $F$ and its partial derivatives resulting from the Taylor expansions are linearly independent so that when the regression discrepancy goes to zero as $n\to\infty$, the coefficients of those terms, which encompass the parameter discrepancies, also converge to zero. Therefore, we need to establish a non-trivial algebraic independence condition on the expert function $\mathcal{E}(x,\eta)$ called \emph{strong identifiability} in Definition~\ref{def:softmax_condition} to ensure such a linear independent property. This requires us to adopt new techniques as previous works on the softmax gating MoE \cite{nguyen2023demystifying,nguyen2024general} employ only linear experts.

\begin{definition}[Strong Identifiability]
    \label{def:softmax_condition}
   An expert function $x \mapsto \mathcal{E}(x,\eta)$ is said to be strongly identifiable if it is twice differentiable with respect to its parameter $\eta$ for almost every $x$, and the set of functions in $x$
    \begin{align*}
        &\Big\{x^{\nu}\cdot\frac{\partial^{|\rho|}\mathcal{E}}{\partial\eta^{\rho}}(x,\eta_j): j\in[k], \nu\in\mathbb{N}^d, \ \rho\in\mathbb{N}^{q}, 0\leq |\nu|+|\rho|\leq 2\Big\}
    \end{align*}
    is linearly independent for almost every $x$ for any $k\geq 1$ and distinct parameters $\eta_1,\eta_2\ldots,\eta_k$.
\end{definition}
\noindent
\textbf{Example.} We can justify that several experts employed in practice, namely FFNs with activation functions like $\mathrm{GELU}$ \cite{hendrycks2023gaussian}, $\mathrm{sigmoid}$, $\mathrm{tanh}$, and non-linear transformed input, satisfy the strong identifiability condition. For instance, let us consider a two-layer FFN with normalized input, i.e.
\begin{align*}
    \mathcal{E}(x,(a,b,c))=c\sigma\Big(a\frac{x}{\|x\|}+b\Big),
\end{align*}
where $\sigma$ is one among the activation functions $\mathrm{GELU}$, $\mathrm{sigmoid}$ and $\mathrm{tanh}$, and $x,a\in\mathbb{R}^d$, $b,c,\in\mathbb{R}$. On the other hand, the strong identifiability will be violated if the activation function $\sigma$ is of polynomial form, e.g., $\sigma(z)=z^p$ for all $z\in\mathbb{R}$ for some positive integer $p\in\mathbb{N}$.\\

\noindent
Intuitively, the linear independence
of functions in Definition 3.1 helps eliminate potential interactions among parameters expressed in the language of
partial differential equations (see e.g., equation (10) and
equation (16) where gating parameters $\beta_1$ interact with expert parameters a). Such interactions are demonstrated to
result in significantly slow expert estimation rates (see Theorem 4.4 and Theorem 4.6).\\

\noindent
Now, it is necessary to construct a loss function among parameters to capture the convergence rate of parameter estimation. In previous works, Nguyen et al. \cite{nguyen2016latentmixing} utilized the generalized Wasserstein divergence between mixing measures to determine parameter estimation rates under the setting of classical mixture models. Then, this divergence was adopted again in \cite{ho2022gaussian} to establish the convergence rates of parameter estimation in Gaussian mixture of experts. However, there is an inherent drawback of the generalized Wasserstein. In particular, this divergence implies estimation rates for all individual parameters while these rates should vary with the number of fitted parameters. In response to this issue, we propose using a loss function built based on the concept of Voronoi cells \cite{manole22refined} to accurately characterize parameter estimation rates.

\begin{figure*}[!ht]
    \centering
    \includegraphics[scale =.1]{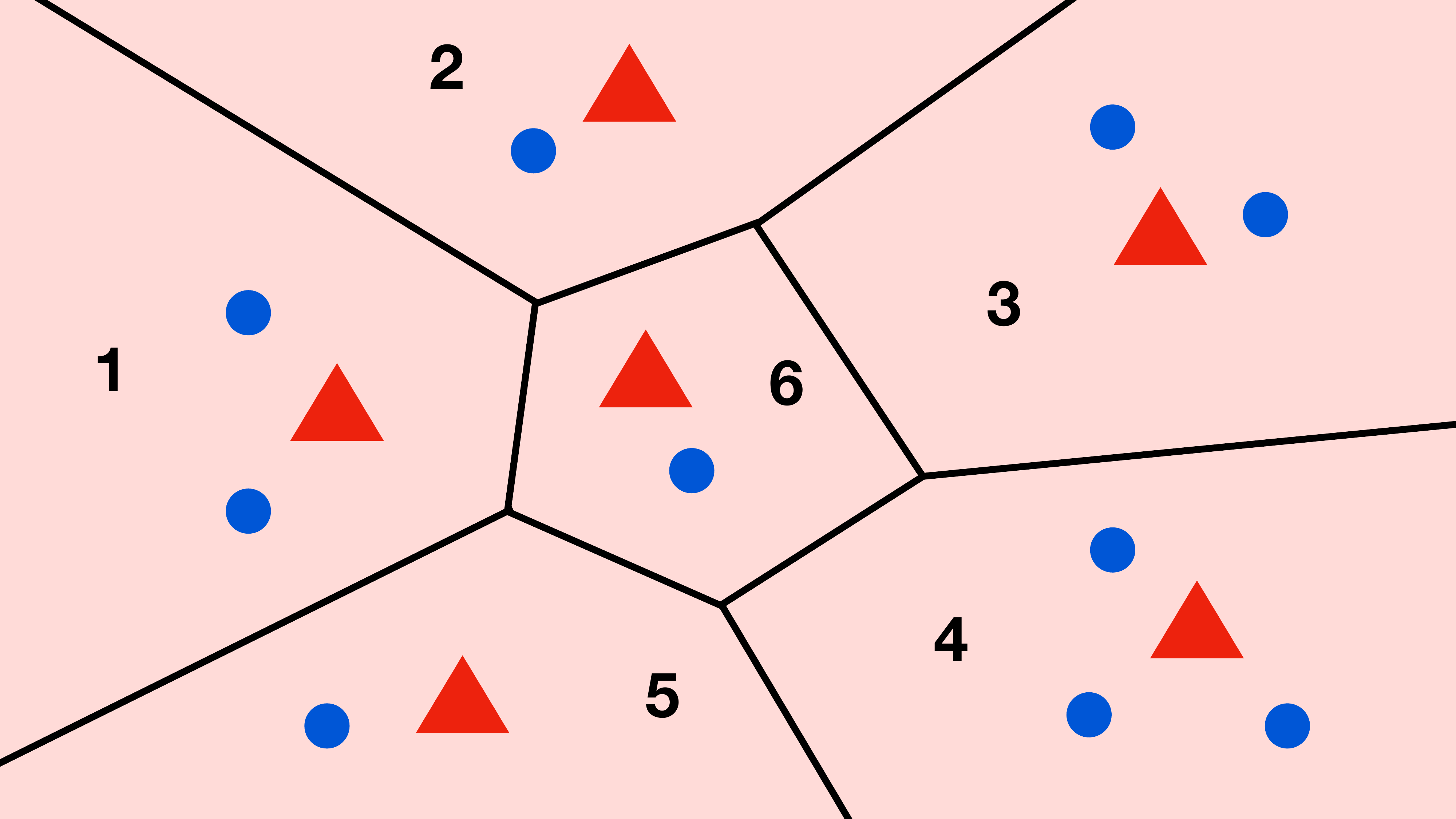}
    \caption{Illustration of Voronoi cells generated by $k_*=6$ atoms of the ground-truth $G_*$ (red triangles) and $k=10$ fitted atoms of the estimator $\widehat{G}_n$ (blue rounds). By definition, each Voronoi cell is generated by one ground-truth atom, and its cardinality equals the number of corresponding fitted atoms. For instance, the red triangle in cell 4 is fitted by three blue rounds, implying that the cardinality of Voronoi cell 4 is three.}	
    \label{fig:voronoi_cells}
\end{figure*}
\noindent
\textbf{Voronoi loss function.} For a mixing measure $G$ with $1\leq k'\leq k$ atoms, we partition its atoms into the set of Voronoi cells $\{\mathcal{A}_j\equiv
    \mathcal{A}_j(G):j\in[k_*] \}$ generated by the atoms of the ground-truth mixing measure $G_*$ defined as
\begin{align}
    \hspace{-0.5em}\mathcal{A}_j:=
    \left\{
    i\in[k']:
    \| \theta_i-\theta_j^* \|
    \leq
    \| \theta_i-\theta_{\ell}^* \|,
    \forall \ell\neq j
    \right\},
\end{align}
with $\theta_i:=(\omega_i,\eta_i)$ and
$\theta^*_j:=(\omega^*_j,\eta^*_j)$ for all $j\in[k_*]$ (see Figure~\ref{fig:voronoi_cells}). Then, the Voronoi loss of interest is given by
\begin{align}
    \mathcal{L}_{1}(G,G_*):=\sum_{j=1}^{k_*}\Big|\sum_{i\in\mathcal{A}_j}\exp(\beta_{i})-\exp(\beta^*_{j})\Big|&+\sum_{j:|\mathcal{A}_j|=1}\sum_{i\in\mathcal{A}_j}\exp(\beta_{i})\Big[\|\Delta\omega_{ij}\|+\|\Delta \eta_{ij}\|\Big]\nonumber\\
    \label{eq:loss_d1}
    &+\sum_{j:|\mathcal{A}_j|>1}\sum_{i\in\mathcal{A}_j}\exp(\beta_{i})\Big[\|\Delta\omega_{ij}\|^2+\|\Delta \eta_{ij}\|^2\Big],
\end{align}
where we denote $\Delta\omega_{ij}:=\omega_{i}-\omega_{j}$ and $\Delta \eta_{ij}:=\eta_i-\eta^*_j$. It is worth noting that the cardinality of a Voronoi cell $\mathcal{A}_j$ is exactly the number of parameters approaching the ground-truth atom $\theta^*_j$ for all $j\in[k_*]$. By convention, if a Voronoi cell $\mathcal{A}_j$ is an empty set, then we set the respective summation to zero. Regarding the Voronoi loss function, we notice that $\mathcal{L}_{1}(G,G_*)=0$ holds if and only if $G\equiv G_*$. This property indicates that when the Voronoi loss $\mathcal{L}_{1}(G,G_*)$ becomes small enough, the parameter discrepancies $\Delta\omega_{ij}$ and $\Delta\eta_{ij}$ are also small accordingly. Therefore, it is a suitable loss function for the sake of capturing parameter estimation rates.
Furthermore, the Voronoi loss $\mathcal{L}_1(G,G_*)$ induces more accurate convergence rates of individual parameter estimation than the generalized Wasserstein divergence. In particular, suppose that there exist a sequence of mixing measures $(G_n)$ such that $\mathcal{L}_1(G_n,G_*)$ converges to zero at a rate $\gamma_n=o(1)$ as $n\to\infty$, we deduce that the convergence rates for estimating \emph{exactly-specified parameters} $\theta^*_j$, whose Voronoi cells $\mathcal{A}_j$ have only one element, are also $\gamma_n$. Meanwhile, \emph{over-specified parameters} $\theta^*_j$, whose Voronoi cells have more than one element, admit the estimation rates of order $\gamma_n^{\frac{1}{2}}$. On the other hand, if we used the generalized Wasserstein divergence, then the parameter estimation rates would be the same irrespective of the Voronoi cell cardinality. However, as the Voronoi loss $\mathcal{L}_1(G,G_*)$ is not symmetric, it is not a proper metric. \\

\noindent
Given the above loss function, we provide the convergence rate of parameter estimation in Theorem~\ref{theorem:general_experts}.

\begin{theorem}
    \label{theorem:general_experts}
    Suppose that the expert function $x\mapsto\mathcal{E}(x,\eta)$ satisfies the strong identifiability condition presented in Definition~\ref{def:softmax_condition}, then the lower bound $\normf{f_{G}-f_{G_*}}\gtrsim\mathcal{L}_{1}(G,G_*)$ holds for all $G\in\mathcal{G}_{k}(\Theta)$, indicating that
    \begin{align}
    \label{eq:general_experts_bound}
        \mathcal{L}_{1}(\widehat{G}_n,G_*)=\mathcal{O}_{P}([\log(n)/n]^{\frac{1}{2}}).
    \end{align} 
\end{theorem}
\noindent
The proof of Theorem~\ref{theorem:general_experts} can be found in Appendix~\ref{appendix:general_experts}. A few remarks regarding the result of this theorem are in order.\\

\noindent
\emph{(i) Parameter estimation rates:} Putting the bound~\eqref{eq:general_experts_bound} and the definition of the Voronoi loss $\mathcal{L}_1$ in equation~\eqref{eq:loss_d1} together, it follows that the rates for estimating exactly-specified parameters $\omega^*_j,\eta^*_j$ are of the order $\mathcal{O}_{P}([\log(n)/n]^{\frac{1}{2}})$, while over-specified parameters $\omega^*_j,\eta^*_j$ admit slower estimation rates of the order $\mathcal{O}_{P}([\log(n)/n]^{\frac{1}{4}})$. This highlights the benefit of the Voronoi loss function over the generalized Wasserstein divergence as the former can distinguish the estimation rates of exactly-specified and over-specified parameters while the latter induces identical convergence rates for all parameter estimations.\\

\noindent
\emph{(ii) Expert estimation rates:} By definition, since $\mathcal{E}(x,\eta)$ is a strongly identifiable expert function, it is twice differentiable over a bounded domain, implying that it is also a Lipschitz function. Assume that the least square estimator $\widehat{G}_n$ is represented as $\widehat{G}_n=\sum_{i=1}^{\widehat{k}_n}\exp(\widehat{\beta}_{i})\delta_{(\widehat{\omega}^n_{i},\widehat{\eta}^n_i)}$, then the following inequality holds for all $i\in\mathcal{A}_j(\widehat{G}_n)$ and $j\in[k_*]$:
\begin{align}
    \sup_{x} |\mathcal{E}(x,\widehat{\eta}^n_i)-\mathcal{E}(x,\eta^*_j)| &\lesssim\|\widehat{\eta}^n_i-\eta^*_j\|.  \label{eq:expert_rate}
\end{align}
From this result, we deduce that the convergence rates of estimating experts $\mathcal{E}(x,\eta^*_j)$ are identical to the rates for estimating parameters $\eta^*_j$, which are of the order $\mathcal{O}_{P}([\log(n)/n]^{\frac{1}{2}})$ when $|\mathcal{A}_j|=1$ and become slower at the order $\mathcal{O}_{P}([\log(n)/n]^{\frac{1}{4}})$ when $|\mathcal{A}_j|>1$. These rates imply that to achieve an approximation of the experts $\mathcal{E}(x,\eta^*_j)$ with a given error $\epsilon>0$, we need polynomially many data points of order $\mathcal{O}(\epsilon^{-2})$ or $\mathcal{O}(\epsilon^{-4})$. To see the effects of the strong identifiability condition on the parameter and expert estimation problem more clearly, we will demonstrate in the next section that it requires a substantially larger sample size to estimate non-strongly identifiable experts. 


\subsection{Linear experts}
\label{sec:linear_experts}
\noindent
Moving to this section, we complement our convergence analysis of parameter and expert estimations under the softmax gating MoE by taking into account a class of linear experts of the form $\mathcal{E}(x,(a,b))=a^{\top}x+b$ for $(a,b)\in\mathbb{R}^d\times\mathbb{R}$, which violate the strong identifiability condition in Definition~\ref{def:softmax_condition}. \\

\noindent
\textbf{Parameter interaction.} Let us recall that a key step in establishing the $L^2$-lower bound in Theorem~\ref{theorem:general_experts} is to decompose the regression discrepancy $f_{\widehat{G}_n}(x)-f_{G_*}(x)$ into a combination of linearly independent terms by applying Taylor expansions to the function $x\mapsto F(x;\omega,\eta)=\exp(\omega^{\top}x)\mathcal{E}(x,\eta)$. since the strong identifiability condition is not satisifed under the scenario of linear experts $\mathcal{E}(x,(a,b))=a^{\top}x+b$, there exist undesirable linear dependence among the function $F$ and its derivative with respect to parameters $a,b,\eta$ expressed via the following partial differential equation (PDE):
\begin{align}
    \label{eq:PDE_polynomial}
    \frac{\partial^2 F}{\partial\omega\partial b}(x;\omega^*_{i},a^*_{i},b^*_{i})=\frac{\partial F}{\partial a}(x;\omega^*_{i},a^*_{i},b^*_{i}).
\end{align}
We refer to the above PDE as an interaction between the gating parameters $\omega$ and the expert parameters $a,b$. This interaction has been encountered in the setting of softmax gating Gaussian mixture of linear experts \cite{nguyen2023demystifying}. More specifically, Nguyen et al. \cite{nguyen2023demystifying} showed that although parameter estimation rates were still of polynomial orders given
the aforementioned parameter interaction, these rates were significantly slow, as they hinged upon the solvability of some intrinsic system of polynomial equations. Subsequently, we will demonstrate in Theorem~\ref{theorem:linear_experts} that the parameter estimation rates become significantly slower than polynomial orders under the deterministic softmax gating MoE due to the parameter interaction in equation~\eqref{eq:PDE_polynomial}. For that purpose, let us define a Voronoi loss function tailored to the setting of linear experts as follows:
\begin{align}
    \label{eq:loss_D_2}
    \mathcal{L}_{2,r}(G,G_*):=\sum_{j=1}^{k_*}\Big|\sum_{i\in\mathcal{A}_j}\exp(\beta_{i})-\exp(\beta^*_{j})\Big|+\sum_{j=1}^{k_*}\sum_{i\in\mathcal{A}_j}\exp(\beta_{i})\Big[\|\Delta\omega_{ij}\|^r+\|\Delta a_{ij}\|^r+\|\Delta b_{ij}\|^r\Big],
\end{align}
for any $r\geq 1$, where we denote $\Delta a_{ij}:=a_i-a^*_j$ and $\Delta b_{ij}:=b_i-b^*_j$.
\begin{theorem}
    \label{theorem:linear_experts}
    Assume that the experts take the linear form $a^{\top}x+b$, then the following minimax lower bound of estimating $G_*$ holds true for any $r\geq 1$:
    \begin{align}
        \label{eq:linear_expert_bound}
        \inf_{\overline{G}_n\in\mathcal{G}_{k}(\Theta)}\sup_{G\in\mathcal{G}_{k}(\Theta)\setminus\mathcal{G}_{k_*-1}(\Theta)}\bbE_{f_{G}}[\mathcal{L}_{2,r}(\overline{G}_n,G)]\gtrsim n^{-1/2},
    \end{align}
    where $\bbE_{f_{G}}$ indicates the expectation taken with respect to the product measure with $f^n_{G}$ and the infimum is over all estimators taking values in $\mathcal{G}_{k}(\Theta)$.
\end{theorem}
\noindent
The proof of Theorem~\ref{theorem:linear_experts} is in Appendix~\ref{appendix:linear_experts}. The bound in equation~\eqref{eq:linear_expert_bound} reveals that the convergence rates of estimating ground-truth parameters $\omega^*_j$, $a^*_j$, and $b^*_j$ are slower than polynomial rates $\mathcal{O}_P(n^{-1/2r})$ for any $r\geq 1$. Therefore, these rates can become as slow as $\mathcal{O}_P(1/\log^{\lambda}(n))$ for some positive constant $\lambda$. Regarding the expert estimation, it is worth noting that
\begin{align*}
    \sup_{x} \Big|((\widehat{a}^n_i)^{\top}x+\widehat{b}^n_i)-((a^*_j)^{\top}x+b^*_j)\Big|\leq \sup_{x} \|\widehat{a}^n_i-a^*_j\|\cdot\|x\|+|\widehat{b}^n_i-b^*_j|.
\end{align*}
Recall from Assumption (A.1) that the input space $\mathcal{X}$ is bounded. Thus, the above inequality implies that the rates for estimating experts $(a^*_j)^{\top}x+b^*_j$ are also slower than any polynomial rates and could be of the order $\mathcal{O}_P(1/\log^{\lambda}(n))$. In that case, it requires  exponentially many data points of order $\mathcal{O}(\exp(\epsilon^{-1/\lambda}))$ to obtain an expert approximation with a predetermined error $\epsilon$. Compared to the results in Theorem~\ref{theorem:general_experts}, we see that using strongly identifiable experts is more sample efficient as we need only a polynomial number of data points to estimate them. 

\section{Dense-to-sparse Gating MoE}
\label{sec:dense-to-sparse_MoE}
\noindent
In this section, we investigate the convergence analysis of parameter estimation and expert estimation under the dense-to-sparse gating MoE. We begin our analysis with a linear router inside the dense-to-sparse gating in Section~\ref{sec:linear_router}, and then extend to the setting of a class of general routers, which are optimal for the parameter and expert convergence, in Section~\ref{sec:general_router}.

\subsection{Linear Router}
\label{sec:linear_router}
\noindent
Firstly, we will exhibit the problem setup for analyzing the convergence of dense-to-sparse gating MoE with a linear router. \\

\noindent
\textbf{Problem setting.} We assume that an i.i.d. sample $(X_{1}, Y_{1}), (X_{2}, Y_{2}), \ldots, (X_{n}, Y_{n})$ in $\mathcal{X}\times \mathcal{Y}\subseteq\mathbb{R}^d\times\mathbb{R}$ is also generated according to a regression framework as in equation~\eqref{eq:moe_regression_model} but with another MoE-based regression function as $Y_{i} = g_{G_{*}}(X_{i}) + \varepsilon_{i}$ for all $i\in[k_*]$, where the regression function $g_{G_{*}}(\cdot)$ is defined as
\begin{align}
    \label{eq:dense-to-sparse_MoE}
    g_{G_{*}}(x) := \sum_{i=1}^{k_*} \softmax\Big(\frac{(\omega^*_{i})^{\top}x+\beta^*_{i}}{\tau^*}\Big)\cdot \mathcal{E}(x,\eta^*_{i}).
\end{align}
Here, we keep the definitions and assumptions of the input $X_i$ and the noise variable $\varepsilon_i$ unchanged for all $i\in[k_*]$. Recall that the temperature $\tau^*$ is involved to control the sparsity level of the MoE model, making the expert selection process smooth and stable. In particular, when the temperature parameter goes to infinity, the weight distribution becomes roughly uniform, implying that all the experts are activated. In contrast, when the temperature parameter tends to zero, the mixture weights behave as a one-hot vector, meaning that only one expert is activated. Additionally, by abuse of notation, we still denote  $G_{*} := \sum_{i = 1}^{k_{*}} \exp(\beta^{*}_{i}) \delta_{(\omega^{*}_{i},\tau^*,\eta^*_{i})}$ as a mixing measure associated with unknown parameters $(\beta^*_{i},\omega^*_{i},\tau^*,\eta^*_{i})_{i=1}^{k_*}\in\Theta\subseteq\mathbb{R}\times\mathbb{R}^d\times\mathbb{R}_+\times\mathbb{R}^q$ although this formulation is different from that in the setting of softmax gating MoE due to the appearance of the temperature parameter $\tau^*$. Then, the least square estimator of the ground-truth mixing measure $G_*$ is now given by:
\begin{align}
    \label{eq:least_square_estimator_2}
    \widetilde{G}_n:=\argmin_{G\in\mathcal{G}_{k}(\Theta)}\sum_{i=1}^{n}\Big(Y_i-g_{G}(X_i)\Big)^2,
\end{align}
where $\mathcal{G}_{k}(\Theta):=\{G=\sum_{i=1}^{k'}\exp(\beta_{i})\delta_{(\omega_{i},\tau,\eta_{i})}:1\leq k'\leq k, \  (\beta_{i},\omega_{i},\tau,\eta_{i})\in\Theta\}$ denotes the set of all mixing measures with no more than $k$ atoms. Analogously to Section~\ref{sec:softmax_MoE}, we provide in the following proposition the convergence rate of estimating the ground-truth regression function $g_{G_*}(x)$.
\begin{proposition}
    \label{theorem:regression_rate_dense-to-sparse}
    For a least square estimator $\widetilde{G}_n$ in equation~\eqref{eq:least_square_estimator_2}, the convergence rate of regression function $g_{\widetilde{G}_n}$ is given by
    \begin{align}
        \label{eq:model_bound_2}
        \normf{g_{\widetilde{G}_n}-g_{G_*}}=\mathcal{O}_{P}([\log(n)/n]^{\frac{1}{2}}).
    \end{align}
\end{proposition}
\noindent
Since this proposition can be proved in a similar fashion to Proposition~\ref{theorem:regression_rate_softmax}, its proof is omitted. The above result reveals that the regression function estimation $g_{\widetilde{G}_n}$ converges to its ground-truth version $g_{G_*}$ at the parametric rate on the sample size, that is, $\mathcal{O}_{P}([\log(n)/n]^{\frac{1}{2}})$. Next, we proceed to determine the convergence rates of parameter and expert estimations based on this result. It comes to our attention that although the temperature helps stabilize and smooth the expert selection process, it induces an intrinsic interaction with gating parameters which might harm the parameter and expert convergence.\\

\noindent
\textbf{Interaction between the temperature and gating parameters.} Given the bound in equation~\eqref{eq:model_bound_2}, we continue to leverage the same strategy to determine  parameter and expert estimation rates as in Section~\ref{sec:softmax_MoE}, that is, to derive the $L^2$ lower bound $\normf{g_{\widetilde{G}_n}-g_{G_*}}\gtrsim\mathcal{L}(\widetilde{G}_n,G_*)$ for some loss function $\mathcal{L}$ that will be defined later. However, an obstacle arises when decomposing the discrepancy $\normf{g_{\widetilde{G}_n}-g_{G_*}}$ into a combination of linearly independent terms, which is a key step in the strategy. In particular, we observe that there is an interaction between the temperature parameter $\tau$ and the gating parameter $\omega$ via the following PDE:
\begin{align}
    \label{eq:PDE_dense-to-sparse}
    \frac{\partial F}{\partial\tau}(x;\omega^*_i,\tau^*,\eta^*_i)=-\frac{1}{\tau^*}(\omega^*_i)^{\top}\frac{\partial F}{\partial \omega}(x;\omega^*_i,\tau^*,\eta^*_i),
\end{align}
where we define $F(x;\omega,\tau,\eta):=\exp\Big(\frac{\omega^{\top}x}{\tau}\Big)\mathcal{E}(x,\eta)$ by abuse of notation. Unlike the interaction in equation~\eqref{eq:PDE_polynomial} which occurs when the expert function takes a linear form, the above interaction holds true irrespective of the expert structure. As a result, we will illustrate in Theorem~\ref{theorem:linear_dense-to-sparse_experts} that the parameter interaction~\eqref{eq:PDE_dense-to-sparse} negatively affects the convergence of parameter and expert estimations under the dense-to-sparse gating MoE by involving the following Voronoi loss function:
\begin{align}
    \label{eq:loss_log_over}
    \mathcal{L}_{3,r}(G,G_*)&:=
    \sum_{j=1}^{k_*}\Big|\sum_{i\in\mathcal{A}_j}\exp\Big(\frac{\beta_{0i}}{\tau}\Big)-\exp\Big(\frac{\beta^*_{0j}}{\tau^*}\Big)\Big|+\sum_{j=1}^{k_*}\sum_{i\in\mathcal{A}_j}\exp\Big(\frac{\beta_{0i}}{\tau}\Big)[\|\Delta \omega_{ij}\|^r+\|\Delta\tau\|^r+\|\Delta\eta_{ij}\|^r],
\end{align}
for any $r\geq 1$, where we denote $\Delta\tau:=\tau-\tau^*$.
\begin{theorem}
    \label{theorem:linear_dense-to-sparse_experts}
    The following minimax lower bound of estimating $G_*$ holds true for any $r\geq 1$:
    \begin{align*}
    \inf_{\overline{G}_n\in\mathcal{G}_{k}(\Theta)}\sup_{G\in\mathcal{G}_{k}(\Theta)\setminus\mathcal{G}_{k_*-1}(\Theta)}\bbE_{g_{G}}[\mathcal{L}_{3,r}(\overline{G}_n,G)]\gtrsim n^{-1/2}.
    \end{align*}
    Here, the notation $\bbE_{g_{G}}$ indicates the expectation taken with respect to the product measure with mixture density $g^n_{G}$.
\end{theorem}
\noindent
The proof of this theorem is deferred to Appendix~\ref{appendix:linear_dense-to-sparse_experts}. Similarly to Theorem~\ref{theorem:linear_experts}, the result of Theorem~\ref{theorem:linear_dense-to-sparse_experts} indicates that the rates for estimating parameters $\omega^*_j$, $\tau^*$, and $\eta^*_j$ are slower than polynomial rates of order $\mathcal{O}_P(n^{-1/2r})$ for any $r\geq 1$. Furthermore, it follows from the inequality~\eqref{eq:expert_rate} that the expert estimation shares the same convergence behavior. Thus, both the parameter and expert estimation rates could be as slow as $\mathcal{O}_P(1/\log^{\lambda}(n))$ for some constant $\lambda>0$. It should be noted that this slow estimation rate applies for any expert function employed in the dense-to-sparse gating MoE model with the linear router~\eqref{eq:dense-to-sparse_MoE}. This situation necessitates a new router formulation to avoid the parameter interaction~\eqref{eq:PDE_dense-to-sparse}, thereby improving the parameter and expert estimation rates. We will address this issue in the next section.

\subsection{General Router}
\label{sec:general_router}
\noindent
In this section, we aim to explore a class of new router structures such that the interaction between the temperature and gating parameters via the PDE~\eqref{eq:PDE_dense-to-sparse}, which induces slow expert estimation rates, no longer exists. For that purpose, we first generalize the ground-truth regression function $g_{G_*}(x)$ in equation~\eqref{eq:dense-to-sparse_MoE} by replacing a linear router $(\omega^*_j)^{\top}x$ with a general router $\pi(x,\omega^*_j)$ as follows:
\begin{align}
    \label{eq:general_dense-to-sparse_MoE}
    g_{G_{*}}(x) := \sum_{i=1}^{k_*} \softmax\Big(\frac{\pi(x,\omega^*_{i})+\beta^*_{i}}{\tau^*}\Big)\cdot \mathcal{E}(x,\eta^*_{i}).
\end{align}
Here, we will keep all the assumptions imposed on the setting of linear router unchanged throughout this section unless stating otherwise. Next, we introduce a so-called \emph{algebraic independence} condition on the router function $x\mapsto\pi(x,\omega)$ and the expert function $\mathcal{E}(x,\eta)$ in Definition~\ref{def:distinguishability_condition} to prevent any interaction among the temperature, gating and expert parameters expressed in the language of PDEs such as those in equation~\eqref{eq:PDE_dense-to-sparse} from happening.\\
\begin{definition}[Algebraic Independence]
    \label{def:distinguishability_condition}
    We say that a router function $x\mapsto\pi(x,\omega)$ and an expert function $x \mapsto \mathcal{E}(x,\eta)$ are algebraically independent if they are twice differentiable w.r.t their parameters $\omega$ and $\eta$, and the set of functions in $x$
    \begin{align*}
        \Big\{&\mathcal{E}(x,\eta_j),\pi(x,\omega_j)\mathcal{E}(x,\eta_j),\frac{\partial\pi}{\partial\omega^{(u)}}(x,\omega_j)\mathcal{E}(x,\eta_j),\frac{\partial\pi}{\partial\omega^{(u)}}(x,\omega_j)\frac{\partial\pi}{\partial\omega^{(v)}}(x,\omega_j)\mathcal{E}(x,\eta_j),\\
    &\frac{\partial^2\pi}{\partial\omega^{(u)}\partial\omega^{(v)}}(x,\omega_j)\mathcal{E}(x,\eta_j), \pi(x,\omega_j)\frac{\partial\pi}{\partial\omega^{(u)}}(x,\omega_j)\mathcal{E}(x,\eta_j), \frac{\partial\mathcal{E}}{\partial\eta^{(u)}}(x,\eta_j), \pi(x,\omega_j)\frac{\partial\mathcal{E}}{\partial\eta^{(u')}}(x,\eta_j),\\
    &\hspace{2cm}\frac{\partial\pi}{\partial\omega^{(u)}}(x,\omega_j)\frac{\partial\mathcal{E}}{\partial\eta^{(u')}}(x,\eta_j),\frac{\partial^2\mathcal{E}}{\partial\eta^{(u')}\partial\eta^{(v')}}(x,\eta_j):j\in[k],1\leq u,v\leq d, 1\leq u',v'\leq q\Big\}.
    \end{align*}
    is linearly independent for almost every $x$ for any $k\geq 1$, distinct gating parameters $\omega_1,\omega_2,\ldots,\omega_k$ and distinct expert parameters $\eta_1,\eta_2,\ldots,\eta_k$.
\end{definition}

\noindent
\textbf{Example.} It can be validated that the router function $\pi(x,\omega)=\sigma_1(\omega^{\top}x)$ and the expert function $\mathcal{E}(x,\eta)=c\sigma_2\Big(a\frac{x}{\|x\|}+b\Big)$ are algebraically independent, where $\sigma_1,\sigma_2$ are two different activation functions among the functions $\mathrm{GELU}$, $\mathrm{sigmoid}$ and $\mathrm{tanh}$, and $\omega\in\mathbb{R}^{q}$, $x,a\in\mathbb{R}^d$, $b,c,\in\mathbb{R}$. However, if the activation functions $\sigma_1$ and $\sigma_2$ take polynomial forms $z\in\mathbb{R}\mapsto z^p$, for some $p\in\mathbb{N}$, then the previous pair of router function and expert function violates the algebraic independence condition.\\


\noindent
Given the algebraic independence condition on the router function and the expert function, we are now ready to analyze the convergence of parameter and expert estimations under the dense-to-sparse gating MoE in Theorem~\ref{theorem:general_dense-to-sparse_experts}, which involves the following Voronoi loss function:
\begin{align}
    \mathcal{L}_{4}(G,G_*):=
    \sum_{j=1}^{k_*}\Big|\sum_{i\in\mathcal{A}_j}\exp\Big(\frac{\beta_{0i}}{\tau}\Big)-\exp\Big(\frac{\beta^*_{0j}}{\tau^*}\Big)\Big|&+\sum_{j:|\mathcal{A}_j|=1 }\sum_{i\in\mathcal{A}_j}\exp\Big(\frac{\beta_{0i}}{\tau}\Big)[\|\Delta \omega_{ij}\|+\|\Delta\tau\|+\|\Delta\eta_{ij}\|]\nonumber\\
    &+\sum_{j=1}^{k_*}\sum_{i\in\mathcal{A}_j}\exp\Big(\frac{\beta_{0i}}{\tau}\Big)[\|\Delta \omega_{ij}\|^2+\|\Delta\tau\|^2+\|\Delta\eta_{ij}\|^2].
\end{align}
\begin{theorem}
    \label{theorem:general_dense-to-sparse_experts}
    Suppose that the router function $x\mapsto\pi(x,\omega)$ and the expert function $x\mapsto\mathcal{E}(x,\eta)$ are algebraically independent, then the lower bound $\normf{g_{G}-g_{G_*}}\gtrsim\mathcal{L}_{4}(G,G_*)$ holds for all $G\in\mathcal{G}_{k}(\Theta)$, indicating that
    \begin{align}
    \label{eq:general_dense-to-sparse_experts_bound}
        \mathcal{L}_{4}(\widetilde{G}_n,G_*)=\mathcal{O}_{P}([\log(n)/n]^{\frac{1}{2}}).
    \end{align} 
\end{theorem}
\noindent
The proof of Theorem~\ref{theorem:general_dense-to-sparse_experts} is in Appendix~\ref{appendix:general_dense-to-sparse_experts}.\\

\noindent
\emph{(i) Parameter estimation rates:} From the  bound~\eqref{eq:general_dense-to-sparse_experts_bound} and the formulation of the Voronoi loss $\mathcal{L}_4$, we deduce that exactly-specified parameters $\omega^*_j,\tau^*,\eta^*_j$ enjoy estimation rates of the order $\mathcal{O}_{P}([\log(n)/n]^{\frac{1}{2}})$, whereas those for their over-specified counterparts are of the order $\mathcal{O}_{P}([\log(n)/n]^{\frac{1}{4}})$. Compared to the linear router scenario in Theorem~\ref{theorem:linear_dense-to-sparse_experts}, we see that the convergence rates of parameter estimation are significantly improved when the router function and the expert function are algebraically independent.\\

\noindent
\emph{(ii) Expert estimation rates:} By employing the inequality~\eqref{eq:expert_rate}, we notice that the convergence rates of expert estimation also benefit from the enhancement of parameter estimation rates. In particular, ground-truth experts $\mathcal{E}(x,\eta^*_j)$ admit estimation rates of the order $\mathcal{O}_{P}([\log(n)/n]^{\frac{1}{2}})$ when they are exactly-specified, and of the order $\mathcal{O}_{P}([\log(n)/n]^{\frac{1}{4}})$ when they are over-specified. As a consequence, only polynomially many data points of order $\mathcal{O}(\epsilon^{-2})$ or $\mathcal{O}(\epsilon^{-4})$ are required to obtain the expert approximation with the error $\epsilon$, which are substantially improved in comparison with exponentially many data points required in the linear router scenario. This result indicates that the algebraic independence between the router function and the expert function helps increase the model sample efficiency.

\section{Hierarchical Softmax Gating MoE}
\label{sec:hierarchical_MoE}
\noindent
In this section, we extend our previous convergence analysis of parameter estimation and expert estimation to the setting of hierarchical MoE (HMoE) model. For simplicity, we will consider only the two-level structure of the HMoE with a note that higher-level HMoE models can be analyzed in a similar fashion. \\

\noindent
\textbf{Problem setting.} Here, we continue to assume that $(X_{1}, Y_{1}), (X_{2}, Y_{2}), \ldots, (X_{n}, Y_{n})$ in $\mathcal{X}\times \mathcal{Y}\subseteq\mathbb{R}^d\times\mathbb{R}$ is drawn from a regression framework $Y_{i} = h_{G_{*}}(X_{i}) + \varepsilon_{i}$ for all $i\in[k_*]$, with the HMoE-based regression function $h_{G_{*}}(\cdot)$ given by 
\begin{align}
    \label{eq:hierarchical_MoE}
    h_{G_{*}}(x) := \sum_{i_1=1}^{k_1^*} \softmax((\omega^*_{i_1})^{\top}x+\beta^*_{i_1})\sum_{i_2=1}^{k_2^*}\softmax((\kappa^*_{i_2|i_1})^{\top}x+\nu^*_{i_2|i_1})\cdot \mathcal{E}(x,\eta^*_{i_1i_2}).
\end{align}
Note that all the definitions and assumptions of the input $X_i$ and the noise variable $\varepsilon_i$ from previous sections will still be applied in this setting. Meanwhile, the ground-truth mixing measure $G_*$ inherits the hierarchical structure of the HMoE and is formulated as $G_{*} := \sum_{i_1=1}^{k^*_1}\exp(\beta^*_{i_1})\sum_{i_2=1}^{k^*_2}\exp(\nu^*_{i_2|i_1})\delta_{(\omega^*_{i_1},\kappa^*_{i_2|i_1},\eta^*_{i_1i_2})},$ where  $(\beta^*_{i_1},\omega^*_{i_1},\nu^*_{i_2|i_1},\kappa^*_{i_2|i_1},\eta^*_{i_1i_2})_{i_1\in[k^*_1],i_2\in[k^*_2]}$ denotes true yet unknown parameters in the compact space $\Theta\subseteq\mathbb{R}\times\mathbb{R}^d\times\mathbb{R}\times\mathbb{R}^d\times\mathbb{R}^q$. Throughout this work, we refer to $k^*_1$ as the number of expert groups, while $k^*_2$ represents the number of experts in each group. As the convergence analysis would become needlessly complicated if the number of expert groups $k^*_1$ was unknown, we assume that its value is known for ease of presentation. Nevertheless, the number of experts in each group $k^*_2$ still remains unknown. Under these assumptions, we over-specify the ground-truth model~\eqref{eq:hierarchical_MoE} by taking into account a least square estimator within a class of HMoE models with $k^*_1$ expert groups, each of which has at most $k_2>k^*_2$ experts, as follows:
\begin{align}
    \label{eq:least_square_estimator_3}
    \check{G}_n:=\argmin_{G\in\mathcal{G}_{k^*_1k_2}(\Theta)}\sum_{i=1}^{n}\Big(Y_i-h_{G}(X_i)\Big)^2,
\end{align}
with
\begin{align*}
    \mathcal{G}_{k^*_1k_2}(\Theta):=\{G=\sum_{i_1=1}^{k^*_1}\exp(\beta_{i_1})\sum_{i_2=1}^{k'_2}\exp(\nu_{i_2|i_1})\delta_{(\omega_{i_1},\kappa_{i_2|i_1},\eta_{i_2|i_1})}&:1\leq k'_2\leq k_2, \\
    &(\beta_{i_1},\omega_{i_1},\nu_{i_2|i_1},\kappa_{i_2|i_1},\eta_{i_1i_2})\in\Theta\}
\end{align*}
standing for the set of all feasible mixing measures. Given the above estimator, we will demonstrate that the corresponding regression function estimation $h_{\check{G}_n}(x)$ converges to the ground-truth regression function $h_{G_*}(x)$ at the parametric rate on the sample size in Proposition~\ref{theorem:regression_rate_hmoe} whose proof is left in Appendix~\ref{appendix:regression_rate_hmoe}.
\begin{proposition}
    \label{theorem:regression_rate_hmoe}
    For a least square estimator $\check{G}_n$ in equation~\eqref{eq:least_square_estimator_3}, the convergence rate of regression function $h_{\check{G}_n}$ is given by
    \begin{align}
        \label{eq:model_bound_3}
        \normf{h_{\check{G}_n}-h_{G_*}}=\mathcal{O}_{P}([\log(n)/n]^{\frac{1}{2}}).
    \end{align}
\end{proposition}
\noindent
Toward the goal of analyzing the convergence behavior of parameter and expert estimations in the HMoE, we also need to decompose the difference of regression functions $h_{\check{G}_n}(x)-h_{G_*}(x)$ into a combination of linearly independent terms. However, due to the hierarchical structure of the HMoE, this decomposition will be achieved by applying Taylor expansions to the function $x\mapsto \check{F}(x;\omega,\kappa,\eta):=\exp(\omega^{\top}x)\exp(\kappa^{\top}x)\mathcal{E}(x,\eta)$ rather than the function $x\mapsto F(x;\omega,\eta):=\exp(\omega^{\top}x)\mathcal{E}(x,\eta)$ as in the MoE setting. Recall that, the function $\check{F}$ and its partial derivatives from the Taylor expansions are forced to be linearly independent so that when the previous regression difference approaches zero as $n\to\infty$, the coefficients of those terms, which contain parameter discrepancies, also go to zero, ensuring the convergence of parameter estimation. Notably, we will show in Section~\ref{sec:hierarchical_strongly_identifiable_experts} that if the expert function $\mathcal{E}(x,\eta)$ meets the strong identifiability condition in Definition~\ref{def:softmax_condition}, then the aforementioned linear independence condition is secured and, therefore, ground-truth parameters and experts admit estimation rates of polynomial orders. On the other hand, if the expert function takes a linear form $\mathcal{E}(x,(a,b))=a^{\top}x+b$, then it is not strongly identifiable, leading to slow parameter and expert estimation rates exhibited in Section~\ref{sec:hierarchical_linear_experts}.

\subsection{Strongly identifiable experts}
\label{sec:hierarchical_strongly_identifiable_experts}
\noindent
Before presenting the result statement, it is necessary to construct a Voronoi loss function tailored to the HMoE setting.\\

\begin{figure*}[!ht]
    \centering
    \includegraphics[scale =.3]{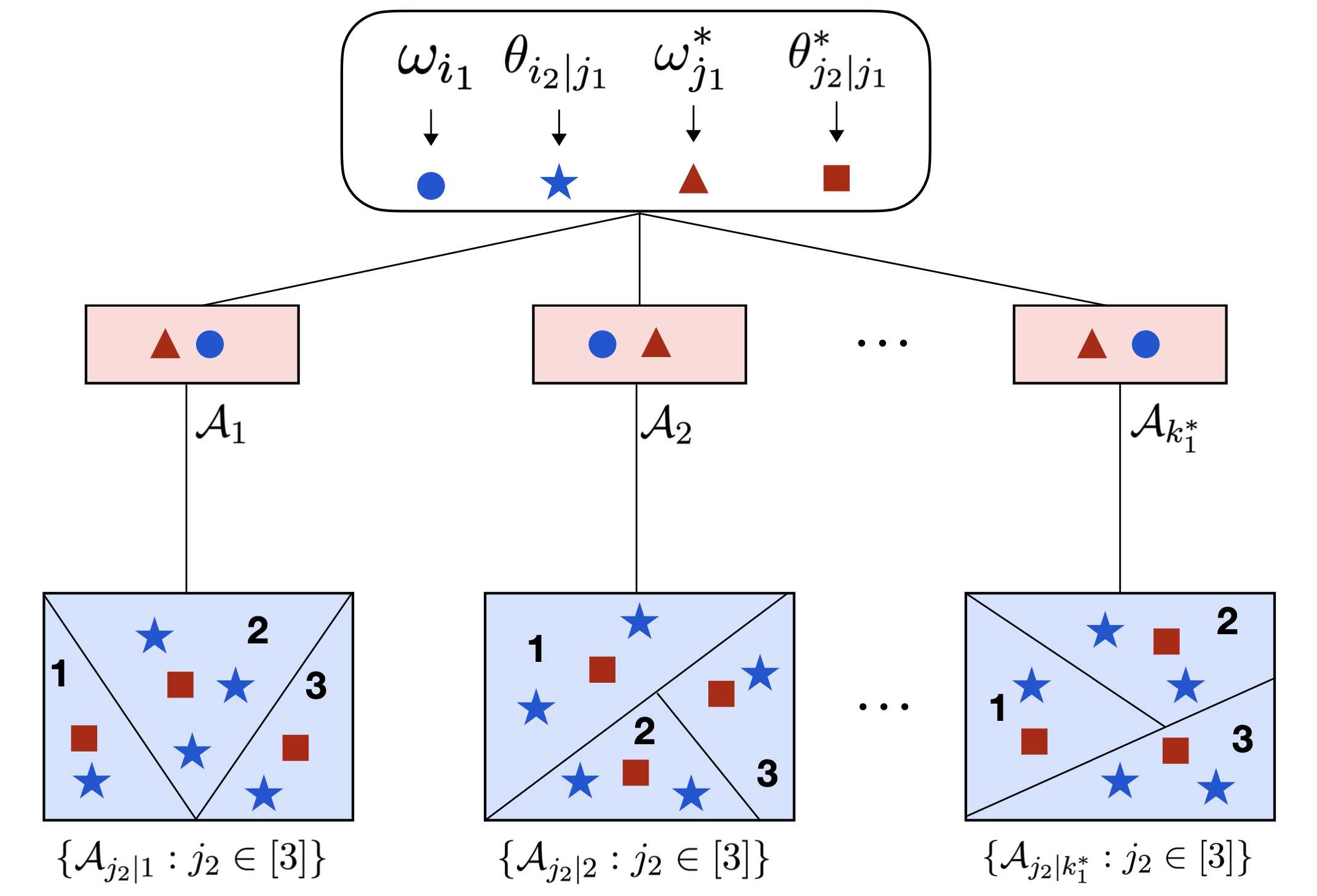}
    \caption{Illustration of Voronoi cells given in equations~\eqref{eq:Voronoi_cells_level_1} and \eqref{eq:Voronoi_cells_level_2}. Above, Voronoi cells $\mathcal{A}_{1},\mathcal{A}_{2},\ldots,\mathcal{A}_{k^*_1}$ in the first level are generated by ground-truth parameters $\omega^*_{1},\omega^*_{2},\ldots,\omega^*_{k^*_1}$ (red triangles), respectively. As the number of expert groups $k_1^*$ is known, each Voronoi cell $\mathcal{A}_{j_1}$ contains one fitted parameter $\omega_{i_1}$ (blue round). In the second level, each rectangle represent a set of $k_2^*=3$ Voronoi cells $\{\mathcal{A}_{j_2|j_1}:j_2\in[k_2^*]\}$ generated by ground-truth parameters $\theta^*_{j_2|j_1}:=(\kappa^*_{j_2|j_1},\eta^*_{j_1j_2})$(red squares), for $j_1\in[k_1^*]$, each of which contains a total of $k_2=5$ fitted parameters $\theta_{i_2|j_1}:=(\kappa_{i_2|j_1},\eta_{i_2j_1})$ (blue stars). 
    }	
    \label{fig:hierarchical_voronoi_cells}
\end{figure*}

\noindent
\textbf{Voronoi loss function.} Due to the two-level hierarchical structure of the HMoE, we need to employ distinct sets of Voronoi cells to capture the convergence of parameter estimations in each level. In particular, given an arbitrary mixing measure $G\in\mathcal{G}_{k^*_1k_2}(\Theta)$, we partition its atoms into the Voronoi cells $\{\mathcal{A}_{j_1}(G):j_1 \in [k^*_1]\}$ and $\{\mathcal{A}_{j_2|j_1}(G):j_1 \in [k^*_1],j_2\in[k^*_2]\}$ generated by the atoms of the ground-truth mixing measure $G_*$ (see Figure~\ref{fig:hierarchical_voronoi_cells}), where
\begin{align}
    \label{eq:Voronoi_cells_level_1}
    \mathcal{A}_{j_1}\equiv\mathcal{A}_{j_1}(G)&:=\{i_1\in[k^*_1]:\|\omega_{i_1}-\omega^*_{j_1}\|\leq\|\omega_{i_1}-\omega^*_{\ell_1}\|,\forall \ell_1\neq j_1\},\\
    \label{eq:Voronoi_cells_level_2}
    \mathcal{A}_{j_2|j_1}\equiv\mathcal{A}_{j_2|j_1}(G)&:=\{i_2\in[k_2]:\|\theta_{i_2|j_1}-\theta^*_{j_2|j_1}\|\leq\|\theta_{i_2|j_1}-\theta^*_{\ell_2|j_1}\|,\forall \ell_2\neq j_2\},
\end{align}
with $\theta_{i_2|j_1}:=(\kappa_{i_2|j_1},\eta_{j_1i_2})$ and $\theta^*_{j_2|j_1}:=(\kappa^*_{j_2|j_1},\eta^*_{j_1j_2})$. Then, the Voronoi loss function of interest is defined as 
\begin{align}
    &\mathcal{L}_{5}(G,G_*):=\sum_{j_1=1}^{k^*_1}\Big|\sum_{i_1\in\mathcal{A}_{j_1}}\exp(\beta_{i_1})-\exp(\beta^*_{j_1})\Big|+\sum_{j_1=1}^{k^*_1}\sum_{i_1\in\mathcal{A}_{j_1}}\exp(\beta_{i_1})\|\omega_{i_1}-\omega^*_{j_1}\|\nonumber\\
    &\hspace{2cm}+\sum_{j_1=1}^{k^*_1}\sum_{i_1\in\mathcal{A}_{j_1}}\exp(\beta_{i_1})\Bigg[\sum_{j_2:|\mathcal{A}_{j_2|j_1}|=1}\sum_{i_2\in\mathcal{A}_{j_2|j_1}}\exp(\nu_{i_2|i_1})(\|\kappa_{i_2|i_1}-\kappa^*_{j_2|j_1}\|+\|\eta_{i_1i_2}-\eta^*_{j_1j_2}\|)\nonumber\\
    &\hspace{5.2cm}+\sum_{j_2:|\mathcal{A}_{j_2|j_1}|>1}\sum_{i_2\in\mathcal{A}_{j_2|j_1}}\exp(\nu_{i_2|i_1})(\|\kappa_{i_2|i_1}-\kappa^*_{j_2|j_1}\|^{2}+\|\eta_{i_1i_2}-\eta^*_{j_1j_2}\|^{2})\Bigg]\nonumber\\
    &\hspace{2cm}+\sum_{j_1=1}^{k^*_1}\sum_{i_1\in\mathcal{A}_{j_1}}\exp(\beta_{i_1})\sum_{j_2=1}^{k^*_2}\Big|\sum_{i_2\in\mathcal{A}_{j_2|j_1}}\exp(\nu_{i_2|i_1})-\exp(\nu^*_{j_2|j_1})\Big|.
\end{align}
Equipped with the Voronoi loss $\mathcal{L}_5$, we will illustrate that strongly identifiable experts in the HMoE model admit estimation rates of polynomial orders in Theorem~\ref{theorem:hierarchical_general_experts} whose proof is left in Appendix~\ref{appendix:hierarchical_general_experts}.
\begin{theorem}
    \label{theorem:hierarchical_general_experts}
    Suppose that the expert function $x\mapsto\mathcal{E}(x,\eta)$ satisfies the strong identifiability condition presented in Definition~\ref{def:softmax_condition}, then the lower bound $\normf{h_{G}-h_{G_*}}\gtrsim\mathcal{L}_{5}(G,G_*)$ holds for all $G\in\mathcal{G}_{k^*_1k_2}(\Theta)$, indicating that
    \begin{align}
        \label{eq:bound_hierarchical}
        \mathcal{L}_{5}(\check{G}_n,G_*)=\mathcal{O}_{P}([\log(n)/n]^{\frac{1}{2}}).
    \end{align}
\end{theorem}
\noindent
We provide below some implications on the parameter and expert estimation rates from the above results.\\

\noindent
\emph{(i) Parameter estimation rates:} It follows from the formulation of the Voronoi loss $\mathcal{L}_{5}$ and the bound~\eqref{eq:bound_hierarchical} that all the first-level parameters $\omega^*_{j_1}$ share the same estimation rate of the order $\mathcal{O}_{P}([\log(n)/n]^{\frac{1}{2}})$. Regarding the second-level parameters $\kappa^*_{j_2|j_1}$ and $\eta^*_{j_1j_1}$, the rates for estimating them are of the order $\mathcal{O}_{P}([\log(n)/n]^{\frac{1}{2}})$ if they are exactly-specified, that is, $|\mathcal{A}_{j_2|j_1}|=1$. However, if they are over-specified, that is, $|\mathcal{A}_{j_2|j_1}|>1$, their estimation rates become slower of the order $\mathcal{O}_{P}([\log(n)/n]^{\frac{1}{4}})$. \\

\noindent
\emph{(ii) Expert estimation rates:} Recall that a strongly identifiable expert function is twice differentiable over a bounded domain, so it is also a Lipschitz function, which implies that 
\begin{align}
    \sup_{x} |\mathcal{E}(x,\check{\eta}^n_{i_1i_2})-\mathcal{E}(x,\eta^*_{j_1j_2})| &\lesssim\|\check{\eta}^n_{i_1i_2}-\eta^*_{j_1j_2}\|.  \label{eq:expert_rate_hierarchical}
\end{align}
Together with the above parameter estimation rates, this inequality indicates that exactly-specified expert $\mathcal{E}(x,\eta^*_{j_1j_2})$ admit the estimation rates of the order $\mathcal{O}_{P}([\log(n)/n]^{\frac{1}{2}})$, while those for their over-specified counterparts are slower of the order $\mathcal{O}_{P}([\log(n)/n]^{\frac{1}{4}})$. As a consequence, we need polynomially many data points of the order $\mathcal{O}(\epsilon^{-2})$ and $\mathcal{O}(\epsilon^{-4})$ to approximate these experts with a given error $\epsilon$. It can be seen that the expert convergence behavior under the two-level HMoE is similar to that under the MoE considered in Section~\ref{sec:softmax_MoE}.

\subsection{Linear experts}
\label{sec:hierarchical_linear_experts}
\noindent
In this section, we draw our attention to the convergence analysis for the hierarchical mixture of linear experts taking the form $\mathcal{E}(x,(a,b))=a^{\top}x+b$ for $(a,b)\in\mathbb{R}^d\times\mathbb{R}$, which fail to meet the strong identifiability condition in Definition~\ref{def:softmax_condition}.\\

\noindent
\textbf{Parameter interaction.} In order to obtain an $L^2$-lower bound as in Theorem~\ref{theorem:hierarchical_general_experts}, a crucial step is to apply Taylor expansions to the function $x\mapsto \check{F}(x;\omega,\kappa,a,b)=\exp(\omega^{\top}x)\exp(\kappa^{\top}x)(a^{\top}x+b)$ to decompose the regression discrepancy $h_{\check{G}_n}(x)-h_{G_*}(x)$ into a combination of linearly independent terms. Nevertheless, we discover that this step is hindered by interactions among parameters $\omega,\kappa,a,b$ from both levels of the HMoE model via the following PDEs:
\begin{align}
    \label{eq:PDE_hierarchical_polynomial}
    \frac{\partial^2 \check{F}}{\partial\omega\partial b}(x;\omega^*_{j_1},\kappa^*_{j_2|j_1},a^*_{j_1j_2},b^*_{j_1j_2})=\frac{\partial^2 \check{F}}{\partial\kappa\partial b}(x;\omega^*_{j_1},\kappa^*_{j_2|j_1},a^*_{j_1j_2},b^*_{j_1j_2})=\frac{\partial \check{F}}{\partial a}(x;\omega^*_{j_1},\kappa^*_{j_2|j_1},a^*_{j_1j_2},b^*_{j_1j_2}).
\end{align}
As a result of the above parameter interactions, we will exhibit in Theorem~\ref{theorem:hierarchical_linear_experts} that the convergence rates of parameter and expert estimations in the HMoE are negatively affected to be slower than polynomial orders. It is worth noting that since the number of expert groups $k^*_1$ is known, the first-level parameters $\omega^*_{j_1}$ are exactly-specified and, thus, should have estimation rates of polynomial orders as in Theorem~\ref{theorem:hierarchical_general_experts}. Nevertheless, due to their interactions with second-level parameters $\kappa^*_{j_2|j_1}$ and $a^*_{j_1j_2},b^*_{j_1j_2}$ in equation~\eqref{eq:PDE_hierarchical_polynomial}, the rates for estimating them will also be decelerated. To capture such convergence behavior, we will employ the following Voronoi loss function for our analysis:
\begin{align}
    \label{eq:loss_D_6}
    &\mathcal{L}_{6,r}(G,G_*):=\sum_{j_1=1}^{k^*_1}\Big|\sum_{i_1\in\mathcal{A}_{j_1}}\exp(\beta_{i_1})-\exp(\beta^*_{j_1})\Big|+\sum_{j_1=1}^{k^*_1}\sum_{i_1\in\mathcal{A}_{j_1}}\exp(\beta_{i_1})\|\omega_{i_1}-\omega^*_{j_1}\|^r\nonumber\\
    &+\sum_{j_1=1}^{k^*_1}\sum_{i_1\in\mathcal{A}_{j_1}}\exp(\beta_{i_1})\sum_{j_2=1}^{k^*_2}\sum_{i_2\in\mathcal{A}_{j_2|j_1}}\exp(\nu_{i_2|i_1})(\|\kappa_{i_2|i_1}-\kappa^*_{j_2|j_1}\|^r+\|a_{i_1i_2}-a^*_{j_1j_2}\|^r+|b_{i_1i_2}-b^*_{j_1j_2}|^r)\nonumber\\
    &+\sum_{j_1=1}^{k^*_1}\sum_{i_1\in\mathcal{A}_{j_1}}\exp(\beta_{i_1})\sum_{j_2=1}^{k^*_2}\Big|\sum_{i_2\in\mathcal{A}_{j_2|j_1}}\exp(\nu_{i_2|i_1})-\exp(\nu^*_{j_2|j_1})\Big|,
\end{align}
for any $r\geq 1$. Now, we are ready to present the main result of this section.
\begin{theorem}
    \label{theorem:hierarchical_linear_experts}
    Assume that the experts take the linear form $a^{\top}x+b$, then the following minimax lower bound of estimating $G_*$ holds true for any $r\geq 1$:
    \begin{align}
        \label{eq:hierarchical_linear_expert_bound}
        \inf_{\overline{G}_n\in\mathcal{G}_{k^*_1k_2}(\Theta)}\sup_{G\in\mathcal{G}_{k^*_1k_2}(\Theta)\setminus\mathcal{G}_{k^*_1(k^*_2-1)}(\Theta)}\bbE_{h_{G}}[\mathcal{L}_{6,r}(\overline{G}_n,G)]\gtrsim n^{-1/2},
    \end{align}
    where $\bbE_{h_{G}}$ indicates the expectation taken with respect to the product measure with $h^n_{G}$ and the infimum is over all estimators taking values in $\mathcal{G}_{k^*_1k_2}(\Theta)$.
\end{theorem}
\noindent
The proof of Theorem~\ref{theorem:hierarchical_linear_experts} is in Appendix~\ref{appendix:hierarchical_linear_experts}. Putting the above result and the formulation of the Voronoi loss $\mathcal{L}_{6,r}$ together, it follows that the estimation rates for the parameters $\omega^*_{j_1},\kappa^*_{j_2|j_1},a^*_{j_1j_2},b^*_{j_1j_2}$, which are involved in the interactions~\eqref{eq:PDE_hierarchical_polynomial}, are slower than polynomial orders $\mathcal{O}_P(n^{-1/2r})$ for any $r\geq 1$. Thus, these parameter estimation rates can become as slow as $\mathcal{O}_P(1/\log^{\lambda}(n))$ for some constant $\lambda>0$. Furthermore, expert estimation rates are also affected by slow estimation rates of the expert parameters $a^*_{j_1j_2},b^*_{j_1j_2}$ via the following inequality:
\begin{align*}
    \sup_{x} \Big|((\check{a}^n_{i_1i_2})^{\top}x+\check{b}^n_{i_1i_2})-((a^*_{j_1j_2})^{\top}x+b^*_{j_1j_2})\Big|\leq \sup_{x} \|\check{a}^n_{i_1i_2}-a^*_{j_1j_2}\|\cdot\|x\|+|\check{b}^n_{i_1i_2}-b^*_{j_1j_2}|.
\end{align*}
In particular, since the input space $\mathcal{X}$ is assumed to be bounded, then the estimation rates for the experts $(a^*_{j_1j_2})^{\top}x+b^*_{j_1j_2}$ are slower than any polynomial order and, therefore, could be as slow as $\mathcal{O}_P(1/\log^{\lambda}(n))$. Consequently, we would need exponentially many data points of order $\mathcal{O}(\exp(\epsilon^{-1/\lambda}))$ to approximate them with a predetermined error $\epsilon$, which is significantly larger than a polynomial number of data points needed for strongly identifiable experts. Hence, the claim that using strongly identifiable experts is more sample efficient than employing linear experts still holds under the HMoE setting.

\section{Discussion}
\label{sec:conclusion}

In this paper, we aim to investigate the impacts of the softmax gating and its variants, namely the dense-to-sparse gating and the hierarchical softmax gating, on the convergence behavior of parameter estimation and expert estimation in the mixture of experts (MoE). We show that the convergence rates of expert estimation are of polynomial orders when the expert function meets the strong identifiability condition. However, when using the dense-to-sparse gating with a linear router, the expert estimation rates become slower than any polynomial rates regardless of the expert structure. In response to this issue, we establish the algebraic independence condition to characterize the pairs of a router function and an expert function that lead to improved parameter and expert estimation rates of polynomial orders. On the other hand, our findings reveal that the convergence rates of estimating linear experts are always slower than polynomial rates irrespective of the gating choice. In conclusion, our convergence analysis provide two following important practical implications for the design of expert and gating structures:

\vspace{0.5em}
\noindent
\emph{(i) Expert design:} Using strongly identifiable experts, namely experts formulated as two-layer feed-forward networks, is more sample-efficient than employing linear experts. 

\vspace{0.5em}
\noindent
\emph{(ii) Gating design:} When incorporating the temperature parameter into the MoE to smoothly adjust the model sparsity, it is necessary to replace a linear router with a general router that is algebraically independent of the expert function. For example, when experts are two-layer feed-forward networks, the router should take the form of a multi-layer perceptron following by a non-linear activation function.

\vspace{0.5em}
\noindent
At the same time, there are two inherent limitations in the analysis. Firstly, we assume that the data are generated from an MoE-based regression framework, which could not be satisfied in the real-world scenario where the regression function $s(\cdot)$ takes an arbitrary form. In that case, the least square estimator $\widehat{G}_n$ converges to the set of mixing measures $\overline{G}\in\mathcal{G}_{k}(\Theta)$ which minimizes the distance $\|f_{G}-s\|_{L^2(\mu)}$. Note that the current techniques of analyzing the convergence of least square estimation are tailored to the scenario when the function space is convex \cite{vandeGeer-00}. Since the space $\mathcal{G}_{k}(\Theta)$ is non-convex, it is necessary to develop further techniques to handle this real-world setting. Secondly, for the sake of theory, we consider only a single MoE layer, while practitioners often employ multiple MoE layers in practice \cite{dai2024deepseekmoe,jiang2024mixtral}. As these two potential directions stay beyond the scope of our work, we leave them for future development.\\

\appendix
\centering
\textbf{\Large{Supplement to \vspace{.1in}\\
``Convergence Rates for Softmax Gating Mixture of Experts''}}

\justifying
\setlength{\parindent}{0pt}

\section{Proof of Theorem~\ref{theorem:general_experts}}
\label{appendix:general_experts}
\noindent
\textbf{Proof overview.} In this proof, we focus on establishing the following inequality:
\begin{align}
    \label{eq:activation_universal_inequality}
    \inf_{G\in\mathcal{G}_{k}(\Theta)}\normf{f_{G}-f_{G_*}}/\mathcal{L}_{1}(G,G_*)>0
\end{align}
For this sake, we first derive the local part of the inequality~\eqref{eq:activation_universal_inequality}, which is given by
\begin{align}
    \label{eq:general_local_inequality}
    \lim_{\varepsilon\to0}\inf_{G\in\mathcal{G}_{k}(\Theta):\mathcal{L}_1(G,G_*)\leq\varepsilon}\normf{f_{G}-f_{G_*}}/\mathcal{L}_1(G,G_*)>0.
\end{align}
We will prove the local part by contradiction. In particular, assume that it does not hold, then there exists a mixing measure sequence $(G_n)$ such that $\mathcal{L}_1(G_n,G_*)\to0$ and $\|f_{G_n}-f_{G_*}\|_{L^2(\mu)}/\mathcal{L}_1(G_n,G_*)\to0$ as $n\to\infty$. Then, we divide the proof into three main stages.

\vspace{0.5em}
\noindent
\emph{Step 1 - Decompose the difference between regression functions.} Firstly, we aim to decompose the regression discrepancy $f_{G_n}(x)-f_{G_*}(x)$ using Taylor expansions.

\vspace{0.5em}
\noindent
\emph{Step 2 - Non-vanishing coefficients.} Secondly, we show that not all the coefficients in the decomposition of $[f_{G_n}(x)-f_{G_*}(x)]/\mathcal{L}_1(G_n,G_*)$ in Step 1 converge to zero. 

\vspace{0.5em}
\noindent
\emph{Step 3 - Applications of the Fatou's lemma.} Finally, we show a contradiction to the result in Step 2. Indeed, by means of the Fatou's lemma, since the term $\|f_{G_n}-f_{G_*}\|_{L^2(\mu)}/\mathcal{L}_1(G_n,G_*)\to0$ as $n\to\infty$, we deduce $|f_{G_n}-f_{G_*}|/\mathcal{L}_1(G_n,G_*)\to0$. Furthermore, as the expert function satisfies the strong identifiability condition in Definition~\ref{def:softmax_condition}, that is, the expert function and its derivatives are linearly independent, then all the coefficients in the representation of $[f_{G_n}(x)-f_{G_*}(x)]/\mathcal{L}_1(G_n,G_*)$ go to zero, contradicting to the result of Step 2. Hence, we obtain the local part~\eqref{eq:general_local_inequality}.

\vspace{0.5em}
\noindent
As a consequence, we can find a positive constant $\varepsilon'$ such that
\begin{align*}
    \inf_{G\in\mathcal{G}_{k}(\Theta):\mathcal{L}_1(G,G_*)\leq\varepsilon'}\normf{f_{G}-f_{G_*}}/\mathcal{L}_1(G,G_*)>0.
\end{align*}
Therefore, it suffices to prove the inequality
\begin{align}
    \label{eq:general_global_inequality}
     \inf_{G\in\mathcal{G}_{k}(\Theta):\mathcal{L}_1(G,G_*)>\varepsilon'}\normf{f_{G}-f_{G_*}}/\mathcal{L}_1(G,G_*)>0,
\end{align}
which we refer to as the global part of the inequality~\eqref{eq:activation_universal_inequality}. Putting the local part~\eqref{eq:general_local_inequality} and the global part~\eqref{eq:general_global_inequality} together, we achieve our desired inequality in equation~\eqref{eq:activation_universal_inequality}.

\vspace{0.5em}
\noindent
In the sequel, we will streamline the detailed proofs of the local part and the global part, respectively.

\vspace{0.5em}
\noindent
\textbf{Proof of the local part~\eqref{eq:general_local_inequality}.} Suppose that the local part~\eqref{eq:general_local_inequality} does not hold, then we can find a sequence of mixing measures $G_n=\sum_{i=1}^{k_*}\exp(\beta^n_{i})\delta_{(\omega^n_{i},\eta^n_i)}$ in $\mathcal{G}_{k}(\Theta)$ satisfying $\mathcal{L}_{1n}:=\mathcal{L}_1(G_n,G_*)\to0$ and
\begin{align}
    \label{eq:general_ratio_limit}
    \normf{f_{G_n}-f_{G_*}}/\mathcal{L}_{1n}\to0,
\end{align}
as $n\to\infty$. For ease of presentation, we denote $\mathcal{A}^n_j:=\mathcal{A}_j(G_n)$ as a Voronoi cell of $G_n$ generated by the $j$-th atom of $G_*$, for all $j\in[k_*]$. As we use asymptotic arguments throughout this proof, these Voronoi cells can be assumed to be independent of the sample size $n$, that is, $\mathcal{A}_j=\mathcal{A}^n_j$. Then, we can rewrite the Voronoi loss $\mathcal{L}_{1n}$ as
\begin{align}
    \label{eq:general_loss_proof}
   \mathcal{L}_{1n}:=\sum_{j=1}^{k_*}\Big|\sum_{i\in\mathcal{A}_j}\exp(\bzin)-\exp(\beta^*_{j})\Big|&+\sum_{j:|\mathcal{A}_j|>1}\sum_{i\in\mathcal{A}_j}\exp(\bzin)\Big[\|\dboijn\|^2+\|\deijn\|^2\Big]\nonumber\\
    &+\sum_{j:|\mathcal{A}_j|=1}\sum_{i\in\mathcal{A}_j}\exp(\beta^n_{i})\Big[\|\dboijn\|+\|\deijn\|\Big],
\end{align}
where we denote $\dboijn:=\omega^n_{i}-\omega^*_{j}$ and $\deijn:=\eta^n_i-\eta^*_j$. 

\vspace{0.5em}
\noindent
Recall that $\mathcal{L}_{1n}\to0$ as $n\to\infty$, then we have that $(\boin,\ein)\to(\boj,\ej)$ and $\sum_{i\in\mathcal{A}_j}\exp(\bzin)\to\exp(\bzj)$ for any $i\in\mathcal{A}_j$ and $j\in[k_*]$. Subsequently, we divide the proof of local part into three main steps. In the first step, we attempt to decompose the difference $f_{G_n}(x)-f_{G_*}(x)$ into a combination of linearly independent terms using Taylor expansions. Then, we demonstrate that not all the coefficients in the representation of $[f_{G_n}(x)-f_{G_*}(x)]/\mathcal{L}_{1n}$ converge to zero as $n\to\infty$ in the second step. In the final step, we employ the Fatou's lemma to show a contradiction that all the previous coefficients necessarily go to zero and, thus, complete the proof of the local part.

\vspace{0.5em}
\noindent
\textbf{Step 1 - Decompose the difference between regression functions.} To begin with, we use Taylor expansions to decompose the quantity $Q_n(x):=[\sum_{j=1}^{k_*}\exp((\omega^*_{j})^{\top}x+\beta^*_{j})]\cdot[f_{G_n}(x)-f_{G_*}(x)]$ into a combination of linearly independent terms. For this sake, we denote $F(x;\omega,\eta):=\exp(\omega^{\top}x)\mathcal{E}(x,\eta)$ and $H(x;\omega)=\exp(\omega^{\top}x)f_{G_n}(x)$. Then, $Q_n(x)$ can be decomposed as
\begin{align}
    \label{eq:general_Q_n}
    Q_n(x)&=\sum_{j=1}^{k_*}\sum_{i\in\mathcal{A}_j}\exp(\beta^n_{i})\Big[F(x;\boin,\ein)-F(x;\boj,\ej)\Big]\nonumber\\
    &-\sum_{j=1}^{k_*}\sum_{i\in\mathcal{A}_j}\exp(\beta^n_{i})\Big[H(x;\boin)-H(x;\boj)\Big]\nonumber\\
    &+\sum_{j=1}^{k_*}\Big(\sum_{i\in\mathcal{A}_j}\exp(\bzin)-\exp(\bzj)\Big)\Big[F(x;\boj,\ej)-H(x;\boj)\Big]\nonumber\\
    &:=A_n(x)-B_n(x)+C_{n}(x).
\end{align}
Next, we will decompose the terms $A_n(x)$ and $B_n(x)$, respectively. 

\vspace{0.5em}
\noindent
\textbf{Step 1A - Decompose $A_n(x)$.} We can represent $A_n(x)$ as
\begin{align*}
    A_n(x)&:=\sum_{j:|\mathcal{A}_j|=1}\sum_{i\in\mathcal{A}_j}\exp(\beta^n_{i})\Big[F(x;\boin,\ein)-F(x;\boj,\ej)\Big]\\
    &+\sum_{j:|\mathcal{A}_j|>1}\sum_{i\in\mathcal{A}_j}\exp(\beta^n_{i})\Big[F(x;\boin,\ein)-F(x;\boj,\ej)\Big]\\
    &:=A_{n,1}(x)+A_{n,2}(x).
\end{align*}
Regarding the term $A_{n,1}(x)$, by applying the first-order Taylor expansion to the function $F(x;\boin,\ein)$ around the point $(\boj,\ej)$, we have
\begin{align*}
    A_{n,1}(x)=\sum_{j:|\mathcal{A}_j|=1}\sum_{i\in\mathcal{A}_j}\exp(\beta^n_{i})\sum_{|\alpha|=1}(\dboijn)^{\alpha_1}(\deijn)^{\alpha_2}\cdot\frac{\partial F}{\partial\omega^{\alpha_1}\partial \eta^{\alpha_2}}(x;\boj,\ej)+R_1(x),
\end{align*}
where $R_1(x)$ is a Taylor remainder such that $R_1(x)/\mathcal{L}_{1n}\to0$ as $n\to\infty$. Note that the first derivatives of $F$ w.r.t its parameters $\omega$and $\eta$ are given by
\begin{align*}
    \frac{\partial F}{\partial\omega}(x;\boj,\ej)&=x\exp((\boj)^{\top}x)\mathcal{E}(x,\ej)=x\cdot F(x;\boj,\ej),\\
    \frac{\partial F}{\partial \eta}(x;\boj,\ej)&=\exp((\boj)^{\top}x)\cdot\frac{\partial \mathcal{E}}{\partial\eta}(x,\ej):=F_1(x;\boj,\ej).
\end{align*}
Then, $A_{n,1}(x)$ can be rewritten as
\begin{align}
    \label{eq:general_A_n_1}
    A_{n,1}(x)&=\sum_{j:|\mathcal{A}_j|=1}E_{n,1,j}(x)+R_1(x),
\end{align}
in which $E_{n,1,j}(x)=\sum_{i\in\mathcal{A}_j}\exp(\beta^n_{i})\Big[(\dboijn)^{\top}x\cdot F(x;\boj,\ej)+(\deijn)^{\top} F_1(x;\boj,\ej)\Big]$.

\vspace{0.5em}
\noindent
Analogously, by means of the second-order Taylor expansion, we can decompose $A_{n,2}(x)$ as
\begin{align*}
    A_{n,2}(x)=\sum_{j:|\mathcal{A}_j|>1}\sum_{i\in\mathcal{A}_j}\exp(\beta^n_{i})\sum_{|\alpha|=1}^{2}\frac{1}{\alpha!}(\dboijn)^{\alpha_1}(\deijn)^{\alpha_2}\cdot\frac{\partial^{|\alpha_1|+|\alpha_2|} F}{\partial\omega^{\alpha_1}\partial \eta^{\alpha_2}}(x;\boj,\ej)+R_2(x),
\end{align*}
where $R_2(x)$ is a Taylor remainder such that $R_2(x)/\mathcal{L}_{1n}\to0$ as $n\to\infty$. By taking the second derivatives of $F$ w.r.t its parameters, we have
\begin{align*}
    \frac{\partial^2F}{\partial\omega\partial\omega^{\top}}(x;\boj,\ej)&=xx^{\top}\cdot F(x;\boj,\ej),\quad \frac{\partial^2F}{\partial\omega\partial\eta^{\top}}(x;\boj,\ej)=x\cdot [F_1(x;\boj,\ej)]^{\top},\\
    \frac{\partial^2F}{\partial\eta\partial\eta^{\top}}(x;\boj,\ej)&=\exp((\boj)^{\top}x)\cdot\frac{\partial^2 \mathcal{E}}{\partial\eta\partial\eta^{\top}}:=F_2(x;\boj,\ej)
\end{align*}
Then, we can represent $A_{n,2}(x)$ as
\begin{align}
     \label{eq:general_A_n_2}
    A_{n,2}(x)=\sum_{j:|\mathcal{A}_j|>1}[E_{n,1,j}(x)+E_{n,2,j}(x)]+R_2(x),
\end{align}
where 
\begin{align*}
    &E_{n,2,j}(x):=\sum_{i\in\mathcal{A}_j}\exp(\bzin)\Bigg\{\Big[x^{\top}\Big(M_{d}\odot (\dboijn)(\dboijn)^{\top}\Big)x\Big]\cdot F(x;\boj,\aj,\bj)\\
    &+\Big[x^{\top}(\dboijn)(\deijn)^{\top}F_1(x;\boj,\ej)\Big]+\Big[(\deijn)^{\top}\Big(M_{d}\odot F_2(x;\boj,\ej)\Big)(\deijn)\Big]\Bigg\},
\end{align*}
with $M_d$ being an $d\times d$ matrix whose diagonal entries are $\frac{1}{2}$ while other entries are 1.

\vspace{0.5em}
\noindent
\textbf{Step 1B - Decompose $B_n(x)$.} Next, we apply the same strategy of decomposing $A_n(x)$ for partitioning the term $B_n(x)$. In particular, we first have
\begin{align*}
    B_n(x)&=\sum_{j:|\mathcal{A}_j|=1}\sum_{i\in\mathcal{A}_j}\exp(\beta^n_{i})\Big[H(x;\boin)-H(x;\boj)\Big]\\
    &+\sum_{j:|\mathcal{A}_j|>1}\sum_{i\in\mathcal{A}_j}\exp(\beta^n_{i})\Big[H(x;\boin)-H(x;\boj)\Big]\\
    &:=B_{n,1}(x)+B_{n,2}(x).
\end{align*}
Regarding the term $B_{n,1}(x)$, by employing the first-order Taylor expansion to the function $H(x;\boin)$ around the point $\boj$, we have
\begin{align}
    \label{eq:general_B_n_1}
    B_{n,1}(x)=\sum_{j:|\mathcal{A}_j|=1}\sum_{i\in\mathcal{A}_j}\exp(\bzin)(\dboijn)^{\top}x\cdot H(x;\boj)+R_3(x),
\end{align}
where $R_3(x)$ is a Taylor remainder such that $R_3(x)/\mathcal{L}_{1n}\to0$ as $n\to\infty$. Similarly, by means of the second-order Taylor expansion, we get
\begin{align}
      \label{eq:general_B_n_2}
    B_{n,2}(x)=\sum_{j:|\mathcal{A}_j|>1}\sum_{i\in\mathcal{A}_j}\exp(\bzin)\Big[(\dboijn)^{\top}x+(\dboijn)^{\top}\Big(M_{d}\odot xx^{\top}\Big)(\dboijn)\Big]\cdot H(x;\boj) + R_4(x),
\end{align}
where $R_4(x)$ is a Taylor remainder such that $R_4(x)/\mathcal{L}_{1n}\to0$ as $n\to\infty$.\\

\noindent
Putting the above results together, we see that $[A_n(x)-R_1(x)-R_2(x)]/\mathcal{L}_{1n}$, $[B_n(x)-R_3(x)-R_4(x)]/\mathcal{L}_{1n}$ and $C_{n}(x)/\mathcal{L}_{1n}$ can be written as a combination of elements from the following set
\begin{align*}
    &\Big\{F(x;\boj,\ej), \ x^{(u)}F(x;\boj,\ej), \ x^{(u)}x^{(v)}F(x;\boj,\ej):u,v\in[d], \ j\in[k_*]\Big\},\\
    \cup&~\Big\{[F_1(x;\boj,\ej)]^{(u)}, \ x^{(u)}[F_1(x;\boj,\ej)]^{(v)}:u,v\in[d], \ j\in[k_*]\Big\},\\
    \cup&~\Big\{[F_2(x;\boj,\ej)]^{(uv)}:u,v\in[d], \ j\in[k_*]\Big\},\\
    \cup&~\Big\{H(x;\boj), \ x^{(u)}H(x;\boj), \ x^{(u)}x^{(v)}H(x;\boj):u,v\in[d], \ j\in[k_*]\Big\}.
\end{align*}
\textbf{Step 2 - Non-vanishing coefficients.} Moving to this step, we utilize the proof by contradiction method to show that not all coefficients in the representations of $[A_n-R_1(x)-R_2(x)]/\mathcal{L}_{1n}$, $[B_n-R_3(x)-R_4(x)]/\mathcal{L}_{1n}$ and $C_{n}(x)/\mathcal{L}_{1n}$ converge to zero as $n\to\infty$. Indeed, assume by contrary that all of them go to zero and consider the coefficients of the term
\begin{itemize}
    \item $F(x;\boj,\ej)$ for $j\in[k_*]$, we have $$\frac{1}{\mathcal{L}_{1n}}\cdot\sum_{j=1}^{k_*}\Big|\sum_{i\in\mathcal{A}_j}\exp(\bzin)-\exp(\bzj)\Big|\to0;$$
    \item $x^{(u)}F(x;\boj,\ej)$ for $u\in[d]$ and $j:|\mathcal{A}_j|=1$, we have $$\frac{1}{\mathcal{L}_{1n}}\cdot\sum_{j:|\mathcal{A}_j|=1}\sum_{i\in\mathcal{A}_j}\exp(\bzin)\|\dboijn\|_1\to0.$$
    Since the $\ell_1$-norm is equivalent to the $\ell_2$-norm, we deduce
    $$\frac{1}{\mathcal{L}_{1n}}\cdot\sum_{j:|\mathcal{A}_j|=1}\sum_{i\in\mathcal{A}_j}\exp(\bzin)\|\dboijn\|\to0;$$
    \item $[F_1(x;\boj,\ej)]^{(u)}$ for $u\in[d]$ and $j:|\mathcal{A}_j|=1$, we have $$\frac{1}{\mathcal{L}_{1n}}\cdot\sum_{j:|\mathcal{A}_j|=1}\sum_{i\in\mathcal{A}_j}\exp(\bzin)\|\deijn\|\to0;$$
    \item $[x^{(u)}]^2F(x;\boj,\ej)$ for $u\in[d]$ and $j:|\mathcal{A}_j|>1$, we have $$\frac{1}{\mathcal{L}_{1n}}\cdot\sum_{j:|\mathcal{A}_j|>1}\sum_{i\in\mathcal{A}_j}\exp(\bzin)\|\dboijn\|^2\to0:$$
    \item $[F_2(x;\boj,\ej)]^{(uu)}$ for $u\in[d]$ and $j:|\mathcal{A}_j|>1$, we have $$\frac{1}{\mathcal{L}_{1n}}\cdot\sum_{j:|\mathcal{A}_j|>1}\sum_{i\in\mathcal{A}_j}\exp(\bzin)\|\deijn\|^2\to0;$$
\end{itemize}
Combine all the above limits, we arrive at $1=\mathcal{L}_{1n}/\mathcal{L}_{1n}\to0$ as $n\to\infty$, which is a contradiction. Thus, it follows that at least one among the coefficients in the representations of $[A_n(x)-R_1(x)-R_2(x)]/\mathcal{L}_{1n}$, $[B_n(x)-R_3(x)-R_4(x)]/\mathcal{L}_{1n}$ and $C_{n}(x)/\mathcal{L}_{1n}$ does not converge to zero.

\vspace{0.5em}
\noindent
\textbf{Step 3 - Application of the Fatou's lemma.} Now, we attempt to exhibit a contradiction to the conclusion of Step 2. Let $m_n$ be the maximum of the absolute values of the coefficients in the representations of $[A_n(x)-R_1(x)-R_2(x)]/\mathcal{L}_{1n}$, $[B_n(x)-R_3(x)-R_4(x)]/\mathcal{L}_{1n}$ and $C_{n}(x)/\mathcal{L}_{1n}$. As at least one among those coefficients does not go zero, we have $1/m_n\not\to\infty$. 

\vspace{0.5em}
\noindent
Recall from the hypothesis in equation~\eqref{eq:general_ratio_limit} that we have $\normf{f_{G_n}-f_{G_*}}/\mathcal{L}_{1n}\to0$ as $n\to\infty$. This result also implies that $\|f_{G_n}-f_{G_*}\|_{L^1(\mu)}/\mathcal{L}_{1n}\to0$. Then, by applying the Fatou's lemma, we get
\begin{align*}
    0=\lim_{n\to\infty}\frac{\|f_{G_n}-f_{G_*}\|_{L^1(\mu)}}{m_n\mathcal{L}_{1n}}\geq \int \liminf_{n\to\infty}\frac{|f_{G_n}(x)-f_{G_*}(x)|}{m_n\mathcal{L}_{1n}}\dint\mu(x)\geq 0.
\end{align*}
As a result, we have $[f_{G_n}(x)-f_{G_*}(x)]/[m_n\mathcal{L}_{1n}]\to0$ for almost every $x$. Since the parameter $\Theta$ is compact, the term $\sum_{j=1}^{k_*}\exp((\omega^*_{j})^{\top}x+\beta^*_{j})$ is bounded. Thus, it follows that $Q_n(x)/[m_n\mathcal{L}_{1n}]\to0$ as $n\to\infty$, or equivalently, 
\begin{align}
    \label{eq:general_zero_limit}
    \frac{1}{m_n\mathcal{L}_{1n}}\cdot\Big[(A_{n,1}(x)-R_1(x)+A_{n,2}(x)-R_2(x))-(B_{n,1}(x)-R_3(x)+&B_{n,2}(x)-R_4(x))+C_{n}(x)\Big]\to0.
\end{align}
For ease of presentation, we denote
\begin{align*}
    \frac{1}{m_n\mathcal{L}_{1n}}\cdot\sum_{i\in\mathcal{A}_j}\exp(\bzin)(\dboijn)\to\phi_{1,j}&,\quad  \frac{1}{m_n\mathcal{L}_{1n}}\cdot\sum_{i\in\mathcal{A}_j}\exp(\bzin)(\dboijn)(\dboijn)^{\top}\to\phi_{2,j},\\
    \frac{1}{m_n\mathcal{L}_{1n}}\cdot\sum_{i\in\mathcal{A}_j}\exp(\bzin)(\deijn)\to\varphi_{1,j}&,\quad \frac{1}{m_n\mathcal{L}_{1n}}\cdot\sum_{i\in\mathcal{A}_j}\exp(\bzin)(\deijn)(\deijn)^{\top}\to\varphi_{2,j},\\
    \frac{1}{m_n\mathcal{L}_{1n}}\cdot\sum_{i\in\mathcal{A}_j}\exp(\bzin)(\dboijn)(\deijn)^{\top}\to\zeta_{j}&, \quad \frac{1}{m_n\mathcal{L}_{1n}}\cdot\Big(\sum_{i\in\mathcal{A}_j}\exp(\bzin)-\exp(\bzj)\Big)\to\xi_j,\\
    F_{\rho j}:=F_{\rho}(x;\boj,\ej)&, \quad H_j=H(x;\boj).
\end{align*}
Above, at least one among the limits $\phi^{(u)}_{1,j}$, $\phi^{(uu)}_{2,j}$, $\varphi^{(u)}_{1,j}$, $\varphi^{(uu)}_{2,j}$ and $\xi_j$, for $j\in[k_*]$, is non-zero as a result of the conclusion in Step 2. From the formulation of 
\begin{itemize}
    \item $A_{n,1}(x)$ in equation~\eqref{eq:general_A_n_1}, we have 
    \begin{align}
        \label{eq:general_limit_1}
        \frac{A_{n,1}(x)-R_1(x)}{m_n\mathcal{L}_{1n}}\to\sum_{j:|\mathcal{A}_j|=1}\Big[\phi^{\top}_{1,j}x\cdot F_j+\varphi_{1,j}^{\top}F_{1j}\Big].
    \end{align}
    \item $A_{n,2}(x)$ in equation~\eqref{eq:general_A_n_2}, we have
    \begin{align}
    \label{eq:general_limit_2}
        \frac{A_{n,2}(x)-R_2(x)}{m_n\mathcal{L}_{2n}}\to\sum_{j:|\mathcal{A}_j|>1}\Bigg\{\Big[\phi^{\top}_{1,j}x+x^{\top}\Big(M_{d}\odot \phi_{2,j}\Big)x\Big]\cdot F_j+[\varphi_{1,j}^{\top}+x^{\top}\zeta_{j}]\cdot F_{1j}\nonumber\\
        +\Big[M_{d}\odot \varphi_{2,j}\Big]\odot F_{2j}\Bigg\}.
    \end{align}
    \item $B_{n,1}(x)$ in equation~\eqref{eq:general_B_n_1}, we have
    \begin{align}
        \label{eq:general_limit_3}
        \frac{B_{n,1}(x)-R_3(x)}{m_n\mathcal{L}_{2n}}\to\sum_{j:|\mathcal{A}_j|=1}[\phi^{\top}_{1,j}x\cdot H_j].
    \end{align}
    \item $B_{n,2}(x)$ in equation~\eqref{eq:general_B_n_2}, we have
    \begin{align}
        \label{eq:general_limit_4}
        \frac{B_{n,2}(x)-R_4(x)}{m_n\mathcal{L}_{2n}}\to\sum_{j:|\mathcal{A}_j|>1}\Big[\phi^{\top}_{1,j}x+x^{\top}\Big(M_{d}\odot \phi_{2,j}\Big)x\Big]\cdot H_j.
    \end{align}
    \item $C_{n}(x)$ in equation~\eqref{eq:general_Q_n}, we have
    \begin{align}
        \label{eq:general_limit_5}
        \frac{C_{n}(x)}{m_n\mathcal{L}_{2n}}\to\sum_{j=1}^{k_*}\xi_j[F_j-H_j].
    \end{align}
\end{itemize}
Note that from equation~\eqref{eq:general_zero_limit}, it follows that the sum of the limits in equations~\eqref{eq:general_limit_1}, \eqref{eq:general_limit_2}, \eqref{eq:general_limit_3}, \eqref{eq:general_limit_4} and \eqref{eq:general_limit_5} is equal to zero, that is,
\begin{align}
    \label{eq:sum_limit_zero}
    &\sum_{j:|\mathcal{A}_j|=1}\Big[\phi^{\top}_{1,j}x\cdot F_j+\varphi_{1,j}^{\top}F_{1j}\Big]+\sum_{j:|\mathcal{A}_j|>1}\Bigg\{\Big[\phi^{\top}_{1,j}x+x^{\top}\Big(M_{d}\odot \phi_{2,j}\Big)x\Big]\cdot F_j+[\varphi_{1,j}^{\top}+x^{\top}\zeta_{j}]\cdot F_{1j}\nonumber\\
    &+\Big[M_{d}\odot \varphi_{2,j}\Big]\odot F_{2j}\Bigg\}-\sum_{j:|\mathcal{A}_j|=1}[\phi^{\top}_{1,j}x\cdot H_j]-\sum_{j:|\mathcal{A}_j|>1}\Big[\phi^{\top}_{1,j}x+x^{\top}\Big(M_{d}\odot \phi_{2,j}\Big)x\Big]\cdot H_j\nonumber\\
    &\hspace{10cm}+\sum_{j=1}^{k_*}\xi_j[F_j-H_j]=0.
\end{align}
\noindent
Subsequently, we aim to demonstrate that the values of $\phi^{(u)}_{1,j}$, $\phi^{(uu)}_{2,j}$, $\varphi^{(u)}_{1,j}$, $\varphi^{(uu)}_{2,j}$ and $\xi_j$ are all zero for all $j\in[k_*]$. Indeed, let $P_1,P_2,\ldots,P_{\ell}$ be the partition of the set $\{\exp((\boj)^{\top}x):j\in[k_*]\}$, for some $\ell\in[k_*]$, such that 
\begin{itemize}
    \item[(i)] $\omega^*_{j}=\omega^*_{j'}$ for any $j,j'\in P_i$ and $i\in[\ell]$;
    \item[(ii)] $\omega^*_{j}\neq\omega^*_{j'}$ when $j$ and $j'$ do not belong to the same set $P_i$ for any $i\in[\ell]$.
\end{itemize}
Then, the set $\{\exp((\omega^*_{j_1})^{\top}x),\ldots,\exp((\omega^*_{j_\ell})^{\top}x)\}$, where $j_i\in P_i$, is linearly independent. This result together with equation~\eqref{eq:sum_limit_zero} implies that for any $i\in[\ell]$, we have
\begin{align*}
    &\sum_{j\in P_i:|\mathcal{A}_j|=1}\Big[(\xi_j+\phi^{\top}_{1,j}x)\cdot \mathcal{E}_j+\varphi_{1,j}^{\top}\mathcal{E}_{1j}\Big]+\sum_{j\in P_i:|\mathcal{A}_j|>1}\Bigg\{\Big[\phi^{\top}_{1,j}x+x^{\top}\Big(M_{d}\odot \phi_{2,j}\Big)x\Big]\cdot \mathcal{E}_j\\
    &+[\varphi_{1,j}^{\top}+x^{\top}\zeta_{j}]\cdot \mathcal{E}_{1j}+\Big[M_{d}\odot \varphi_{2,j}\Big]\odot \mathcal{E}_{2j}\Bigg\}-\sum_{j\in P_i:|\mathcal{A}_j|=1}[(\phi^{\top}_{1,j}x+\xi_j)\cdot f_{G_*}(x)]\\
    &\hspace{4cm}-\sum_{j\in P_i:|\mathcal{A}_j|>1}\Big[\xi_j+\phi^{\top}_{1,j}x+x^{\top}\Big(M_{d}\odot \phi_{2,j}\Big)x\Big]\cdot f_{G_*}(x)=0,
\end{align*}
where we denote $\mathcal{E}_{j}:=\mathcal{E}(x,\ej)$, $\mathcal{E}_{1j}:=\frac{\partial \mathcal{E}}{\partial \eta}(x,\ej)$ and $\mathcal{E}_{2j}:=\frac{\partial^2 \mathcal{E}}{\partial\eta\partial\eta^{\top}}(x,\ej)$. Furthermore, since the expert function $x\mapsto\mathcal{E}(x,\eta)$ satisfies the strong identifiability condition in Definition~\ref{def:softmax_condition}, then the set 
\begin{align*}
        &\Big\{x^{\nu}\cdot\frac{\partial^{|\rho|}\mathcal{E}}{\partial\eta^{\rho}}(x,\eta^*_j): j\in[k_*], \nu\in\mathbb{N}^d, \ \rho\in\mathbb{N}^{q}, 0\leq |\nu|+|\rho|\leq 2\Big\}
    \end{align*}
is linearly independent for almost every $x$. Thus, it follows that $\xi_j=0$, $\phi_{1,j}=\varphi_{1,j}=\zerod$ and $\phi_{2,j}=\varphi_{2,j}=\zeta_{j}={0}_{d\times d}$ for any $j\in P_i$ and $i\in[\ell]$. This contradicts the fact that not all the values of $\phi^{(u)}_{1,j}$, $\phi^{(uu)}_{2,j}$, $\varphi^{(u)}_{1,j}$, $\varphi^{(uu)}_{2,j}$ and $\xi_j$, for $j\in[k_*]$, are zero. Therefore, we obtain the local part~\eqref{eq:general_local_inequality}, that is,
\begin{align*}
    \lim_{\varepsilon\to0}\inf_{G\in\mathcal{G}_{k}(\Theta):\mathcal{L}_1(G,G_*)\leq\varepsilon}\normf{f_{G}-f_{G_*}}/\mathcal{L}_1(G,G_*)>0.
\end{align*}
Consequently, we can find a positive constant $\varepsilon'$ such that
\begin{align*}
    \inf_{G\in\mathcal{G}_{k}(\Theta):\mathcal{L}_1(G,G_*)\leq\varepsilon'}\normf{f_{G}-f_{G_*}}/\mathcal{L}_1(G,G_*)>0.
\end{align*}
\textbf{Proof of the global part~\eqref{eq:general_global_inequality}:} Given the above result, it is sufficient to prove that 
\begin{align*}
     \inf_{G\in\mathcal{G}_{k}(\Theta):\mathcal{L}_1(G,G_*)>\varepsilon'}\normf{f_{G}-f_{G_*}}/\mathcal{L}_1(G,G_*)>0.
\end{align*}
Suppose that the global part does not hold, then there exists a sequence of mixing measures $G'_n\in\mathcal{G}_{k}(\Theta)$ satisfying $\mathcal{L}_1(G'_n,G_*)>\varepsilon'$ and
\begin{align*}
    \lim_{n\to\infty}\frac{\normf{f_{G'_n}-f_{G_*}}}{\mathcal{L}_1(G'_n,G_*)}=0,
\end{align*}
indicating that $\normf{f_{G'_n}-f_{G_*}}\to0$ as $n\to\infty$. Since the parameter space $\Theta$ is compact, we can substitute the sequence $(G'_n)$ with one of its subsequences that converges to $G'\in\mathcal{G}_{k}(\Theta)$. In addition, as we have $\mathcal{L}_1(G'_n,G_*)>\varepsilon'$, it follows that $\mathcal{L}_1(G',G_*)>\varepsilon'$. 
Now, by applying the Fatou's lemma, we get
\begin{align*}
    0=\lim_{n\to\infty}\normf{f_{G'_n}-f_{G_*}}^2\geq \int\liminf_{n\to\infty}\Big|f_{G'_n}(x)-f_{G_*}(x)\Big|^2~\dint\mu(x).
\end{align*}
The above result implies that $f_{G'}(x)=f_{G_*}(x)$ for almost every $x$. From Proposition~\ref{prop:general_identifiability} in Appendix~\ref{appendix:identifiability}, we obtain $G'\equiv G_*$. As a consequence, we have $\mathcal{L}_1(G',G_*)=0$, which contradicts the fact that $\mathcal{L}_1(G',G_*)>\varepsilon'>0$. 
Hence, we achieve the global part and complete the proof of Theorem~\ref{theorem:general_experts}.

\section{Proofs of other Theorems}
\label{appendix:softmax_MoE}

\subsection{Proof of Theorem~\ref{theorem:linear_experts}}
\label{appendix:linear_experts}
\noindent
\textbf{Proof overview.} In this proof, we first demonstrate that the limit
\begin{align}
    \label{eq:ratio_zero_limit}
    \lim_{\varepsilon\to0}\inf_{G\in\mathcal{G}_{k}(\Theta):\mathcal{L}_{2,r}(G,G_*)\leq\varepsilon}\normf{f_{G}-f_{G_*}}/\mathcal{L}_{2,r}(G,G_*)=0
\end{align}
holds for any $r\geq 1$. Given the above result, we proceed to derive the minimax lower bound in Theorem~\ref{theorem:linear_experts}:
\begin{align}
    \label{eq:minimax_activation_experts}
    \inf_{\overline{G}_n\in\mathcal{G}_{k}(\Theta)}\sup_{G\in\mathcal{G}_{k}(\Theta)\setminus\mathcal{G}_{k_*-1}(\Theta)}\bbE_{f_{G}}[\mathcal{L}_{2,r}(\overline{G}_n,G)]\gtrsim n^{-1/2}.
\end{align}
\textbf{Proof of equation~\eqref{eq:ratio_zero_limit}:} It suffices to construct a mixing measure sequence $(G_n)$ satisfying $\mathcal{L}_{2,r}(G_n,G_*)\to0$ and 
\begin{align}
    \label{eq:ratio_zero_linear}
    \normf{f_{G_n}-f_{G_*}}/\mathcal{L}_{2,r}(G_n,G_*)\to0,
\end{align}
as $n\to\infty$. Now, let us take into account the sequence $(G_n)$ defined as  $G_n:=\sum_{i=1}^{k_*+1}\exp(\bzin)\delta_{(\boin,\ain,\bin)}$, where we set 

\vspace{0.5em}
$\exp(\beta^n_{1})=\exp(\beta^n_{2})=\frac{1}{2}\exp(\beta^*_{1})+\frac{1}{2n^{r+1}}$ and  $\exp(\beta^n_{i})=\exp(\beta^n_{(i-1)})$ for any $3\leq i\leq k_*+1$;

\vspace{0.5em}
$\omega^n_{1}=\omega^n_{2}=\omega^*_{1}$ and  $\omega^n_{i}=\omega^n_{(i-1)}$ for any $3\leq i\leq k_*+1$;

\vspace{0.5em}
$a^n_1=a^n_2=a^*_1$ and $a^n_i=a^n_{i-1}$ for any $3\leq i\leq k_*+1$;

\vspace{0.5em}
$b^n_1=b^*_1+\frac{1}{n}$, $b^n_2=b^*_1-\frac{1}{n}$ and $b^n_{i}=b^*_{i-1}$ for any $3\leq i\leq k_*+1$.

\vspace{0.5em}
\noindent
Then, the Voronoi loss $\mathcal{L}_{2,r}(G_n,G_*)$ can be rewritten as
\begin{align}
    \label{eq:D_r_formulation}
    \mathcal{L}_{2,r}(G_n,G_*)=\frac{1}{n^{r+1}}+\Big[\exp(\beta^*_{1})+\frac{1}{n^{r+1}}\Big]\cdot\frac{1}{n^r}=\mathcal{O}(n^{-r}).
\end{align}
The above formulation implies that $\mathcal{L}_{2,r}(G_n,G_*)\to0$ as $n\to\infty$. Thus, it suffices to derive equation~\eqref{eq:ratio_zero_linear}. To this end, we decompose the quantity $Q_n(x):=[\sum_{j=1}^{k_*}\exp((\omega^*_{j})^{\top}x+\beta^*_{j})]\cdot[f_{G_n}(x)-f_{G_*}(x)]$ as
\begin{align*}
    Q_n(x)&=\sum_{j=1}^{k_*}\sum_{i\in\mathcal{A}_j}\exp(\beta^n_{i})\Big[\exp((\boin)^{\top}x)((\ain)^{\top}x+\bin)-\exp((\boj)^{\top}x)((\aj)^{\top}x+\bj)\Big]\\
    &-\sum_{j=1}^{k_*}\sum_{i\in\mathcal{A}_j}\exp(\beta^n_{i})\Big[\exp((\boin)^{\top}x)-\exp((\boj)^{\top}x)\Big]f_{G_n}(x)\\
    &+\sum_{j=1}^{k_*}\Big(\sum_{i\in\mathcal{A}_j}\exp(\bzin)-\exp(\bzj)\Big)\Big[\exp((\boj)^{\top}x)((\aj)^{\top}x+\bj)-\exp((\boj)^{\top}x)f_{G_n}(x)\Big]\\
    &:=A_n(x)-B_n(x)+C_{n}(x).
\end{align*}
By plugging the values of $\exp(\beta^n_i)$, $\omega^n_{i},a^n_i$ and $b^n_i$ into the terms $A_n(x)$, $B_n(x)$ and $C_n(x)$, we have
\begin{align*}
    A_n(x)&=\sum_{i=1}^{2}\exp(\beta^n_{i})\exp((\omega^*_{1})^{\top}x)(b^n_i-b^*_1)=\frac{1}{2}\Big[\exp(\beta^*_{1})+\frac{1}{n^{r+1}}\Big]\exp((\omega^*_{1})^{\top}x)[(b^n_1-b^*_1)+(b^n_2-b^*_1)]=0,\\
    B_{n}(x)&=\sum_{i=1}^{2}\exp(\beta^n_i)[\exp((\boin)^{\top}x)-\exp((\omega^*_1)^{\top}x)]f_{G_n}(x)=0,\\
    C_{n}(x)&=\Big(\sum_{i=1}^{2}\exp(\bzin)-\exp(\beta^*_{1})\Big)\Big[\exp((\beta^*_1)^{\top}x)((a^*_1)^{\top}x+b^*_1)-\exp((\beta^*_1)^{\top}x)f_{G_n}(x)\Big]=\mathcal{O}(n^{-(r+1)})
\end{align*}
Since $\mathcal{L}_{2,r}(G_n,G_*)=\mathcal{O}(n^{-r})$, we can justify that $C_{n}(x)/\mathcal{L}_{2,r}(G_n,G_*)\to0$, thereby leading to $Q_n(x)/\mathcal{L}_{2,r}(G_n,G_*)\to0$ as $n\to\infty$ for almost every $x$. Note that the term $\sum_{j=1}^{k_*}\exp((\boj)^{\top}x+\bzj)$ is bounded, then we also have that $[f_{G_n}(x)-f_{G_*}(x)]/\mathcal{L}_{2,r}\to0$ for almost every $x$. As a consequence, we obtain $\normf{f_{G_n}-f_{G_*}}/\mathcal{L}_{2,r}\to0$ as $n\to\infty$. Hence, the proof of claim~\eqref{eq:ratio_zero_limit} is completed.\\

\noindent
\textbf{Proof of equation~\eqref{eq:minimax_activation_experts}:}
Since the noise variables $\varepsilon_i$ given the input $X_i$ follows Gaussian distributions, we have $Y_{i}|X_{i} \sim \mathcal{N}(f_{G_{*}}(x_{i}), \nu)$ for all $i \in [n]$. Next, it follows from equation~\eqref{eq:ratio_zero_limit} that for a small enough constant $\varepsilon>0$ and a fixed constant $c>0$ that will be determined later, there exists a mixing measure $G'_* \in \mathcal{G}_{k}(\Theta)$ satisfying $\mathcal{L}_{2,r}(G'_*,G_*)=2 \varepsilon$ and $\|f_{G'_*} - f_{G_*}\|_{L^2(\mu)} \leq c\cdot\varepsilon$. According to Le Cam's lemma~\cite{yu97lecam}, since the Voronoi loss $\mathcal{L}_{2,r}$ satisfies the weak triangle inequality, we have
\begin{align}
    &\inf_{\overline{G}_n\in\mathcal{G}_{k}(\Theta)}\sup_{G\in\mathcal{G}_{k}(\Theta)\setminus\mathcal{G}_{k_*-1}(\Theta)}\bbE_{f_{G}}[\mathcal{L}_{2,r}(\overline{G}_n,G)]\nonumber\\ 
    & \gtrsim \frac{\mathcal{L}_{2,r}(G'_*,G_*)}{8} \text{exp}(- n \mathbb{E}_{X \sim \mu}[\text{KL}(\mathcal{N}(f_{G'_{*}}(x), \nu),\mathcal{N}(f_{G_{*}}(x), \nu))]) \nonumber \\
    & \gtrsim \varepsilon \cdot \text{exp}(-n \|f_{G'_*} - f_{G_*}\|_{L^2(\mu)}^2)\nonumber \\
    & \gtrsim \varepsilon \cdot \text{exp}(-c n \varepsilon^2), \label{eq:LeCam_inequality}
\end{align}
where the second inequality is due to the fact that $\text{KL}(\mathcal{N}(f_{G'_{*}}(x), \nu),\mathcal{N}(f_{G_{*}}(x), \nu)) = \frac{[f_{G'_*}(x) - f_{G_*}(x)]^2}{2 \nu}$.

\vspace{0.5em}
\noindent
By setting $\varepsilon=n^{-1/2}$, we get $\varepsilon\cdot\exp(-c n\varepsilon^2)=n^{-1/2}\exp(-c)$. Hence, we obtain the desired minimax lower bound in equation~\eqref{eq:minimax_activation_experts} and complete the proof.

\subsection{Proof of Theorem~\ref{theorem:linear_dense-to-sparse_experts}}
\label{appendix:linear_dense-to-sparse_experts}
\noindent
In this proof, we will focus on showing that the limit
\begin{align}
    \label{eq:ratio_zero_dense-to-sparse}
        \lim_{\varepsilon\to0}\inf_{G\in\mathcal{G}_{k}(\Theta):\mathcal{L}_{3,r}(G,G_*)\leq\varepsilon}\frac{\normf{g_{G}-g_{G_*}}}{\mathcal{L}_{3,r}(G,G_*)}=0,
    \end{align}
holds for any $r\geq 1$. Then, by employing the arguments for proving equation~\eqref{eq:minimax_activation_experts} in Appendix~\ref{appendix:linear_experts}, we achieve the minimax lower bound in Theorem~\ref{theorem:linear_dense-to-sparse_experts}, that is,
\begin{align*}
        \inf_{\overline{G}_n\in\mathcal{G}_{k}(\Theta)}\sup_{G\in\mathcal{G}_{k}(\Theta)\setminus\mathcal{G}_{k_*-1}(\Theta)}\bbE_{g_{G}}[\mathcal{L}_{3,r}(\overline{G}_n,G)]\gtrsim n^{-1/2}.
    \end{align*}
\textbf{Proof of equation~\eqref{eq:ratio_zero_dense-to-sparse}:} It is sufficient to build a mixing measure sequence $(G_n)$ that satisfies $\mathcal{L}_{3,r}(G_n,G_*)\to0$ and
\begin{align}
    \label{eq:ratio_zero_over}
   \normf{g_{G}-g_{G_*}}/\mathcal{L}_{3,r}(G_n,G_*)\to0,
\end{align}
as $n\to\infty$. For this sake, let us consider the sequence defined as $G_n:=\sum_{i=1}^{k_*}\exp\Big(\frac{\bzin}{\tau^n}\Big)\delta_{(\boin,\tau^n,\ein)}$, where we define for any $j\in[k_*]$ that

\vspace{0.5em}
$\eta^n_{i}=\eta^*_{j}$, for any $i\in\mathcal{A}_j$;

\vspace{0.5em}
$\omega^n_{i}=\omega^*_{j}+s_{n,j}$, for any $i\in\mathcal{A}_j$;

\vspace{0.5em}
$\tau^n=\tau^*+t_n$;

\vspace{0.5em}
$\bzin=\tau^n\cdot\Big[\frac{\bzj}{\tau^*}-\log(|\mathcal{A}_j|)\Big]$, which implies that $\sum_{i\in\mathcal{A}_j}\exp\Big(\frac{\bzin}{\tau^n}\Big)=\exp\Big(\frac{\bzj}{\tau^*}\Big)$, for any $i\in\mathcal{A}_j$,

\vspace{0.5em}
\noindent
where $s_{n,j}:=(s^{(1)}_{n,j},\ldots,s^{(d)}_{n,j})\in\mathbb{R}^d$ and $t_n\in\mathbb{R}$ will be determined later such that $s^{(u)}_{n,j}\to0$ and $t_n\to0$ as $n\to\infty$ for any $u\in[d]$ and $j\in[k_*]$. Then, we can represent the Voronoi loss $\mathcal{L}_{3,r}(G_n,G_*)$ as
\begin{align*}
    \mathcal{L}_{3,r}(G_n,G_*)=\sum_{j=1}^{k_*}|\mathcal{A}_j|\cdot\exp\Big(\frac{\bzj}{\tau^*}\Big)(\|s_{n,j}\|^r+t_n^r).
\end{align*}
Due to the aforementioned properties of sequences $(s_{n,j})$ and $(t_n)$, it follows that $\mathcal{L}_{3,r}(G_n,G_*)\to0$ as $n\to\infty$. Thus, we can complete the proof by deriving equation~\eqref{eq:ratio_zero_over}.

\vspace{0.5em}
\noindent
To achieve this goal, we take into account the quantity $Q_n(x):=\Big[\sum_{j=1}^{k_*}\exp\Big(\dfrac{(\boi)^{\top}x+\bzi}{\tau^*}\Big)\Big]\cdot\Big[g_{G_n}(x)-g_{G_*}(x)\Big]$, which can be decomposed as
\begin{align*}
    Q_n(x)&=\sum_{j=1}^{k_*}\sum_{i\in\mathcal{A}_j}\exp\Big(\frac{\bzin}{\tau^n}\Big)\Big[\exp\Big(\frac{(\boin)^{\top}x}{\tau^n}\Big)\mathcal{E}(x,\ein)-\exp\Big(\frac{(\boj)^{\top}x}{\tau^*}\Big)\mathcal{E}(x,\ej)\Big]\\
    &-\sum_{j=1}^{k_*}\sum_{i\in\mathcal{A}_j}\exp\Big(\frac{\bzin}{\tau^n}\Big)\Big[\exp\Big(\frac{(\boin)^{\top}x}{\tau^n}\Big)g_{G_n}(x)-\exp\Big(\frac{(\boj)^{\top}x}{\tau^*}\Big)g_{G_n}(x)\Big]\\
    &+\sum_{j=1}^{k_*}\Big[\sum_{i\in\mathcal{A}_j}\exp\Big(\frac{\bzin}{\tau^n}\Big)-\exp\Big(\frac{\bzj}{\tau^*}\Big)\Big]\Big[\exp\Big(\frac{(\boj)^{\top}x}{\tau^*}\Big)\mathcal{E}(x,\ej)-\exp\Big(\frac{(\boj)^{\top}x}{\tau^*}\Big)g_{G_n}(x)\Big]\\
    &:=A_n(x)-B_n(x)+C_{n}(x).
\end{align*}
By plugging in the values of $\eta^n_i$, $\omega^n_i$, $\tau^n$ and $\beta^n_i$, the term $A_n(x)$ becomes
\begin{align*}
    A_n(x)=\sum_{i=1}^{k_*}\sum_{i\in\mathcal{A}_j}\exp\Big(\frac{\bzin}{\tau^n}\Big)\Big[\exp\Big(\frac{(\boin)^{\top}x}{\tau^n}\Big)-\exp\Big(\frac{(\boj)^{\top}x}{\tau^*}\Big)\Big]\mathcal{E}(x,\ej).
\end{align*}
By applying the first-order Taylor expansions to the function $\exp\Big(\frac{(\boin)^{\top}x}{\tau^n}\Big)$ around the point $(\boj,\tau^*)$, we have
\begin{align*}
    A_n(x)=\sum_{j=1}^{k_*}\sum_{i\in\mathcal{A}_j}\sum_{u=1}^{d}\exp\Big(\frac{\bzin}{\tau^n}\Big)\Big[\frac{s^{(u)}_{n,j}}{\tau^*}-\frac{t_n(\boj)^{(u)}}{(\tau^*)^2}\Big]\cdot x^{(u)}\exp\Big(\frac{(\boj)^{\top}x}{\tau^*}\Big) \mathcal{E}(x,\ej) +R_1(x),
\end{align*}
where $R_1(x)$ is a Taylor remainder such that $R_1(x)/\mathcal{L}_{3,r}(G_n,G_*)\to0$ as $n\to\infty$. Then, by setting
\begin{align*}
    t_n=\frac{1}{n}; \qquad s^{(u)}_{n,j}=\dfrac{t_n(\boj)^{(u)}}{\tau^*}=\dfrac{(\boj)^{(u)}}{n\tau^*},
\end{align*}
we get $A_n(x)/\mathcal{L}_{3,r}(G_n,G_*)\to0$ as $n\to\infty$.

\vspace{0.5em}
\noindent
Analogously, we also have $B_n(x)/\mathcal{L}_{3,r}(G_n,G_*)\to0$ as $n\to\infty$. Furthermore, since it can be checked that $C_{n}(x)=0$, we deduce $Q_n(x)/\mathcal{L}_{3,r}(G_n,G_*)\to0$ as $n\to\infty$. Note that the term $\Big[\sum_{j=1}^{k_*}\exp\Big(\dfrac{(\boj)^{\top}x+\bzj}{\tau^*}\Big)\Big]$ is bounded, then it follows that $$|g_{G_n}(x)-g_{G_*}(x)|/\mathcal{L}_{3,r}(G_n,G_*)\to0$$ as $n\to\infty$ for almost every $x$. The above result directly leads to equation~\eqref{eq:ratio_zero_over}. Hence, the proof is completed.

\subsection{Proof of Theorem~\ref{theorem:general_dense-to-sparse_experts}}
\label{appendix:general_dense-to-sparse_experts}
\noindent
In this proof, we aim to establish the following inequality:
\begin{align}
    \label{eq:activation_universal_inequality_temperature}
    \inf_{G\in\mathcal{G}_{k}(\Theta)}\normf{g_{G}-g_{G_*}}/\mathcal{L}_{4}(G,G_*)>0.
\end{align}
By employing the proof framework of Theorem~\ref{theorem:general_experts} in Appendix~\ref{appendix:general_experts}, we will first derive the local part of equation~\eqref{eq:activation_universal_inequality_temperature}, which is given by
\begin{align}
    \label{eq:general_local_inequality_temperature}
    \lim_{\varepsilon\to0}\inf_{G\in\mathcal{G}_{k}(\Theta):\mathcal{L}_4(G,G_*)\leq\varepsilon}\normf{g_{G}-g_{G_*}}/\mathcal{L}_4(G,G_*)>0.
\end{align}
As a consequence, there exists some $\varepsilon'>0$ such that
\begin{align*}
    \inf_{G\in\mathcal{G}_{k}(\Theta):\mathcal{L}_4(G,G_*)\leq\varepsilon'}\normf{g_{G}-g_{G_*}}/\mathcal{L}_4(G,G_*)>0.
\end{align*}
Thus, it is sufficient to prove the global part of equation~\eqref{eq:activation_universal_inequality_temperature}, that is,
\begin{align*}
    \inf_{G\in\mathcal{G}_{k}(\Theta):\mathcal{L}_4(G,G_*)>\varepsilon'}\normf{g_{G}-g_{G_*}}/\mathcal{L}_4(G,G_*)>0.
\end{align*}
Notably, since the global part can be argued in a similar fashion to that in Appendix~\ref{appendix:general_experts}, it is omitted here. In other words, we will provide only the proof of local part here.

\vspace{0.5em}
\noindent
\textbf{Proof of the local part~\eqref{eq:general_local_inequality_temperature}:} Assume by contrary that the above inequality does not hold true, then there exists a sequence of mixing measures $G_n=\sum_{i=1}^{k_*}\exp(\beta^n_{i})\delta_{(\omega^n_{i},\tau^n,\eta^n_i)}$ in $\mathcal{G}_{k}(\Theta)$ such that $\mathcal{L}_{4n}:=\mathcal{L}_4(G_n,G_*)\to0$ and
\begin{align}
    \label{eq:general_ratio_limit_temperature}
    \normf{g_{G_n}-g_{G_*}}/\mathcal{L}_{4n}\to0,
\end{align}
as $n\to\infty$. Let us denote by $\mathcal{A}^n_j:=\mathcal{A}_j(G_n)$ a Voronoi cell of $G_n$ generated by the $j$-th components of $G_*$. Since our arguments are asymptotic, we may assume that those Voronoi cells do not depend on the sample size, i.e. $\mathcal{A}_j=\mathcal{A}^n_j$. Thus, the Voronoi loss $\mathcal{L}_{4n}$ can be represented as
\begin{align}
    \label{eq:general_loss_proof_temperature}
   \mathcal{L}_{4n}&:=\sum_{j=1}^{k_*}\Big|\sum_{i\in\mathcal{A}_j}\exp\Big(\frac{\bzin}{\tau^n}\Big)-\exp\Big(\frac{\beta^*_{j}}{\tau^*}\Big)\Big|+\sum_{j:|\mathcal{A}_j|=1}\sum_{i\in\mathcal{A}_j}\exp\Big(\frac{\bzin}{\tau^n}\Big)\Big[\|\dboijn\|+|\Delta\tau^n|+\|\deijn\|\Big]\nonumber\\
    &\hspace{4cm}+\sum_{j:|\mathcal{A}_j|>1}\sum_{i\in\mathcal{A}_j}\exp\Big(\frac{\bzin}{\tau^n}\Big)\Big[\|\dboijn\|^2+|\Delta\tau^n|^2+\|\deijn\|^2\Big],
\end{align}
where we denote $\dboijn:=\omega^n_{i}-\omega^*_{j}$, $\Delta\tau^n:=\tau^n-\tau^*$, and $\deijn:=\eta^n_i-\eta^*_j$.\\

\noindent
Since $\mathcal{L}_{4n}\to0$, we get that $(\boin,\tau^n,\ein)\to(\boj,\tau^*,\ej)$ and $\sum_{i\in\mathcal{A}_j}\exp\Big(\frac{\bzin}{\tau^n}\Big)\to\exp\Big(\frac{\beta^*_{j}}{\tau^*}\Big)$ as $n\to\infty$ for any $i\in\mathcal{A}_j$ and $j\in[k_*]$. Now, we divide the proof of local part into three steps as follows:\\

\noindent
\textbf{Step 1 - Decompose the difference between regression functions.} To begin with, we use Taylor expansions to decompose the quantity $Q_n(x):=\Big[\sum_{j=1}^{k_*}\exp\Big(\frac{\pi(x,\omega^*_j)+\beta^*_{j}}{\tau^*}\Big)\Big]\cdot[g_{G_n}(x)-g_{G_*}(x)]$ into a combination of linearly independent terms. For this sake, we denote $F(x;\omega,\tau,\eta):=\exp\Big(\frac{\pi(x,\omega)}{\tau}\Big)\mathcal{E}(x,\eta)$ and $H(x;\omega,\tau)=\exp\Big(\frac{\pi(x,\omega)}{\tau}\Big)g_{G_n}(x)$. Then, $Q_n(x)$ can be decomposed as
\begin{align}
    \label{eq:general_Q_n_temperature}
    Q_n(x)&=\sum_{j=1}^{k_*}\sum_{i\in\mathcal{A}_j}\exp\Big(\frac{\beta^n_{i}}{\tau^n}\Big)\Big[F(x;\boin,\tn,\ein)-F(x;\boj,\tj,\ej)\Big]\nonumber\\
    &-\sum_{j=1}^{k_*}\sum_{i\in\mathcal{A}_j}\exp\Big(\frac{\beta^n_{i}}{\tau^n}\Big)\Big[H(x;\boin,\tn)-H(x;\boj,\tj)\Big]\nonumber\\
    &+\sum_{j=1}^{k_*}\Big(\sum_{i\in\mathcal{A}_j}\exp\Big(\frac{\bzin}{\tau^n}\Big)-\exp\Big(\frac{\bzj}{\tau^*}\Big)\Big)\Big[F(x;\boj,\tj,\ej)-H(x;\boj,\tj)\Big]\nonumber\\
    &:=A_n(x)-B_n(x)+C_{n}(x).
\end{align}
Next, we will decompose the terms $A_n(x)$ and $B_n(x)$, respectively. 

\vspace{0.5em}
\noindent
\textbf{Step 1A- Decompose $A_n(x)$.} We can represent $A_n(x)$ as
\begin{align*}
    A_n(x)&:=\sum_{j:|\mathcal{A}_j|=1}\sum_{i\in\mathcal{A}_j}\exp\Big(\frac{\beta^n_{i}}{\tau^n}\Big)\Big[F(x;\boin,\tn,\ein)-F(x;\boj,\tj,\ej)\Big]\\
    &+\sum_{j:|\mathcal{A}_j|>1}\sum_{i\in\mathcal{A}_j}\exp\Big(\frac{\beta^n_{i}}{\tau^n}\Big)\Big[F(x;\boin,\tn,\ein)-F(x;\boj,\tj,\ej)\Big]\\
    &:=A_{n,1}(x)+A_{n,2}(x).
\end{align*}
By applying the first-order Taylor expansion, we have
\begin{align*}
    A_{n,1}(x)=\sum_{j:|\mathcal{A}_j|=1}\sum_{i\in\mathcal{A}_j}\exp\Big(\frac{\beta^n_{i}}{\tau^n}\Big)\sum_{|\alpha|=1}(\dboijn)^{\alpha_1}(\dtn)^{\alpha_2}(\deijn)^{\alpha_3}\cdot\frac{\partial F}{\partial\omega^{\alpha_1}\partial\tau^{\alpha_2}\partial \eta^{\alpha_3}}(x;\boj,\tj,\ej)\\
    +R_1(x),
\end{align*}
where $R_1(x)$ is a Taylor remainder such that $R_1(x)/\mathcal{L}_{4n}\to0$ as $n\to\infty$. Note that the first derivatives of $F$ w.r.t its parameters are given by
\begin{align*}
    \frac{\partial F}{\partial\omega}(x;\boj,\tj,\ej)&=\frac{1}{\tj}\cdot\frac{\partial\pi}{\partial\omega}(x,\boj)\exp\Big(\frac{\pi(x,\boj)}{\tj}\Big)\mathcal{E}(x,\ej)=\frac{1}{\tj}\cdot\frac{\partial\pi}{\partial\omega}(x,\boj)F(x;\boj,\tj,\ej),\\
    \frac{\partial F}{\partial\tau}(x;\boj,\tj,\ej)&=-\frac{\pi(x,\boj)}{(\tj)^2}\exp\Big(\frac{\pi(x,\boj)}{\tj}\Big)\mathcal{E}(x,\ej)=-\frac{\pi(x,\boj)}{(\tj)^2}F(x;\boj,\tj,\ej),\\
    \frac{\partial F}{\partial\eta}(x;\boj,\tj,\ej)&=\exp\Big(\frac{\pi(x,\boj)}{\tj}\Big)\frac{\partial\mathcal{E}}{\partial\eta}(x,\ej):=F_1(x;\boj,\tj,\ej).
\end{align*}
Then, $A_{n,1}(x)$ can be rewritten as
\begin{align}
    \label{eq:general_A_n_1_temperature}
    A_{n,1}(x)&=\sum_{j:|\mathcal{A}_j|=1}A_{n,1,j}(x)+R_1(x),
\end{align}
where
\begin{align*}
    A_{n,1,j}(x):=\sum_{i\in\mathcal{A}_j}\exp\Big(\frac{\beta^n_{i}}{\tau^n}\Big)\Big[\Big(\frac{1}{\tj}(\dboijn)^{\top}\frac{\partial\pi}{\partial\omega}(x,\boj)-(\dtn)\cdot\frac{\pi(x,\boj)}{(\tj)^2}\Big)\cdot F(x;\boj,\tj,\ej)+\\
    (\deijn)^{\top} F_1(x;\boj,\tj,\ej)\Big].
\end{align*}
Next, by means of the second-order Taylor expansion, we can decompose $A_{n,2}(x)$ as
\begin{align*}
    A_{n,2}(x)=\sum_{j:|\mathcal{A}_j|>1}\sum_{i\in\mathcal{A}_j}\exp\Big(\frac{\beta^n_{i}}{\tn}\Big)\sum_{|\alpha|=1}^{2}\frac{1}{\alpha!}(\dboijn)^{\alpha_1}(\dtn)^{\alpha_2}(\deijn)^{\alpha_3}\cdot\frac{\partial^{|\alpha_1|+\alpha_2+|\alpha_3|} F}{\partial\omega^{\alpha_1}\partial\tau^{\alpha_2}\partial \eta^{\alpha_3}}(x;\boj,\tj,\ej)+R_2(x),
\end{align*}
where $R_2(x)$ is a Taylor remainder such that $R_2(x)/\mathcal{L}_{4n}\to0$ as $n\to\infty$. By taking the second derivatives of $F$ w.r.t its parameters, we have
\begin{align*}
    \frac{\partial^2F}{\partial\omega\partial\omega^{\top}}(x;\boj,\tj,\ej)&=\Big[\frac{1}{\tj}\frac{\partial^2\pi}{\partial\omega\partial\omega^{\top}}(x,\boj)+\frac{1}{(\tj)^2}\frac{\partial\pi}{\partial\omega}(x,\boj)\cdot\frac{\partial\pi}{\partial\omega^{\top}}(x,\boj)\Big]F(x;\boj,\tj,\ej),\\
    \frac{\partial^2F}{\partial\tau^2}(x;\boj,\tj,\ej)&=\Big[2\frac{\pi(x,\boj)}{(\tj)^3}+\frac{\pi^2(x,\boj)}{(\tj)^4}\Big]F(x;\boj,\ej,\tj),\\
    \frac{\partial^2F}{\partial\eta\partial\eta^{\top}}(x;\boj,\tj,\ej)&=\exp\Big(\frac{\pi(x,\boj)}{\tj}\Big)\mathcal{E}(x,\ej):=F_2(x;\boj,\tj,\ej),\\
    \frac{\partial^2F}{\partial\omega\partial\tau}(x;\boj,\tj,\ej)&=\Big[-\frac{1}{(\tj)^2}\frac{\partial\pi}{\partial\omega}(x,\boj)-\frac{\pi(x,\boj)}{(\tj)^3}\frac{\partial\pi}{\partial\omega}(x,\boj)\Big]F(x;\boj,\tj,\ej),\\
    \frac{\partial^2F}{\partial\omega\partial\eta^{\top}}(x;\boj,\tj,\ej)&=\frac{1}{\tj}\frac{\partial\pi}{\partial\omega}(x,\boj)[F_1(x;\boj,\tj,\ej)]^{\top},\\
    \frac{\partial^2F}{\partial\tau\partial\eta}(x;\boj,\tj,\ej)&=-\frac{\pi(x,\boj)}{(\tj)^2}F_1(x;\boj,\tj,\ej).
\end{align*}
Then, we can represent $A_{n,2}(x)$ as
\begin{align}
     \label{eq:general_A_n_2_temperature}
    A_{n,2}(x)=\sum_{j:|\mathcal{A}_j|>1}[A_{n,1,j}(x)+A_{n,2,j}(x)]+R_2(x),
\end{align}
where 
\begin{align*}
    &A_{n,2,j}(x):=\sum_{i\in\mathcal{A}_j}\exp\Big(\frac{\beta^n_{i}}{\tau^n}\Big)\Bigg\{\Big[(\dboijn)^{\top}M_d\odot\Big(\frac{1}{\tj}\frac{\partial^2\pi}{\partial\omega\partial\omega^{\top}}(x,\boj)+\frac{1}{(\tj)^2}\frac{\partial\pi}{\partial\omega}(x,\boj)\cdot\frac{\partial\pi}{\partial\omega^{\top}}(x,\boj)\Big)(\dboijn)\\
    &+\Big(\frac{\pi(x,\boj)}{(\tj)^3}+\frac{\pi^2(x,\boj)}{2(\tj)^4}\Big)(\dtn)^2-(\dboijn)^{\top}\Big(\frac{1}{(\tj)^2}\frac{\partial\pi}{\partial\omega}(x,\boj)+\frac{\pi(x,\boj)}{(\tj)^3}\frac{\partial\pi}{\partial\omega}(x,\boj)\Big)(\dtn)\Big]\cdot F(x;\boj,\tj,\ej)\\
    &+\Big[(\dboijn)^{\top}\frac{1}{\tj}\frac{\partial\pi}{\partial\omega}(x,\boj)[F_1(x;\boj,\tj,\ej)]^{\top}(\deijn)-(\deijn)^{\top}\frac{\pi(x,\boj)}{(\tj)^2}F_1(x;\boj,\tj,\ej)(\dtn)\Big]\\
    &+(\deijn)^{\top}M_d\odot F_2(x;\boj,\tj,\ej)(\deijn)\Bigg\},
\end{align*}
with $M_d$ being an $d\times d$ matrix whose diagonal entries are $\frac{1}{2}$ while other entries are 1.

\vspace{0.5em}
\noindent
\textbf{Step 1B - Decompose $B_n(x)$.} Next, we apply the same strategy of decomposing $A_n(x)$ for partitioning the term $B_n(x)$. In particular, we first have
\begin{align*}
    B_n(x)&=\sum_{j:|\mathcal{A}_j|=1}\sum_{i\in\mathcal{A}_j}\exp\Big(\frac{\beta^n_{i}}{\tau^n}\Big)\Big[H(x;\boin,\tn)-H(x;\boj,\tj)\Big]\\
    &+\sum_{j:|\mathcal{A}_j|>1}\sum_{i\in\mathcal{A}_j}\exp\Big(\frac{\beta^n_{i}}{\tau^n}\Big)\Big[H(x;\boin,\tn)-H(x;\boj,\tj)\Big]\\
    &:=B_{n,1}(x)+B_{n,2}(x).
\end{align*}
Regarding the term $B_{n,1}(x)$, by employing the first-order Taylor expansion, we have
\begin{align}
    \label{eq:general_B_n_1_temperature}
    B_{n,1}(x)&=\sum_{j:|\mathcal{A}_j|=1}\sum_{i\in\mathcal{A}_j}\exp\Big(\frac{\beta^n_{i}}{\tau^n}\Big)\sum_{|\alpha|=1}(\dboijn)^{\alpha_1}(\dtn)^{\alpha_2}\cdot\frac{\partial H}{\partial\omega^{\alpha_1}\partial\tau^{\alpha_2}}(x;\boj,\tj)+R_3(x)\nonumber\\
    &=\sum_{j:|\mathcal{A}_j|=1}\sum_{i\in\mathcal{A}_j}\exp\Big(\frac{\beta^n_{i}}{\tau^n}\Big)\Big[(\dboijn)^{\top}\frac{\partial\pi}{\partial\omega}(x,\boj)-(\dtn)\frac{\pi(x,\boj)}{(\tj)^2}\Big]\cdot H(x;\boj,\tj)+R_3(x)\nonumber\\
    &=\sum_{j:|\mathcal{A}_j|=1}B_{n,1,j}(x)+R_3(x),
\end{align}
where $R_3(x)$ is a Taylor remainder such that $R_3(x)/\mathcal{L}_{4n}\to0$ as $n\to\infty$, and
\begin{align*}
    B_{n,1,j}(x):=\sum_{i\in\mathcal{A}_j}\exp\Big(\frac{\beta^n_{i}}{\tau^n}\Big)\Big[\frac{(\dboijn)^{\top}}{\tj}\frac{\partial\pi}{\partial\omega}(x,\boj)-(\dtn)\frac{\pi(x,\boj)}{(\tj)^2}\Big]\cdot H(x;\boj,\tj).
\end{align*}
Similarly, by means of the second-order Taylor expansion, we get
\begin{align}
      \label{eq:general_B_n_2_temperature}
    B_{n,2}(x)&=\sum_{j:|\mathcal{A}_j|>1}\sum_{i\in\mathcal{A}_j}\exp\Big(\frac{\beta^n_{i}}{\tau^n}\Big)\sum_{|\alpha|=1}^{2}(\dboijn)^{\alpha_1}(\dtn)^{\alpha_2}\cdot\frac{\partial H}{\partial\omega^{\alpha_1}\partial\tau^{\alpha_2}}(x;\boj,\tj)+R_4(x)\nonumber\\
    &=\sum_{j:|\mathcal{A}_j|>1}[B_{n,1,j}(x)+B_{n,2,j}(x)]+ R_4(x),
\end{align}
where $R_4(x)$ is a Taylor remainder such that $R_4(x)/\mathcal{L}_{4n}\to0$ as $n\to\infty$, and
\begin{align*}
    &B_{n,2,j}(x):=\sum_{i\in\mathcal{A}_j}\exp\Big(\frac{\beta^n_{i}}{\tau^n}\Big)\Big[(\dboijn)^{\top}M_d\odot\Big(\frac{1}{\tj}\frac{\partial^2\pi}{\partial\omega\partial\omega^{\top}}(x,\boj)+\frac{1}{(\tj)^2}\frac{\partial\pi}{\partial\omega}(x,\boj)\cdot\frac{\partial\pi}{\partial\omega^{\top}}(x,\boj)\Big)(\dboijn)\\
    &+\Big(\frac{\pi(x,\boj)}{(\tj)^3}+\frac{\pi^2(x,\boj)}{2(\tj)^4}\Big)(\dtn)^2-(\dboijn)^{\top}\Big(\frac{1}{(\tj)^2}\frac{\partial\pi}{\partial\omega}(x,\boj)+\frac{\pi(x,\boj)}{(\tj)^3}\frac{\partial\pi}{\partial\omega}(x,\boj)\Big)(\dtn)\Big]\cdot H(x;\boj).
\end{align*}
Putting the above results together, we see that $[A_n(x)-R_1(x)-R_2(x)]/\mathcal{L}_{4n}$, $[B_n(x)-R_3(x)-R_4(x)]/\mathcal{L}_{4n}$ and $C_{n}(x)/\mathcal{L}_{4n}$ can be written as a combination of elements from the following set
\begin{align*}
    \Big\{F_j, \pi(x,\boj) F_j, \pi^2(x,\boj) F_j, \frac{\partial\pi}{\partial\omega^{(u)}}(x,\boj) F_j, \pi(x,\boj)\frac{\partial\pi}{\partial\omega^{(u)}}(x,\boj)F_j, \frac{\partial\pi}{\partial\omega^{(u)}}(x,\boj)\frac{\partial\pi}{\partial\omega^{(v)}}(x,\boj)F_j,\\
    \frac{\partial^2\pi}{\partial\omega^{(u)}\partial\omega^{(v)}}(x,\boj)F_j:j\in[k_*], u,v\in[d]\Big\}\\
    \cup~\Big\{(F_{1j})^{(u)}, \pi(x,\boj)(F_{1j})^{(u)}, \frac{\partial\pi}{\partial\omega^{(u)}}(x,\boj)(F_{1j})^{(v)}:j\in[k_*],u,v\in[d]\Big\}\\
    \cup~\Big\{(F_{2j})^{(uv)}:j\in[k_*],u,v\in[d]\Big\}\\
    \cup~\Big\{H_j, \pi(x,\boj)H_j, \pi^2(x,\boj)H_j, \frac{\partial\pi}{\partial\omega^{(u)}}(x,\boj)H_j,\pi(x,\boj)\frac{\partial\pi}{\partial\omega^{(u)}}(x,\boj)H_j,\frac{\partial\pi}{\partial\omega^{(u)}}(x,\boj)\frac{\partial\pi}{\partial\omega^{(v)}}(x,\boj)H_j,\\
    \frac{\partial^2\pi}{\partial\omega^{(u)}\partial\omega^{(v)}}(x,\boj)H_j:j\in[k_*], u,v\in[d]\Big\},
\end{align*}
where we denote
\begin{align*}
    F_j&:=F(x;\boj,\tj,\ej),\\
    F_{1j}&:=\exp\Big(\frac{\pi(x,\boj)}{\tj}\Big)\frac{\partial\mathcal{E}}{\partial\eta}(x,\ej),\\
    F_{2j}&:=\exp\Big(\frac{\pi(x,\boj)}{\tj}\Big)\frac{\partial^2\mathcal{E}}{\partial\eta\partial\eta^{\top}}(x,\ej).
\end{align*}
\textbf{Step 2 - Non-vanishing coefficients.} Moving to this step, we utilize the proof by contradiction method to show that not all coefficients in the representations of $[A_n-R_1(x)-R_2(x)]/\mathcal{L}_{4n}$, $[B_n-R_3(x)-R_4(x)]/\mathcal{L}_{4n}$ and $C_{n}(x)/\mathcal{L}_{4n}$ converge to zero as $n\to\infty$. Indeed, assume by contrary that all of them go to zero and consider the coefficients of the term
\begin{itemize}
    \item $F(x;\boj,\tj,\ej)$ for $j\in[k_*]$, we get that
    \begin{align*}
        \frac{1}{\mathcal{L}_{4n}}\sum_{j=1}^{k_*}\Big|\sum_{i\in\mathcal{A}_j}\exp\Big(\frac{\beta^n_{i}}{\tau^n}\Big)-\exp\Big(\frac{\beta^*_{j}}{\tau^*}\Big)\Big|\to0;
    \end{align*}
    \item $\frac{\partial\pi}{\partial\omega^{(u)}}(x,\boj)F(x;\boj,\tj,\ej)$ for $u\in[d]$ and $j\in[k_*]:|\mathcal{A}_j|=1$, we get that
    \begin{align*}
        \frac{1}{\mathcal{L}_{4n}}\sum_{j:|\mathcal{A}_j|=1}\sum_{i\in\mathcal{A}_j}\exp\Big(\frac{\beta^n_{i}}{\tau^n}\Big)\|\dboijn\|_1\to0.
    \end{align*}
    Due to the equivalency between the $\ell_1$-norm and the $\ell_2$-norm, we deduce that
    \begin{align*}
        \frac{1}{\mathcal{L}_{4n}}\sum_{j:|\mathcal{A}_j|=1}\sum_{i\in\mathcal{A}_j}\exp\Big(\frac{\beta^n_{i}}{\tau^n}\Big)\|\dboijn\|\to0.
    \end{align*}
    \item $\pi(x,\boj)F(x;\boj,\tj,\ej)$ for $u\in[d]$ and $j:|\mathcal{A}_j|=1$, we get that
    \begin{align*}
        \frac{1}{\mathcal{L}_{4n}}\sum_{j:|\mathcal{A}_j|=1}\sum_{i\in\mathcal{A}_j}\exp\Big(\frac{\beta^n_{i}}{\tau^n}\Big)\|\dtn\|\to0.
    \end{align*}
    \item $[F_1(x;\boj,\tj,\ej)]^{(u)}$ for $u\in[d]$ and $j:|\mathcal{A}_j|=1$, we get that
    \begin{align*}
        \frac{1}{\mathcal{L}_{4n}}\sum_{j:|\mathcal{A}_j|=1}\sum_{i\in\mathcal{A}_j}\exp\Big(\frac{\beta^n_{i}}{\tau^n}\Big)\|\deijn\|\to0.
    \end{align*}
    \item $\frac{\partial^2\pi}{\partial\omega^{(u)}\partial\omega^{(u)}}(x,\boj)F(x;\boj,\tj,\ej)$ for $u\in[d]$ and $j:|\mathcal{A}_j|>1$, we get that
    \begin{align*}
        \frac{1}{\mathcal{L}_{4n}}\sum_{j:|\mathcal{A}_j|>1}\sum_{i\in\mathcal{A}_j}\exp\Big(\frac{\beta^n_{i}}{\tau^n}\Big)\|\dboijn\|^2\to0.
    \end{align*}
    \item $\pi^2(x,\boj)F(x;\boj,\tj,\ej)$ for $u\in[d]$ and $j:|\mathcal{A}_j|>1$, we get that
    \begin{align*}
        \frac{1}{\mathcal{L}_{4n}}\sum_{j:|\mathcal{A}_j|>1}\sum_{i\in\mathcal{A}_j}\exp\Big(\frac{\beta^n_{i}}{\tau^n}\Big)\|\dtn\|^2\to0.
    \end{align*}
    \item $[F_2(x;\boj,\tj,\ej)]^{(uu)}$ for $u\in[d]$ and $j:|\mathcal{A}_j|>1$, we get that
    \begin{align*}
        \frac{1}{\mathcal{L}_{4n}}\sum_{j:|\mathcal{A}_j|>1}\sum_{i\in\mathcal{A}_j}\exp\Big(\frac{\beta^n_{i}}{\tau^n}\Big)\|\deijn\|^2\to0.
    \end{align*}
\end{itemize}
Combine all the above limits, we arrive at $1=\mathcal{L}_{4n}/\mathcal{L}_{4n}\to0$ as $n\to\infty$, which is a contradiction. Therefore, not all the coefficients in the representations of $[A_n(x)-R_1(x)-R_2(x)]/\mathcal{L}_{4n}$, $[B_n(x)-R_3(x)-R_4(x)]/\mathcal{L}_{4n}$ and $C_{n}(x)/\mathcal{L}_{4n}$ go to zero as $n\to\infty$.\\

\noindent
\textbf{Step 3 - Application of the Fatou's lemma.} Now, we attempt to exhibit a contradiction to the conclusion of Step 2. Let $m_n$ be the maximum of the absolute values of the coefficients in the representations of $[A_n(x)-R_1(x)-R_2(x)]/\mathcal{L}_{4n}$, $[B_n(x)-R_3(x)-R_4(x)]/\mathcal{L}_{4n}$ and $C_{n}(x)/\mathcal{L}_{4n}$. As at least one among those coefficients does not go zero, we have $1/m_n\not\to\infty$. 

\vspace{0.5em}
\noindent
Recall from the hypothesis in equation~\eqref{eq:general_ratio_limit_temperature} that $\normf{g_{G_n}-g_{G_*}}/\mathcal{L}_{4n}\to0$ as $n\to\infty$. This result also implies that  $\|g_{G_n}-g_{G_*}\|_{L^1(\mu)}/\mathcal{L}_{4n}\to0$. Then, by applying the Fatou's lemma, we get
\begin{align*}
    0=\lim_{n\to\infty}\frac{\|g_{G_n}-g_{G_*}\|_{L^1(\mu)}}{m_n\mathcal{L}_{4n}}\geq \int \liminf_{n\to\infty}\frac{|g_{G_n}(x)-g_{G_*}(x)|}{m_n\mathcal{L}_{4n}}\dint\mu(x)\geq 0.
\end{align*}
As a result, we have $[g_{G_n}(x)-g_{G_*}(x)]/[m_n\mathcal{L}_{4n}]\to0$ for almost every $x$. Since the parameter $\Theta$ is compact, the term $\sum_{j=1}^{k_*}\exp\Big(\frac{\pi(x,\boj)}{\tj}+\bzj\Big)$ is bounded. Thus, it follows that $Q_n(x)/[m_n\mathcal{L}_{4n}]\to0$, or equivalently, 
\begin{align}
    \label{eq:general_zero_limit_temperature}
    \frac{1}{m_n\mathcal{L}_{1n}}\cdot\Big[(A_{n,1}(x)-R_1(x)+A_{n,2}(x)-R_2(x))-(B_{n,1}(x)-R_3(x)+&B_{n,2}(x)-R_4(x))+C_{n}(x)\Big]\to0.
\end{align}
For ease of presentation, we denote
\begin{align*}
    \frac{1}{m_n\mathcal{L}_{4n}}\cdot\sum_{i\in\mathcal{A}_j}\exp\Big(\frac{\bzin}{\tn}\Big)(\dboijn)\to\phi_{1,j}&,\quad  \frac{1}{m_n\mathcal{L}_{4n}}\cdot\sum_{i\in\mathcal{A}_j}\exp\Big(\frac{\bzin}{\tn}\Big)(\dboijn)(\dboijn)^{\top}\to\phi_{2,j},\\
    \frac{1}{m_n\mathcal{L}_{4n}}\cdot\sum_{i\in\mathcal{A}_j}\exp\Big(\frac{\bzin}{\tn}\Big)(\dtn)\to\chi_{1,j}&,\quad \frac{1}{m_n\mathcal{L}_{4n}}\cdot\sum_{i\in\mathcal{A}_j}\exp\Big(\frac{\bzin}{\tn}\Big)(\dtn)^2\to\chi_{2,j},\\
    \frac{1}{m_n\mathcal{L}_{4n}}\cdot\sum_{i\in\mathcal{A}_j}\exp\Big(\frac{\bzin}{\tn}\Big)(\deijn)\to\varphi_{1,j}&,\quad \frac{1}{m_n\mathcal{L}_{4n}}\cdot\sum_{i\in\mathcal{A}_j}\exp\Big(\frac{\bzin}{\tn}\Big)(\deijn)(\deijn)^{\top}\to\varphi_{2,j},\\
    \frac{1}{m_n\mathcal{L}_{4n}}\cdot\sum_{i\in\mathcal{A}_j}\exp\Big(\frac{\bzin}{\tn}\Big)(\dboijn)(\dtn)\to\zeta_{1,j}&, \quad \frac{1}{m_n\mathcal{L}_{4n}}\cdot\sum_{i\in\mathcal{A}_j}\exp\Big(\frac{\bzin}{\tn}\Big)(\dboijn)(\deijn)^{\top}\to\zeta_{2,j}\\
    \frac{1}{m_n\mathcal{L}_{4n}}\cdot\sum_{i\in\mathcal{A}_j}\exp\Big(\frac{\bzin}{\tn}\Big)(\deijn)(\dtn)\to\zeta_{3,j}&, \quad \frac{1}{m_n\mathcal{L}_{4n}}\cdot\Big(\sum_{i\in\mathcal{A}_j}\exp\Big(\frac{\bzin}{\tn}\Big)-\exp\Big(\frac{\bzj}{\tj}\Big)\Big)\to\xi_j.
\end{align*}
Above, at least one among the limits $\phi^{(u)}_{1,j}$, $\phi^{(uu)}_{2,j}$, $\chi_{1,j}$, $\chi_{2,j}$, $\varphi^{(u)}_{1,j}$, $\varphi^{(uu)}_{2,j}$ and $\xi_j$, for $j\in[k_*]$, is non-zero as a result of the conclusion in Step 2. 
Additionally, let us denote $F_{\rho j}:=F_{\rho}(x;\boj,\tj,\ej)$ for $\rho\in\{1,2\}$ and $H_j=H(x;\boj,\tj)$. From the formulation of 
\begin{itemize}
    \item $A_{n,1}(x)$ in equation~\eqref{eq:general_A_n_1_temperature}, we get 
    \begin{align}
        \label{eq:general_limit_1_temperature}
        \frac{A_{n,1}(x)-R_1(x)}{m_n\mathcal{L}_{4n}}\to\sum_{j:|\mathcal{A}_j|=1}\Big[\Big(\frac{(\phi_{1,j})^{\top}}{\tj}\frac{\partial\pi}{\partial\omega}(x,\boj)-\chi_{1,j}\cdot\frac{\pi(x,\boj)}{(\tj)^2}\Big)\cdot F_j+\varphi_{1,j}^{\top} F_{1j}\Big].
    \end{align}
    \item $A_{n,2}(x)$ in equation~\eqref{eq:general_A_n_2_temperature}, we get
    \begin{align}
    \label{eq:general_limit_2_temperature}
        &\frac{A_{n,2}(x)-R_2(x)}{m_n\mathcal{L}_{2n}}\to\sum_{j:|\mathcal{A}_j|>1}\Bigg\{\Big[\frac{(\phi_{1,j})^{\top}}{\tj}\frac{\partial\pi}{\partial\omega}(x,\boj)-\chi_{1,j}\cdot\frac{\pi(x,\boj)}{(\tj)^2}\nonumber\\
        &+M_d\odot\Big(\frac{1}{\tj}\frac{\partial^2\pi}{\partial\omega\partial\omega^{\top}}(x,\boj)+\frac{1}{(\tj)^2}\frac{\partial\pi}{\partial\omega}(x,\boj)\cdot\frac{\partial\pi}{\partial\omega^{\top}}(x,\boj)\Big)\odot \phi_{2,j}\nonumber\\
        &+\Big(\frac{\pi(x,\boj)}{(\tj)^3}+\frac{\pi^2(x,\boj)}{2(\tj)^4}\Big)\chi_{2,j}-\zeta_{1,j}^{\top}\Big(\frac{1}{(\tj)^2}\frac{\partial\pi}{\partial\omega}(x,\boj)+\frac{\pi(x,\boj)}{(\tj)^3}\frac{\partial\pi}{\partial\omega}(x,\boj)\Big)\Big]\cdot F_j\nonumber\\
        &+[\varphi_{1,j}^{\top} F_{1j}+\Big(\frac{1}{\tj}F_{1j}\frac{\partial\pi}{\partial\omega^{\top}}(x,\boj)\Big)\odot\zeta_{2,j}-\frac{\pi(x,\boj)}{(\tj)^2}\cdot\zeta_{3,j}^{\top}F_{1j}]+\Big(M_{d}\odot \varphi_{2,j}\Big)\odot F_{2j}\Bigg\}.
    \end{align}
    \item $B_{n,1}(x)$ in equation~\eqref{eq:general_B_n_1_temperature}, we get
    \begin{align}
        \label{eq:general_limit_3_temperature}
        \frac{B_{n,1}(x)-R_3(x)}{m_n\mathcal{L}_{2n}}\to\sum_{j:|\mathcal{A}_j|=1}\Big[\frac{\phi^{\top}_{1,j}}{\tj}\frac{\partial\pi}{\partial  \omega}(x,\boj)-\chi_{1,j}\cdot\frac{\pi(x,\boj)}{(\tj)^2}\Big]\cdot H_j.
    \end{align}
    \item $B_{n,2}(x)$ in equation~\eqref{eq:general_B_n_2_temperature}, we get
    \begin{align}
        \label{eq:general_limit_4_temperature}
        &\frac{B_{n,2}(x)-R_4(x)}{m_n\mathcal{L}_{2n}}\to\sum_{j:|\mathcal{A}_j|>1}\Big[\frac{(\phi_{1,j})^{\top}}{\tj}\frac{\partial\pi}{\partial\omega}(x,\boj)-\chi_{1,j}\cdot\frac{\pi(x,\boj)}{(\tj)^2}\nonumber\\
        &+M_d\odot\Big(\frac{1}{\tj}\frac{\partial^2\pi}{\partial\omega\partial\omega^{\top}}(x,\boj)+\frac{1}{(\tj)^2}\frac{\partial\pi}{\partial\omega}(x,\boj)\cdot\frac{\partial\pi}{\partial\omega^{\top}}(x,\boj)\Big)\odot \phi_{2,j}\nonumber\\
        &+\Big(\frac{\pi(x,\boj)}{(\tj)^3}+\frac{\pi^2(x,\boj)}{2(\tj)^4}\Big)\chi_{2,j}-\zeta_{1,j}^{\top}\Big(\frac{1}{(\tj)^2}\frac{\partial\pi}{\partial\omega}(x,\boj)+\frac{\pi(x,\boj)}{(\tj)^3}\frac{\partial\pi}{\partial\omega}(x,\boj)\Big)\Big]\cdot H_j.
    \end{align}
    \item $C_{n}(x)$ in equation~\eqref{eq:general_Q_n_temperature}, we get
    \begin{align}
        \label{eq:general_limit_5_temperature}
        \frac{C_{n}(x)}{m_n\mathcal{L}_{2n}}\to\sum_{j=1}^{k_*}\xi_j[F_j-H_j].
    \end{align}
\end{itemize}
Note that from equation~\eqref{eq:general_zero_limit_temperature}, it follows that the sum of the limits in equations~\eqref{eq:general_limit_1_temperature}, \eqref{eq:general_limit_2_temperature}, \eqref{eq:general_limit_3_temperature}, \eqref{eq:general_limit_4_temperature} and \eqref{eq:general_limit_5_temperature} is equal to zero.

\vspace{0.5em}
\noindent
Subsequently, we aim to demonstrate that the values of $\phi^{(u)}_{1,j}$, $\phi^{(uu)}_{2,j}$, $\chi_{1,j}$, $\chi_{2,j}$, $\varphi^{(u)}_{1,j}$, $\varphi^{(uu)}_{2,j}$ and $\xi_j$ are all zero for $j\in[k_*]$. Indeed, let $P_1,P_2,\ldots,P_{\ell}$ be the partition of the set $\Big\{\exp\Big(\frac{\pi(x,\boj)}{\tj}\Big):j\in[k_*]\Big\}$, for some $\ell\leq k_*$, such that 
\begin{itemize}
    \item[(i)] $\omega^*_{j}=\omega^*_{j'}$ for any $j,j'\in P_i$ and $i\in[\ell]$;
    \item[(ii)] $\omega^*_{j}\neq\omega^*_{j'}$ when $j$ and $j'$ do not belong to the same set $P_i$ for any $i\in[\ell]$.
\end{itemize}
Then, the set $\Big\{\exp\Big(\frac{\pi(x,\omega^*_{j_1})}{\tau^*}\Big),\ldots,\exp\Big(\frac{\pi(x,\omega^*_{j_\ell})}{\tau^*}\Big)\Big\}$, where $P_i\in P_i$, is linearly independent. Since the limits in equations~\eqref{eq:general_limit_1_temperature}, \eqref{eq:general_limit_2_temperature}, \eqref{eq:general_limit_3_temperature}, \eqref{eq:general_limit_4_temperature} and \eqref{eq:general_limit_5_temperature} sum up to zero, we get for any $i\in[\ell]$ that
\begin{align*}
    &\sum_{j\in P_i:|\mathcal{A}_j|=1}\Big[\Big(\xi_j+\frac{(\phi_{1,j})^{\top}}{\tj}\frac{\partial\pi}{\partial\omega}(x,\boj)-\chi_{1,j}\cdot\frac{\pi(x,\boj)}{(\tj)^2}\Big)\cdot \mathcal{E}_j+\varphi_{1,j}^{\top} \mathcal{E}_{1j}\Big]\\
    &+\sum_{j\in P_i:|\mathcal{A}_j|>1}\Bigg\{\Big[\xi_{j}+\frac{(\phi_{1,j})^{\top}}{\tj}\frac{\partial\pi}{\partial\omega}(x,\boj)-\chi_{1,j}\cdot\frac{\pi(x,\boj)}{(\tj)^2}\nonumber\\
        &+M_d\odot\Big(\frac{1}{\tj}\frac{\partial^2\pi}{\partial\omega\partial\omega^{\top}}(x,\boj)+\frac{1}{(\tj)^2}\frac{\partial\pi}{\partial\omega}(x,\boj)\cdot\frac{\partial\pi}{\partial\omega^{\top}}(x,\boj)\Big)\odot \phi_{2,j}\nonumber\\
        &+\Big(\frac{\pi(x,\boj)}{(\tj)^3}+\frac{\pi^2(x,\boj)}{2(\tj)^4}\Big)\chi_{2,j}-\zeta_{1,j}^{\top}\Big(\frac{1}{(\tj)^2}\frac{\partial\pi}{\partial\omega}(x,\boj)+\frac{\pi(x,\boj)}{(\tj)^3}\frac{\partial\pi}{\partial\omega}(x,\boj)\Big)\Big]\cdot \mathcal{E}_j\nonumber\\
        &+[\varphi_{1,j}^{\top} \mathcal{E}_{1j}+\Big(\frac{1}{\tj}\mathcal{E}_{1j}\frac{\partial\pi}{\partial\omega^{\top}}(x,\boj)\Big)\odot\zeta_{2,j}-\frac{\pi(x,\boj)}{(\tj)^2}\cdot\zeta_{3,j}^{\top}\mathcal{E}_{1j}]+\Big(M_{d}\odot \varphi_{2,j}\Big)\odot \mathcal{E}_{2j}\Bigg\}\\
        &+\sum_{j:|\mathcal{A}_j|=1}\Big[\xi_{j}+\frac{\phi^{\top}_{1,j}}{\tj}\frac{\partial\pi}{\partial  \omega}(x,\boj)-\chi_{1,j}\cdot\frac{\pi(x,\boj)}{(\tj)^2}\Big]g_{G_*}(x)\\
        &+\sum_{j:|\mathcal{A}_j|>1}\Big[\xi_{j}+\frac{(\phi_{1,j})^{\top}}{\tj}\frac{\partial\pi}{\partial\omega}(x,\boj)-\chi_{1,j}\cdot\frac{\pi(x,\boj)}{(\tj)^2}\nonumber\\
        &+M_d\odot\Big(\frac{1}{\tj}\frac{\partial^2\pi}{\partial\omega\partial\omega^{\top}}(x,\boj)+\frac{1}{(\tj)^2}\frac{\partial\pi}{\partial\omega}(x,\boj)\cdot\frac{\partial\pi}{\partial\omega^{\top}}(x,\boj)\Big)\odot \phi_{2,j}\nonumber\\
        &+\Big(\frac{\pi(x,\boj)}{(\tj)^3}+\frac{\pi^2(x,\boj)}{2(\tj)^4}\Big)\chi_{2,j}-\zeta_{1,j}^{\top}\Big(\frac{1}{(\tj)^2}\frac{\partial\pi}{\partial\omega}(x,\boj)+\frac{\pi(x,\boj)}{(\tj)^3}\frac{\partial\pi}{\partial\omega}(x,\boj)\Big)\Big]g_{G_*}(x)=0,
\end{align*}
where we denote $\mathcal{E}_{j}:=\mathcal{E}(x,\ej)$, $\mathcal{E}_{1j}:=\frac{\partial \mathcal{E}}{\partial \eta}(x,\ej)$ and $\mathcal{E}_{2j}:=\frac{\partial^2 \mathcal{E}}{\partial\eta\partial\eta^{\top}}(x,\ej)$. Recall that the pair of the expert function $\mathcal{E}$ and the router $\pi$ satisfies the algebraic independence condition in Definition~\ref{def:distinguishability_condition}, then the following set is linearly independent for almost every $x$:
\begin{align*}
    \Big\{&\mathcal{E}_j,\pi(x,\boj)\mathcal{E}_j,\frac{\partial\pi}{\partial\omega^{(u)}}(x,\boj)\mathcal{E}_j,\frac{\partial\pi}{\partial\omega^{(u)}}(x,\boj)\frac{\partial\pi}{\partial\omega^{(v)}}(x,\boj)\mathcal{E}_j,\frac{\partial^2\pi}{\partial\omega^{(u)}\partial\omega^{(v)}}(x,\boj)\mathcal{E}_j,\\
    &\pi(x,\boj)\frac{\partial\pi}{\partial\omega^{(u)}}(x,\boj)\mathcal{E}_j, \mathcal{E}_{1j}, \pi(x,\boj)\mathcal{E}_{1j},\frac{\partial\pi}{\partial\omega^{(u)}}(x,\boj)\mathcal{E}_{1j},\mathcal{E}_{2j}:j\in[k_*], 1\leq u,v\leq d\Big\}.
\end{align*}
Thus, it follows that $\xi_j=0$, $\phi_{1,j}=\varphi_{1,j}=\zerod$ and $\phi_{2,j}=\varphi_{2,j}=\zeta_{j}=\mathbf{0}_{d\times d}$ for any $j\in P_i$ and $i\in[\ell]$. This contradicts the fact that not all the values of $\phi^{(u)}_{1,j}$, $\phi^{(uu)}_{2,j}$, $\chi_{1,j}$, $\chi_{2,j}$ $\varphi^{(u)}_{1,j}$, $\varphi^{(uu)}_{2,j}$ and $\xi_j$, for $j\in[k_*]$, are zero. Therefore, we obtain the local part~\eqref{eq:general_local_inequality_temperature}, that is,
\begin{align*}
    \lim_{\varepsilon\to0}\inf_{G\in\mathcal{G}_{k}(\Theta):\mathcal{L}_4(G,G_*)\leq\varepsilon}\normf{g_{G}-g_{G_*}}/\mathcal{L}_4(G,G_*)>0.
\end{align*}
Hence, the proof is completed.

\subsection{Proof of Theorem~\ref{theorem:hierarchical_general_experts}}
\label{appendix:hierarchical_general_experts}
\noindent
Our main goal is to demonstrate that
\begin{align}
    \label{eq:general_universal_inequality_over_softmax}
    \inf_{G\in\mathcal{G}_{k^*_1,k_2}(\Theta)}\normf{h_{G}-h_{G_*}}/\mathcal{L}_{5}(G,G_*)>0.
\end{align}
By streamlining the same arguments as in Appendix~\ref{appendix:general_experts}, it is sufficient to establish the local part of equation~\eqref{eq:general_universal_inequality_over_softmax} provided below as the global part of equation~\eqref{eq:general_universal_inequality_over_softmax} can be achieved similarly:
\begin{align}
    \label{eq:general_local_inequality_over_softmax}
    \lim_{\varepsilon\to0}\inf_{G\in\mathcal{G}_{k^*_1,k_2}(\Theta):\mathcal{L}_{5}(G,G_*)\leq\varepsilon}\normf{h_{G}-h_{G_*}}/\mathcal{L}_{5}(G,G_*)>0.
\end{align}
Assume by contrary that the above claim does not hold, then we can retrieve a sequence of mixing measures $(G_n)$ in $\mathcal{G}_{k^*_1,k_2}(\Theta)$ such that $\mathcal{L}_{5n}:=\mathcal{L}_{5}(G_n,G_*)\to0$ and
\begin{align}
    \label{eq:general_ratio_limit_over_softmax}
    \normf{h_{G_n}-h_{G_*}}/\mathcal{L}_{5n}\to0,
\end{align}
as $n\to\infty$. Since the Voronoi loss $\mathcal{A}^n_{j_1}:=\mathcal{A}_{j_1}(G_n)$ has only one element, we may assume WLOG that $\mathcal{A}_{j_1}=\{j_1\}$ for any $j_1\in[k^*_1]$. 
\begin{align}
&\mathcal{L}_{5n}=\sum_{j_1=1}^{k^*_1}\Big|\exp(\beta^n_{j_1})-\exp(\bzj)\Big|+\sum_{j_1=1}^{k^*_1}\exp(\beta_{j_1})\|\omega^n_{j_1}-\omega^*_{j_1}\|\nonumber\\
    &\hspace{2cm}+\sum_{j_1=1}^{k^*_1}\exp(\beta^n_{j_1})\Bigg[\sum_{j_2:|\mathcal{A}_{j_2|j_1}|=1}\exp(\nu^n_{j_2|j_1})(\|\kappa^n_{j_2|j_1}-\kappa^*_{j_2|j_1}\|+\|\eta^n_{j_1j_2}-\eta^*_{j_1j_2}\|)\nonumber\\
    &\hspace{4cm}+\sum_{j_2:|\mathcal{A}_{j_2|j_1}|>1}\sum_{i_2\in\mathcal{A}_{j_2|j_1}}\exp(\nu^n_{j_2|j_1})(\|\kappa^n_{i_2|j_1}-\kappa^*_{j_2|j_1}\|^{2}+\|\eta^n_{j_1i_2}-\eta^*_{j_1j_2}\|^{2})\Bigg]\nonumber\\
    \label{eq:loss_l1_softmax}
    &\hspace{2cm}+\sum_{j_1=1}^{k^*_1}\exp(\beta_{j_1})\sum_{j_2=1}^{k^*_2}\Big|\sum_{i_2\in\mathcal{A}_{j_2|j_1}}\exp(\nu^n_{i_2|j_1})-\exp(\vj)\Big|,
\end{align}
Since $\mathcal{L}_{5n}\to0$ as $n\to\infty$, we deduce that $\bojn\to\boj$ and $\exp(\bzjn)\to\exp(\bzj)$ as $n\to\infty$.

\vspace{0.5em}
\noindent
\textbf{Step 1 - Decompose the difference between regression functions:} In this step, we utilize the Taylor expansion to decompose the following term:
\begin{align*}
    Q_n(x):=\left[\sum_{j_1=1}^{k^*_1}\exp((\hboj)^{\top}x+\hbzj)\right][h_{G_n}(x)-h_{G_*}(x)].
\end{align*}
Let us denote
\begin{align*}
    h^n_{j_1}(x)&:=\sum_{j_2=1}^{k^*_2}\sum_{i_2\in\mathcal{A}_{j_2|j_1}}\softmax((\kappa^n_{i_2|j_1})^{\top}x+\nu^n_{i_2|j_1}) \mathcal{E}(x,\eta^n_{j_1i_2}),\\
    h^*_{j_1}(x)&:=\sum_{j_2=1}^{k^*_2}\softmax((\kj)^{\top}x+\vj) \mathcal{E}(x,\hej),
\end{align*}
then the term $Q_n(x)$ can be decomposed into three terms as
\begin{align}
    Q_n(x)&=\sum_{j_1=1}^{k^*_1}\exp(\hbzjn)\left[\exp((\hbojn)^{\top}x)h^n_{j_1}(x)-\exp((\hboj)^{\top}x)h^*_{j_1}(x)\right]\nonumber\\
    &-\sum_{j_1=1}^{k^*_1}\exp(\hbzjn)\left[\exp((\hbojn)^{\top}x)-\exp((\hboj)^{\top}x)\right]h_{G_n}(x)\nonumber\\
    &+\sum_{j_1=1}^{k^*_1}\left(\exp(\hbzjn)-\exp(\hbzj)\right)\exp((\hboj)^{\top}x)\left[h^*_{j_1}(x)-h_{G_n}(x)\right]\nonumber\\
     \label{eq:decompose_Qn_softmax}
    :&=A_n(x)-B_n(x)+C_n(x).
\end{align}
\textbf{Step 1A - Decompose $A_n(x)$:} We proceed to decompose this term as follows:
\begin{align*}
    A_n(x)&:=\sum_{j_1=1}^{k^*_1}\frac{\exp(\hbzjn)}{\sum_{j'_2=1}^{k^*_2}\exp((\kappa^*_{j'_2|j_1})^{\top}x+\nu^*_{j'_2|j_1})}[A_{n,j_1,1}(x)-A_{n,j_1,2}(x)+A_{n,j_1,3}(x)],
\end{align*}
where we define
\begin{align*}
    A_{n,j_1,1}(x)&:=\sum_{j_2=1}^{k^*_2}\sum_{i_2\in\mathcal{A}_{j_2|j_1}}\exp(\nu^n_{i_2|j_1})\Big[\exp((\kin)^{\top}x)\exp((\hbojn)^{\top}x)\mathcal{E}(x,\ein)\nonumber\\
    &\hspace{4cm}-\exp((\kj)^{\top}x)\exp((\hboj)^{\top}x)\mathcal{E}(x,\hej)\Big],\\
    A_{n,j_1,2}(x)&:=\sum_{j_2=1}^{k^*_2}\sum_{i_2\in\mathcal{A}_{j_2|j_1}}\exp(\nu^n_{i_2|j_1})\Big[\exp((\kin)^{\top}x)\nonumber-\exp((\kj)^{\top}x)\Big]\exp((\hbojn)^{\top}x)h^n_{j_1}(x),\\
    A_{n,j_1,3}(x)&:=\sum_{j_2=1}^{k^*_2}\Big(\sum_{i_2\in\mathcal{A}_{j_2|j_1}}\exp(\vin)-\exp(\vj)\Big)\exp((\kj)^{\top}x)\\
    &\hspace{4cm}\times[\exp((\hboj)^{\top}x)\mathcal{E}(x,\hej)-\exp((\hbojn)^{\top}x)h^n_{j_1}(x)].
\end{align*}
\textbf{Step 1A.1 - Decompose $A_{n,j_1,1}(x)$:}
Moreover, we continue to divide the term $A_{n,j_1,1}(x)$ into two parts based on the cardinality of the Voronoi cells $\mathcal{A}_{j_2|j_1}$:
\begin{align*}
    A_{n,j_1,1}(x)&=\sum_{j_2:|\mathcal{A}_{j_2|j_1}|=1}\sum_{i_2\in\mathcal{A}_{j_2|j_1}}\exp(\vin)\Big[\exp((\kin)^{\top}x)\exp((\hbojn)^{\top}x)\mathcal{E}(x,\ein)\\
    &\hspace{5cm}-\exp((\kj)^{\top}x)\exp((\hboj)^{\top}x)\mathcal{E}(x,\hej)\Big],\\
    &+\sum_{j_2:|\mathcal{A}_{j_2|j_1}|>1}\sum_{i_2\in\mathcal{A}_{j_2|j_1}}\exp(\vin)\Big[\exp((\kin)^{\top}x)\exp((\hbojn)^{\top}x)\mathcal{E}(x,\ein)\\
    &\hspace{5cm}-\exp((\kj)^{\top}x)\exp((\hboj)^{\top}x)\mathcal{E}(x,\hej)\Big]\\
    :&=A_{n,j_1,1,1}(x)+A_{n,j_1,1,2}(x).
\end{align*}
By employing the first-order Taylor expansion to the term $A_{n,j_1,1,1}(x)$, we get
\begin{align*}
    A_{n,j_1,1,1}(x)&=\sum_{j_2:|\mathcal{A}_{j_2|j_1}|=1}\sum_{i_2\in\mathcal{A}_{j_2|j_1}}\sum_{|\alpha|=1}\frac{\exp(\vin)}{\alpha!}(\dkijn)^{\alpha_1}(\dbojn)^{\alpha_2}(\deijn)^{\alpha_3}\\
    &\hspace{0cm}\times x^{\alpha_1+\alpha_2}\exp((\kj)^{\top}x)\exp((\hboj)^{\top}x)\frac{\partial^{|\alpha_3|}\mathcal{E}}{\partial\eta^{\alpha_3}}(x,\hej)
    +R_{n,1,1}(x),\\
    &=\sum_{j_2:|\mathcal{A}_{j_2|j_1}|=1}\sum_{|\rho_1|+|\alpha_3|=1}S_{n,j_2|j_1,\rho_1,\alpha_3}\cdot x^{\rho_1}\exp((\kj)^{\top}x)\exp((\hboj)^{\top}x) \frac{\partial^{|\alpha_3|}\mathcal{E}}{\partial\eta^{\alpha_3}}(x,\hej)+R_{n,1,1}(x),
\end{align*}
where $R_{n,1,1}(x)$ is a Taylor remainder such that $R_{n,1,1}(x)/\mathcal{L}_{5n}\to0$ as $n\to\infty$, and 
\begin{align*}
    S_{n,j_2|j_1,\rho_1,\alpha_3}:=\sum_{i_2\in\mathcal{A}_{j_2|j_1}}\sum_{\alpha_1+\alpha_2=\rho_1}\frac{\exp(\vin)}{\alpha!}(\dkijn)^{\alpha_1}(\dbojn)^{\alpha_2}(\deijn)^{\alpha_3},
\end{align*}
for any $j_2\in[k^*_2]$ and $(\rho_1,\alpha_3)\neq(\zerod,\zerod)$.

\vspace{0.5em}
\noindent
Similarly, we apply the second-order Taylor expansion to the term $A_{n,j,1,2}(x)$ and get that
\begin{align*}
    A_{n,j,1,2}(x)&=\sum_{j_2:|\mathcal{A}_{j_2|j_1}|>1}\sum_{|\rho_1|+|\alpha_3|=1}^{2}S_{n,j_2|j_1,\rho_1,\alpha_3}\cdot x^{\rho_1}\exp((\kj)^{\top}x)\exp((\hboj)^{\top}x) \frac{\partial^{|\alpha_3|}\mathcal{E}}{\partial\eta^{\alpha_3}}(x,\hej)+R_{n,1,2}(x),
\end{align*}
where $R_{n,1,2}(x)$ is a Taylor remainder such that $R_{n,1,2}(x)/\mathcal{L}_{5n}\to0$ as $n\to\infty$.

\vspace{0.5em}
\noindent
\textbf{Step 1A.2 - Decompose $A_{n,j_1,2}(x)$:} Next, we also separate the quantity $A_{n,j_1,2}(x)$ into two following terms:
\begin{align*}
    A_{n,j_1,2}(x)&=\sum_{j_2:|\mathcal{A}_{j_2|j_1}|=1}\sum_{i_2\in\mathcal{A}_{j_2|j_1}}\exp(\nu^n_{i_2|j_1})\Big[\exp((\kin)^{\top}x)\nonumber-\exp((\kj)^{\top}x)\Big]\exp((\hbojn)^{\top}x)h^n_{j_1}(x)\\
    &+\sum_{j_2:|\mathcal{A}_{j_2|j_1}|>1}\sum_{i_2\in\mathcal{A}_{j_2|j_1}}\exp(\nu^n_{i_2|j_1})\Big[\exp((\kin)^{\top}x)\nonumber-\exp((\kj)^{\top}x)\Big]\exp((\hbojn)^{\top}x)h^n_{j_1}(x)\\
    &:=A_{n,j_1,2,1}(x)+A_{n,j_1,2,2}(x).
\end{align*}
By means of the first-order Taylor expansion, we have
\begin{align*}
    A_{n,j_1,2,1}(x)&=\sum_{j_2:|\mathcal{A}_{j_2|j_1}|=1}\sum_{i_2\in\mathcal{A}_{j_2|j_1}}\sum_{|\rho_2|=1}\frac{\exp(\vin)}{\rho_2!}\cdot(\dkijn)^{\rho_2}x^{\rho_2}\exp((\kj)^{\top}x) \exp((\hbojn)^{\top}x) h^n_{j_1}(x)+R_{n,2,1}(x),\\
    &=\sum_{j_2:|\mathcal{A}_{j_2|j_1}|=1}\sum_{|\rho_2|=1}T_{n,j_2|j_1,\rho_2}\cdot x^{\rho_2}\exp((\kj)^{\top}x)\exp((\hbojn)^{\top}x) h^n_{j_1}(x)+R_{n,2,1}(x),
\end{align*}
where $R_{n,2,1}(x)$ is a Taylor remainder such that $R_{n,2,1}(x)/\mathcal{L}_{5n}\to0$ as $n\to\infty$ and
\begin{align*}
    T_{n,j_2|j_1,\rho_2}:=\sum_{i_2\in\mathcal{A}_{j_2|j_1}}\frac{\exp(\vin)}{\rho_2!}\cdot(\dkijn)^{\rho_2},
\end{align*}
for any $j_2\in[k^*_2]$ and $\rho_2\neq\zerod$.

\vspace{0.5em}
\noindent
Analogously, by means of the second-order Taylor expansion, we have
\begin{align*}
    A_{n,j_1,2,1}(x)=\sum_{j_2:|\mathcal{A}_{j_2|j_1}|>1}\sum_{|\rho_2|=1}^{2}T_{n,j_2|j_1,\rho_2}\cdot x^{\rho_2}\exp((\kj)^{\top}x)\exp((\hbojn)^{\top}x) h^n_{j_1}(x)+R_{n,2,2}(x),
\end{align*}
where $R_{n,2,2}(x)$ is a Taylor remainder such that $R_{n,2,2}(x)/\mathcal{L}_{5n}\to0$ as $n\to\infty$.

\vspace{0.5em}
\noindent
Combining the above decompositions of $A_{n,j_1,1}(x)$ and $A_{n,j_1,2}(x)$, we deduce that 
\begin{align}
    A_n(x)&=\sum_{j_1=1}^{k^*_1}\sum_{j_2=1}^{k^*_2}\frac{\exp(\hbzjn)}{\sum_{j'_2=1}^{k^*_2}\exp((\kappa^*_{j'_2|j_1})^{\top}x+\nu^*_{j'_2|j_1})}[R_{n,1,1}(x)+R_{n,1,2}(x)-R_{n,2,1}(x)-R_{n,2,2}(x)\nonumber\\
    &+\sum_{|\rho_1|+|\alpha_3|=0}^{2}S_{n,j_2|j_1,\rho_1,\alpha_3}\cdot x^{\rho_1}\exp((\kj)^{\top}x)\exp((\hboj)^{\top}x) \frac{\partial^{|\alpha_3|}\mathcal{E}}{\partial\eta^{\alpha_3}}(x,\hej)\nonumber\\
    \label{eq:hAn_decomposition}
    &-\sum_{|\rho_2|=0}^{2}T_{n,j_2|j_1,\rho_2}\cdot x^{\rho_2}\exp((\kj)^{\top}x)\exp((\hbojn)^{\top}x) h^n_{j_1}(x)\Bigg],
\end{align}
where we define $S_{n,j_2|j_1,\rho_1,\alpha_3}=T_{n,j_2|j_1,\rho_2}=\sum_{i_2\in\mathcal{A}_{j_2|j_1}}\exp(\vin)-\exp(\vj)$ for any $j_1\in[k^*_1]$, $j_2\in[k^*_2]$ and $(\rho_1,\alpha_3,\rho_2)=(\zerod,\zeroq,\zerod)$.

\vspace{0.5em}
\noindent
\textbf{Step 1B - Decompose $B_n(x)$:} Subsequently, by invoking the Taylor expansion of first order, we get
\begin{align}
    \label{eq:hBn_decomposition}
    B_n(x)=\sum_{j_1=1}^{k^*_1}\exp(\beta^n_{j_1})\sum_{|\rho_3|=1}(\dbojn)^{\rho_3}\cdot x^{\rho_3}\exp((\hboj)^{\top}x)h_{G_n}(x)+R_{n,3}(x),
\end{align}
where $R_{n,3}(x)$ is a Taylor remainder such that $R_{n,3}(x)/\mathcal{L}_{5n}\to0$ as $n\to\infty$.

\vspace{0.5em}
\noindent
From the decompositions in equations~\eqref{eq:hAn_decomposition}, \eqref{eq:hBn_decomposition} and \eqref{eq:decompose_Qn_softmax}, we can view $A_n(x)$, $B_n(x)$ and $C_n(x)$ as a combination of elements from the following set:
\begin{align*}
    &\Bigg\{\frac{x^{\rho_1}\exp((\kj)^{\top}x)\exp((\hboj)^{\top}x) \frac{\partial^{|\alpha_3|}\mathcal{E}}{\partial\eta^{\alpha_3}}(x,\hej)}{\sum_{j'_2=1}^{k^*_2}\exp((\kappa^*_{j'_2|j_1})^{\top}x+\nu^*_{j'_2|j_1})}:j_1\in[k^*_1],j_2\in[k^*_2],0\leq |\rho_1|+|\alpha_3|\leq 2\Bigg\}\\
    &\cup\Bigg\{\frac{x^{\rho_2}\exp((\kj)^{\top}x)\exp((\hbojn)^{\top}x) h^n_{j_1}(x)}{\sum_{j'_2=1}^{k^*_2}\exp((\kappa^*_{j'_2|j_1})^{\top}x+\nu^*_{j'_2|j_1})}:j_1\in[k^*_1],j_2\in[k^*_2],0\leq|\rho_2|\leq2\Bigg\}\\
    &\cup\Big\{x^{\rho_3}\exp((\hboj)^{\top}x)h_{G_n}(x):j_1\in[k^*_1], 0\leq|\rho_3|\leq 1\Big\}\cup\Big\{\exp((\hboj)^{\top}x)h^*_{j_1}(x):j_1\in[k^*_1]\Big\}.
\end{align*}
\textbf{Step 2 - Non-vanishing coefficients:} In this step, we aim to demonstrate that not all the coefficients in the representations of $A_n(x)/\mathcal{L}_{5n}$, $B_n(x)/\mathcal{L}_{5n}$ and $C_n(x)/\mathcal{L}_{5n}$ converge to zero as $n$ tends to infinity. Assume by contrary that all of them go to zero, then by considering the coefficients of the following terms
\begin{itemize}
    \item $\exp((\hboj)^{\top}x)h^*_{j_1}(x)$ in $C_n(x)/\mathcal{L}_{5n}$, we get
    \begin{align*}
        \frac{1}{\mathcal{L}_{5n}}\cdot\sum_{j_1=1}^{k^*_1}\Big|\exp(\hbzjn)-\exp(\hbzj)\Big|\to0;
    \end{align*}
    \item $x^{\rho_3}\exp((\hboj)^{\top}x)h_{G_n}(x)$ in $B_n(x)/\mathcal{L}_{5n}$ for $\rho_3=e_{d,u}:=(0,\ldots,0,\underbrace{1}_{\textit{u-th}},0,\ldots,0)\in\mathbb{R}^d$, we have
    \begin{align*}
        \frac{1}{\mathcal{L}_{5n}}\cdot\sum_{j_1=1}^{k^*_1}\exp(\hbzjn)\|\hbojn-\hboj\|\to0;
    \end{align*}
    \item $\frac{x^{\rho_2}\exp((\kj)^{\top}x)\exp((\hbojn)^{\top}x) h^n_{j_1}(x)}{\sum_{j'_2=1}^{k^*_2}\exp((\kappa^*_{j'_2|j_1})^{\top}x+\nu^*_{j'_2|j_1})}$ in $A_n(x)/\mathcal{L}_{5n}$ for $j_1\in[k^*_1]$, $j_2\in[k^*_2]:|\mathcal{A}_{j_2|j_1}|=1$ and $\rho_2=e_{d,u}$, we get
    \begin{align*}
        \frac{1}{\mathcal{L}_{5n}}\cdot\sum_{j_1=1}^{k^*_1}\exp(\hbzjn)\sum_{j_2\in[k^*_2]:|\mathcal{A}_{j_2|j_1}|=1}\sum_{i_2\in\mathcal{A}_{j_2|j_1}}\exp(\vin)\|\kin-\kj\|_1\to0;
    \end{align*}
    Due to the equivalency between the $\ell_1$-norm and the $\ell_2$-norm, we deduce that
    \begin{align*}
        \frac{1}{\mathcal{L}_{5n}}\cdot\sum_{j_1=1}^{k^*_1}\exp(\hbzjn)\sum_{j_2\in[k^*_2]:|\mathcal{A}_{j_2|j_1}|=1}\sum_{i_2\in\mathcal{A}_{j_2|j_1}}\exp(\vin)\|\kin-\kj\|\to0;
    \end{align*}
    \item $\frac{\exp((\kj)^{\top}x)\exp((\hboj)^{\top}x) \frac{\partial^{|\alpha_3|}\mathcal{E}}{\partial\eta^{\alpha_3}}(x,\hej)}{\sum_{j'_2=1}^{k^*_2}\exp((\kappa^*_{j'_2|j_1})^{\top}x+\nu^*_{j'_2|j_1})}$ in $A_n(x)/\mathcal{L}_{5n}$ for $j_1\in[k^*_1]$, $j_2\in[k^*_2]:|\mathcal{A}_{j_2|j_1}|=1$ and $\alpha_3=e_{q,u}\in\mathbb{R}^q$, we get
    \begin{align*}
        \frac{1}{\mathcal{L}_{5n}}\cdot\sum_{j_1=1}^{k^*_1}\exp(\hbzjn)\sum_{j_2\in[k^*_2]:|\mathcal{A}_{j_2|j_1}|=1}\sum_{i_2\in\mathcal{A}_{j_2|j_1}}\exp(\vin)\|\hein-\hej\|\to0;
    \end{align*}
    \item $\frac{x^{\rho_2}\exp((\kj)^{\top}x)\exp((\hbojn)^{\top}x) h^n_{j_1}(x)}{\sum_{j'_2=1}^{k^*_2}\exp((\kappa^*_{j'_2|j_1})^{\top}x+\nu^*_{j'_2|j_1})}$ in $A_n(x)/\mathcal{L}_{5n}$ for $j_1\in[k^*_1]$, $j_2\in[k^*_2]:|\mathcal{A}_{j_2|j_1}|>1$ and $\rho_2=2e_{d,u}$, we get
    \begin{align*}
        \frac{1}{\mathcal{L}_{5n}}\cdot\sum_{j_1=1}^{k^*_1}\exp(\hbzjn)\sum_{j_2\in[k^*_2]:|\mathcal{A}_{j_2|j_1}|>1}\sum_{i_2\in\mathcal{A}_{j_2|j_1}}\exp(\vin)\|\kin-\kj\|^2\to0;
    \end{align*}
    \item $\frac{\exp((\kj)^{\top}x)\exp((\hboj)^{\top}x) \frac{\partial^{|\alpha_3|}\mathcal{E}}{\partial\eta^{\alpha_3}}(x,\hej)}{\sum_{j'_2=1}^{k^*_2}\exp((\kappa^*_{j'_2|j_1})^{\top}x+\nu^*_{j'_2|j_1})}$ in $A_n(x)/\mathcal{L}_{5n}$ for $j_1\in[k^*_1]$, $j_2\in[k^*_2]:|\mathcal{A}_{j_2|j_1}|>1$ and $\alpha_3=2e_{q,u}\in\mathbb{R}^q$, we get
    \begin{align*}
        \frac{1}{\mathcal{L}_{5n}}\cdot\sum_{j_1=1}^{k^*_1}\exp(\hbzjn)\sum_{j_2\in[k^*_2]:|\mathcal{A}_{j_2|j_1}|>1}\sum_{i_2\in\mathcal{A}_{j_2|j_1}}\exp(\vin)\|\hein-\hej\|^2\to0;
    \end{align*}
    \item $\frac{\exp((\kj)^{\top}x)\exp((\hboj)^{\top}x) \mathcal{E}(x,\hej)}{\sum_{j'_2=1}^{k^*_2}\exp((\kappa^*_{j'_2|j_1})^{\top}x+\nu^*_{j'_2|j_1})}$ in $A_n(x)/\mathcal{L}_{5n}$, we get
    \begin{align*}
        \frac{1}{\mathcal{L}_{5n}}\cdot\sum_{j_1=1}^{k^*_1}\exp(\hbzjn)\sum_{j_2=1}^{k^*_2}\Big|\sum_{i_2\in\mathcal{A}_{j_2|j_1}}\exp(\vin)-\exp(\vj)\Big|\to0;
    \end{align*} 
\end{itemize}
By taking the summation of the above limits, then it follows from the formulation of the loss $\mathcal{L}_{5n}$ in equation~\eqref{eq:loss_l1_softmax} that $1=\mathcal{L}_{5n}/\mathcal{L}_{5n}$, which is a contradiction. Hence, we obtain at least one among the coefficients in the representations of $A_n(x)/\mathcal{L}_{5n}$, $B_n(x)/\mathcal{L}_{5n}$ and $C_n(x)/\mathcal{L}_{5n}$ does not go to zero as $n\to\infty$.

\vspace{0.5em}
\noindent
\textbf{Step 3 - Application of the Fatou's lemma:} In this step, we point out that all the coefficients in the representations of $A_n(x)/\mathcal{L}_{5n}$, $B_n(x)/\mathcal{L}_{5n}$ and $C_n(x)/\mathcal{L}_{5n}$ converge to zero as $n\to\infty$. In particular, let us denote by $m_n$ the maximum of the absolute values of those coefficients. From the result achieved in Step 2, we deduce that $1/m_n\not\to\infty$. Recall from the hypothesis in equation~\eqref{eq:general_ratio_limit_over_softmax} that $\normf{h_{G_n}-h_{G_*}}/\mathcal{L}_{5n}\to0$ as $n\to\infty$, which indicates that $\|h_{G_n}-h_{G_*}\|_{L^1(\mu)}/\mathcal{L}_{5n}\to0$. Then, according to the Fatou's lemma, we get that
\begin{align*}
    0=\lim_{n\to\infty}\frac{\|h_{G_n}-h_{G_*}\|_{L^1(\mu)}}{m_n\mathcal{L}_{5n}}\geq \int \liminf_{n\to\infty}\frac{|h_{G_n}(x)-h_{G_*}(x)|}{m_n\mathcal{L}_{5n}}\dint\mu(x)\geq 0.
\end{align*}
As a result, it follows that $\frac{1}{m_n\mathcal{L}_{5n}}\cdot[h_{G_n}(x)-h_{G_*}(x)]\to0$ as $n\to\infty$ for $\mu$-almost surely $x$. From the formulation of $Q_n(x)$ in equation~\eqref{eq:decompose_Qn_softmax}, since the quantity $\sum_{j_1=1}^{K}\exp(-\|\hboj-x\|+\hbzj)$ is bounded, we get $\frac{1}{m_n\mathcal{L}_{5n}}\cdot Q_n(x)\to0$ for $\mu$-almost surely $x$.\\

\noindent
Let us denote
\begin{align*}
    \frac{\exp(\hbzjn)S_{n,j_2|j_1,\rho_1,\alpha_3}}{m_n\mathcal{L}_{5n}}\to\phi_{j_2|j_1,\rho_1,\alpha_3}&, \hspace{3cm} \frac{\exp(\hbzjn)T_{n,j_2|j_1,\rho_2}}{m_n\mathcal{L}_{5n}}\to\varphi_{j_2|j_1,\rho_2},\\
    \frac{\exp(\hbzjn)(\Delta\omega^n_{j_1})^{\rho_3}}{m_n\mathcal{L}_{5n}}\to\lambda_{j_1,\rho_3}&, \hspace{3cm} \frac{\exp(\hbzjn)-\exp(\hbzj)}{m_n\mathcal{L}_{5n}}\to\chi_{j_1},
\end{align*}
for all $j_1\in[k^*_1]$, $j_2\in[k^*_2]$, $0\leq|\rho_1|+|\alpha_3|\leq 2$, $0\leq|\rho_2|\leq 2$, $|\rho_3|=1$ with a note that at least one among them is different from zero. 
From the decomposition of $Q_n(x)$ in equation~\eqref{eq:decompose_Qn_softmax}, we have
\begin{align*}
    \lim_{n\to\infty}\frac{Q_n(x)}{m_n\mathcal{L}_{5n}}=\lim_{n\to\infty}\frac{A_n(x)}{m_n\mathcal{L}_{5n}}-\lim_{n\to\infty}\frac{B_n(x)}{m_n\mathcal{L}_{5n}}+\lim_{n\to\infty}\frac{C_n(x)}{m_n\mathcal{L}_{5n}},
\end{align*}
where 
\begin{align*}
    \lim_{n\to\infty}\frac{A_n(x)}{m_n\mathcal{L}_{5n}}&=\sum_{j_1=1}^{k^*_1}\sum_{j_2=1}^{k^*_2}\Bigg[\sum_{|\rho_1|+|\alpha_3|=0}^{2}\phi_{j_2|j_1,\rho_1,\alpha_3}\cdot x^{\rho_1}\exp((\kj)^{\top}x)\exp((\hboj)^{\top}x) \frac{\partial^{|\alpha_3|}\mathcal{E}}{\partial\eta^{\alpha_3}}(x,\hej)\\
    &\hspace{-1.2cm}-\sum_{|\rho_2|=0}^{2}\varphi_{j_2|j_1,\rho_2}\cdot x^{\rho_2}\exp((\kj)^{\top}x)\exp((\hboj)^{\top}x) h^*_{j_1}(x)\Bigg]\cdot\frac{1}{\sum_{j'_2=1}^{k^*_2}\exp((\kappa^*_{j'_2|j_1})^{\top}x+\nu^*_{j'_2|j_1})},\\
    \lim_{n\to\infty}\frac{B_n(x)}{m_n\mathcal{L}_{5n}}&=\sum_{j_1=1}^{k^*_1}\sum_{|\rho_3|=1}\lambda_{j_1,\rho_3}\cdot x^{\rho_3}\exp((\hboj)^{\top}x)h_{G_*}(x),\\
    \lim_{n\to\infty}\frac{C_n(x)}{m_n\mathcal{L}_{5n}}&=\sum_{j_1=1}^{k^*_1}\chi_{j_1}\exp((\hboj)^{\top}x)\left[h^*_{j_1}(x)-h_{G_*}(x)\right].
\end{align*}
Since the expert function $x\mapsto\mathcal{E}(x,\eta)$ meets the strong identifiability condition in Definition~\ref{def:softmax_condition}, the following set of functions in $x$ is linearly independent:
\begin{align*}
    &\Bigg\{\frac{x^{\rho_1}\exp((\kj)^{\top}x)\exp((\hboj)^{\top}x) \frac{\partial^{|\alpha_3|}\mathcal{E}}{\partial\eta^{\alpha_3}}(x,\hej)}{\sum_{j'_2=1}^{k^*_2}\exp((\kappa^*_{j'_2|j_1})^{\top}x+\nu^*_{j'_2|j_1})}:j_1\in[k^*_1],j_2\in[k^*_2],0\leq |\rho_1|+|\alpha_3|\leq 2\Bigg\}\\
    &\cup\Bigg\{\frac{x^{\rho_2}\exp((\kj)^{\top}x)\exp((\hboj)^{\top}x) h^*_{j_1}(x)}{\sum_{j'_2=1}^{k^*_2}\exp((\kappa^*_{j'_2|j_1})^{\top}x+\nu^*_{j'_2|j_1})}:j_1\in[k^*_1],j_2\in[k^*_2],0\leq|\rho_2|\leq2\Bigg\}\\
    &\cup\Big\{x^{\rho_3}\exp((\hboj)^{\top}x)h_{G_*}(x):j_1\in[k^*_1], 0\leq|\rho_3|\leq 1\Big\}\cup\Big\{\exp((\hboj)^{\top}x)h^*_{j_1}(x):j_1\in[k^*_1]\Big\}.
\end{align*}
As a result, we get that $\phi_{j_2|j_1,\rho_1,\alpha_3}=\varphi_{j_2|j_1,\rho_2}=\lambda_{j_1,\rho_3}=\chi_{j_1}$ for all all $j_1\in[k^*_1]$, $j_2\in[k^*_2]$, $0\leq|\rho_1|+|\alpha_3|\leq 2$, $0\leq|\rho_2|\leq 2$, $|\rho_3|=0$, which contradicts to the fact that at least one of them is non-zero. Thus, we achieve the local part~\eqref{eq:general_local_inequality_over_softmax}, that is,
\begin{align*}
    \lim_{\varepsilon\to0}\inf_{G\in\mathcal{G}_{k^*_1,k_2}(\Theta):\mathcal{L}_{5}(G,G_*)\leq\varepsilon}\normf{h_{G}-h_{G_*}}/\mathcal{L}_{5}(G,G_*)>0,
\end{align*}
Hence, the proof is completed.

\subsection{Proof of Theorem~\ref{theorem:hierarchical_linear_experts}}
\label{appendix:hierarchical_linear_experts}
\noindent
Firstly, we will demonstrate that the limit 
\begin{align}
    \label{eq:ratio_zero_limit_hierarchical}
    \lim_{\varepsilon\to0}\inf_{G\in\mathcal{G}_{k^*_1k_2}(\Theta):\mathcal{L}_{6,r}(G,G_*)\leq\varepsilon}\frac{\normf{h_{G}-h_{G_*}}}{\mathcal{L}_{6,r}(G,G_*)}=0
\end{align}
holds true for any $r\geq 1$. Then, by employing the same arguments for proving the equation~\eqref{eq:minimax_activation_experts} in Appendix~\ref{appendix:linear_experts}, we arrive at our desired result
\begin{align*}
        \inf_{\overline{G}_n\in\mathcal{G}_{k^*_1k_2}(\Theta)}\sup_{G\in\mathcal{G}_{k^*_1k_2}(\Theta)\setminus\mathcal{G}_{k^*_1(k^*_2-1)}(\Theta)}\bbE_{h_{G}}[\mathcal{L}_{6,r}(\overline{G}_n,G)]\gtrsim n^{-1/2}.
    \end{align*}
\textbf{Proof for equation~\eqref{eq:ratio_zero_limit_hierarchical}:} We need to construct a sequence of mixing measures $(G_n)\subset\mathcal{G}_{k^*_1k_2}(\Theta)$ satisfying $\mathcal{L}_{6,r}(G_n,G_*)\to0$ and 
\begin{align*}
    \normf{h_{G_n}-h_{G_*}}/\mathcal{L}_{6,r}(G_n,G_*)\to0,
\end{align*}
as $n\to\infty$. For that purpose, let us take into account the mixing measure sequence defined as
\begin{align*}
    G_n=\sum_{i_1=1}^{k^*_1}\exp(\beta^n_{i_1})\sum_{i_2=1}^{k^*_2+1}\exp(\nu^n_{i_2|i_1})\delta_{(\omega^n_{j_1},\kappa^n_{i_2|i_1},a^n_{i_1i_2},b^n_{i_1i_2})},
\end{align*}
where 

\vspace{0.5em}
$\exp(\beta^n_{i_1})=\exp(\beta^*_{i_1})$, for any $1\leq i_1\leq k^*_1$;

\vspace{0.5em}
$\omega^n_{i_1}=\omega^*_{i_1}$, for any $1\leq i_1\leq k^*_1$;

\vspace{0.5em}
$\exp(\nu^n_{1|i_1})=\exp(\nu^n_{2|i_1})=\frac{1}{2}\exp(\nu^*_{1|i_1})+\frac{1}{2n^{r+1}}$ and  $\exp(\nu^n_{i_2|i_1})=\exp(\nu^n_{i_2-1|i_1})$, for any $1\leq i_1\leq k^*_1$ and $3\leq i_2\leq k^*_2+1$;

\vspace{0.5em}
$\kappa^n_{1|i_1}=\kappa^n_{2|i_1}=\kappa^*_{1|i_1}$ and  $\kappa^n_{i_2|i_1}=\kappa^n_{i_2-1|i_1}$, for any $1\leq i_1\leq k^*_1$ and $3\leq i_2\leq k^*_2+1$;

\vspace{0.5em}
$a^n_{i_11}=a^n_{i_12}=a^*_{i_11}$ and $a^n_{i_1i_2}=a^n_{i_1(i_2-1)}$, for any $1\leq i_1\leq k^*_1$ and $3\leq i_2\leq k^*_2+1$;

\vspace{0.5em}
$b^n_{i_11}=b^*_{i_11}+\frac{1}{n}$, $b^n_{i_12}=b^*_{i_11}-\frac{1}{n}$ and  $b^n_{i_1i_2}=b^*_{i_1(i_2-1)}$, for any $1\leq i_1\leq k^*_1$ and $3\leq i_2\leq k^*_2+1$.

\vspace{0.5em}
\noindent
Then, the Voronoi loss $\mathcal{L}_{6,r}(G_n,G_*)$ can be rewritten as
\begin{align}
    \label{eq:D_6r_formulation}
    \mathcal{L}_{6,r}(G_n,G_*)=\frac{1}{n^{r+1}}\sum_{j_1=1}^{k^*_1}\exp(\beta^*_{j_1})+\sum_{j_1=1}^{k^*_1}\exp(\beta^*_{j_1})\Big(\exp(\nu^*_{1|j_1})+\frac{1}{n^{r+1}}\Big)\cdot\frac{1}{n^r}=\mathcal{O}(n^{-r}).
\end{align}
From the above equation, it is clear that $\mathcal{L}_{6,r}(G_n,G_*)\to0$ as $n\to\infty$. Thus, it suffices to show that
\begin{align*}
    \normf{h_{G_n}-h_{G_*}}/\mathcal{L}_{6,r}(G_n,G_*)\to0.
\end{align*}
Let us consider the quantity $Q_n(x):=\left[\sum_{j_1=1}^{k^*_1}\exp((\hboj)^{\top}x+\hbzj)\right][h_{G_n}(x)-h_{G_*}(x)]$, which can be decomposed as follows:
\begin{align*}
    Q_n(x)&=\sum_{j_1=1}^{k^*_1}\exp(\hbzjn)\left[\exp((\hbojn)^{\top}x)h^n_{j_1}(x)-\exp((\hboj)^{\top}x)h^*_{j_1}(x)\right]\nonumber\\
    &-\sum_{j_1=1}^{k^*_1}\exp(\hbzjn)\left[\exp((\hbojn)^{\top}x)-\exp((\hboj)^{\top}x)\right]h_{G_n}(x)\nonumber\\
    &+\sum_{j_1=1}^{k^*_1}\left(\exp(\hbzjn)-\exp(\hbzj)\right)\exp((\hboj)^{\top}x)\left[h^*_{j_1}(x)-h_{G_n}(x)\right],
\end{align*}
where we denote
\begin{align*}
    h^n_{j_1}(x)&:=\sum_{j_2=1}^{k^*_2}\sum_{i_2\in\mathcal{A}_{j_2|j_1}}\softmax((\kappa^n_{i_2|j_1})^{\top}x+\nu^n_{i_2|j_1})((a^n_{j_1i_2})^{\top}x+b^n_{j_1i_2}),\\
    h^*_{j_1}(x)&:=\sum_{j_2=1}^{k^*_2}\softmax((\kj)^{\top}x+\vj) ((a^*_{j_1j_2})^{\top}x+b^*_{j_1j_2}),
\end{align*}
for all $j_1\in[k^*_1]$. Recall that we have $\exp(\beta^n_{i_1})=\exp(\beta^*_{i_1})$ and $\omega^n_{i_1}=\omega^*_{i_1}$ for all $i_1\in[k^*_1]$. Then, the second and third terms in the decomposition of $Q_n(x)$ become zero. Thus, we can represent $Q_n(x)$ as
\begin{align}
    \label{eq:q_n_x}
    Q_n(x)=\sum_{j_1=1}^{k^*_1}\exp(\beta^*_{j_1})\exp((\hboj)^{\top}x)\left[h^n_{j_1}(x)-h^*_{j_1}(x)\right].
\end{align}
Note that we can continue to decompose the term $h^n_{j_1}(x)-h^*_{j_1}(x)$ as
\begin{align*}
    &\Big[\sum_{j_2=1}^{k^*_2}\exp((\kappa^*_{j_2|j_1})^{\top}x+\nu^*_{j_2|j_1})\Big]\cdot[h^n_{j_1}(x)-h^*_{j_1}(x)]\\
    &=\sum_{j_2=1}^{k^*_2}\sum_{i_2\in\mathcal{A}_{j_2|j_1}}\exp(\nu^n_{i_2|j_1})\Big[\exp((\kappa^n_{i_2|j_1})^{\top}x)((a^n_{i_2j_1})^{\top}x+b^n_{i_2j_1})-\exp((\kappa^*_{j_2|j_1})^{\top}x)((a^*_{j_1j_2})^{\top}x+b^*_{j_1j_2})\Big]\nonumber\\
    &-\sum_{j_2=1}^{k^*_2}\sum_{i_2\in\mathcal{A}_{j_2|j_1}}\exp(\nu^n_{i_2|j_1})\Big[\exp((\kappa^n_{i_2|j_1})^{\top}x)-\exp((\kappa^*_{j_2|j_1})^{\top}x)\Big]h^n_{j_1}(x)\nonumber\\
    &+\sum_{j_2=1}^{k^*_2}\Big(\sum_{i_2\in\mathcal{A}_{j_2|j_1}}\exp(\nu^n_{i_2|j_1})-\exp(\nu^*_{j_2|j_1})\Big)\exp((\kappa^*_{j_2|j_1})^{\top}x)[((a^*_{j_1j_2})^{\top}x+b^*_{j_1j_2})-h^n_{j_1}(x)]\\
    &:=A_{n,j_1}(x)-B_{n,j_1}(x)+E_{n,j_1}(x)
\end{align*}
From the definitions of $\kappa^n_{i_2|i_1},a^n_{i_1i_2}$ and $b^n_{i_1i_2}$, we can rewrite $A_{n,j_1}(x)$ as follows:
\begin{align*}
    A_{n,j_1}(x)&=\sum_{i_2=1}^{2}\frac{1}{2}\exp(\nu^n_{1|j_1})\exp((\kappa^*_{1|j_1})^{\top}x)(b^n_{j_1i_2}-b^*_{j_11})\\
    &=\frac{1}{2}\exp(\nu^n_{1|j_1})\exp((\kappa^*_{1|j_1})^{\top}x)[(b^n_{j_11}-b^*_{j_11})+(b^n_{j_12}-b^*_{j_11})]\\
    &=0.
\end{align*}
Next, since $\kappa^n_{i_2|j_1}=\kappa^*_{j_2|j_1}$, we can see that $B_{n,j_1}(x)=0$. Lastly, it can be justified that $E_{n,j_1}(x)=\mathcal{O}(n^{-(r+1)})$, leading to $E_{n,j_1}(x)/\mathcal{L}_{6,r}(G_n,G_*)\to0$. Combined these results, we deduce that
\begin{align*}
    \Big[\sum_{j_2=1}^{k^*_2}\exp((\kappa^*_{j_2|j_1})^{\top}x+\nu^*_{j_2|j_1})\Big]\cdot[h^n_{j_1}(x)-h^*_{j_1}(x)]/\mathcal{L}_{6,r}(G_n,G_*)\to0,
\end{align*}
as $n\to\infty$. Since the term $\sum_{j_2=1}^{k^*_2}\exp((\kappa^*_{j_2|j_1})^{\top}x+\nu^*_{j_2|j_1})$ is bounded, it follows that $[h^n_{j_1}(x)-h^*_{j_1}(x)]/\mathcal{L}_{6,r}(G_n,G_*)\to0$ for almost every $x$. Putting this limit and equation~\eqref{eq:q_n_x} together, we get that $Q_n(x)/\mathcal{L}_{6,r}(G_n,G_*)\to0$ as $n\to\infty$ for almost every $x$. Again, since the term $\sum_{j_1=1}^{k^*_1}\exp((\hboj)^{\top}x+\hbzj)$ is bounded, we deduce that $[h_{G_n}(x)-h_{G_*}(x)]/\mathcal{L}_{6,r}(G_n,G_*)\to0$ as $n\to\infty$ for almost every $x$. This result implies that
\begin{align*}
    \normf{h_{G_n}-h_{G_*}}/\mathcal{L}_{6,r}(G_n,G_*)\to0,
\end{align*}
as $n\to\infty$. Hence, the proof of claim~\eqref{eq:ratio_zero_limit_hierarchical} is completed.

\section{Auxiliary Results}

\subsection{Proof of Proposition~\ref{theorem:regression_rate_softmax}}
\label{appendix:regression_rate_softmax}
\noindent
For the proof of the theorem, we first introduce some necessary concepts and notations. Firstly, we denote by $\mathcal{F}_k(\Theta)$ the set of regression functions w.r.t all mixing measures in $\mathcal{G}_k(\Theta)$, that is, $\mathcal{F}_k(\Theta):=\{f_{G}(x):G\in\mathcal{G}_{k}(\Theta)\}$.
Additionally, for each $\delta>0$, the $L^{2}$ ball centered around the regression function $f_{G_*}(x)$ and intersected with the set $ \mathcal{F}_k(\Theta)$ is defined as
\begin{align*}   \mathcal{F}_k(\Theta,\delta):=\left\{f \in \mathcal{F}_k(\Theta): \|f -f_{G_*}\|_{L^2(\mu)} \leq\delta\right\}.
\end{align*}
In order to measure the size of the above set, \cite{vandeGeer-00} suggest using the following quantity:
\begin{align}
    \label{eq:bracket_size}
    \mathcal{J}_B(\delta, \mathcal{F}_k(\Theta,\delta)):=\int_{\delta^2/2^{13}}^{\delta}H_B^{1/2}(t, \mathcal{F}_k(\Theta,t),\|\cdot\|_{L^2(\mu)})~\dint t\vee \delta,
\end{align}
where $H_B(t, \mathcal{F}_k(\Theta,t),\|\cdot\|_{L^2(\mu)})$ stands for the bracketing entropy \cite{vandeGeer-00} of $ \mathcal{F}_k(\Theta,u)$ under the $L^{2}$-norm, and $t\vee\delta:=\max\{t,\delta\}$. By using the similar proof argument of Theorem 7.4 and Theorem 9.2 in \cite{vandeGeer-00} with notations being adapted to this work, we obtain the following lemma:
\begin{lemma}
    \label{lemma:density_rate}
    Take $\Psi(\delta)\geq \mathcal{J}_B(\delta, \mathcal{F}_k(\Theta,\delta))$ that satisfies $\Psi(\delta)/\delta^2$ is a non-increasing function of $\delta$. Then, for some universal constant $c$ and for some sequence $(\delta_n)$ such that $\sqrt{n}\delta^2_n\geq c\Psi(\delta_n)$, we achieve that
    \begin{align*}
        \mathbb{P}\Big(\|f_{\widehat{G}_n} - f_{G_*}\|_{L^2(\mu)} > \delta\Big)\leq c \exp\left(-\frac{n\delta^2}{\nu^2}\right),
    \end{align*}
    for all $\delta\geq \delta_n$.
\end{lemma}
\noindent
\textbf{Proof overview.} In this proof, we first demonstrate that the following bound holds for any $0 < \varepsilon \leq 1/2$:
\begin{align}    
H_B(\varepsilon,\mathcal{F}_k(\Theta),\|.\|_{L^{2}(\mu)}) \lesssim \log(1/\varepsilon), \label{eq:bracket_entropy_bound}
\end{align}
Then, it follows that 
\begin{align}
    \label{eq:bracketing_integral}
    \mathcal{J}_B(\delta, \mathcal{F}_k(\Theta,\delta))= \int_{\delta^2/2^{13}}^{\delta}H_B^{1/2}(t, \mathcal{F}_k(\Theta,t),\normf{\cdot})~\dint t\vee \delta\lesssim \int_{\delta^2/2^{13}}^{\delta}\log(1/t)dt\vee\delta.
\end{align}
Let $\Psi(\delta)=\delta\cdot[\log(1/\delta)]^{1/2}$, then $\Psi(\delta)/\delta^2$ is a non-increasing function of $\delta$. Furthermore, equation~\eqref{eq:bracketing_integral} indicates that $\Psi(\delta)\geq \mathcal{J}_B(\delta,\mathcal{F}_k(\Theta,\delta))$. In addition, let $\delta_n=\sqrt{\log(n)/n}$, then we get that $\sqrt{n}\delta^2_n\geq c\Psi(\delta_n)$ for some universal constant $c$. Finally, by applying Lemma~\ref{lemma:density_rate}, we achieve the desired conclusion of Proposition~\ref{theorem:regression_rate_softmax}, that is,
\begin{align*}
        \normf{f_{\widehat{G}_n}-f_{G_*}}=\mathcal{O}_{P}([\log(n)/n]^{\frac{1}{2}}).
    \end{align*}
\textbf{Proof of the bound~\eqref{eq:bracket_entropy_bound}.} Let $\zeta\leq\varepsilon$ and $\{\pi_1,\ldots,\pi_N\}$ be the $\zeta$-cover under the $L^{\infty}$-norm of the set $\mathcal{F}_k(\Theta)$ where $N:={N}(\zeta,\mathcal{F}_k(\Theta),\|\cdot\|_{\infty})$ is the $\eta$-covering number of the metric space $(\mathcal{F}_k(\Theta),\|\cdot\|_{\infty})$. Next, since the expert function $\mathcal{E}(\cdot,\eta)$ is bounded, the regression function $f_{G}(\cdot)$ is also bounded, that is, $f_{G}(x) \leq M$ for all $x$, where $M>0$ is the bounded constant of the expert function. Then, we construct the brackets of the form $[L_i(x),U_i(x)]$ for all $i\in[N]$ as follows:
    \begin{align*}
        L_i(x)&:=\max\{\pi_i(x)-\zeta,0\},\\
        U_i(x)&:=\max\{\pi_i(x)+\zeta, M \}.
    \end{align*}
From the above construction, we can validate that $\mathcal{F}_{k}(\Theta)\subset\cup_{i=1}^{N}[L_i(x),U_i(x)]$ and $U_i(x)-L_i(x)\leq \min\{2\zeta,M\}$. Therefore, it follows that 
\begin{align*}
    \normf{U_i-L_i}=\Big(\int(U_i-L_i)^2\dint\mu(x)\Big)^{1/2}\leq\Big(\int 4\zeta^2\dint\mu(x)\Big)^{1/2}=2\zeta.
\end{align*}
By definition of the bracketing entropy, we deduce that
\begin{align}
    \label{eq:bracketing_covering}
    H_B(2\zeta,\mathcal{F}_{k}(\Theta),\normf{\cdot})\leq\log N=\log {N}(\zeta,\mathcal{F}_k(\Theta),\|\cdot\|_{\infty}).
\end{align}
Therefore, we need to provide an upper bound for the covering number $N$. In particular, we denote $\Delta:=\{(\beta,\omega)\in\mathbb{R}\times\mathbb{R}^d:(\beta,\omega,\eta)\in\Theta\}$ and $\Omega:=\{\eta\in\mathbb{R}^q:(\beta,\omega,\eta)\in\Theta\}$. Since $\Theta$ is a compact set, $\Delta$ and $\Omega$ are also compact. Therefore, we can find $\zeta$-covers $\Delta_{\zeta}$ and ${\Omega}_{\zeta}$ for $\Delta$ and $\Omega$, respectively. We can check that 
\begin{align*}
    |\Delta_{\zeta}|\leq \mathcal{O}(\zeta^{-(d+1)k}), \quad |\Omega_{\zeta}|\leq \mathcal{O}(\zeta^{-qk}).
\end{align*}
For each mixing measure $G=\sum_{i=1}^{k}\exp(\beta_{i})\delta_{(\omega_{i},\eta_{i})}\in\mathcal{G}_k(\Theta)$, we consider other two mixing measures:
\begin{align*}
    \widetilde{G}:=\sum_{i=1}^k\exp(\beta_{i})\delta_{({\beta}_{1i},\overline{\eta}_i)}, \qquad \overline{G}:=\sum_{i=1}^k\exp(\overline{\beta}_{i})\delta_{({\overline{\beta}}_{1i},\overline{\eta}_i)}.
\end{align*}
Here, $\overline{\eta}_{i}\in{\Omega}_{\zeta}$ such that $\overline{\eta}_{i}$ is the closest to $\eta_{i}$ in that set, while $(\overline{\beta}_{i},\overline{\omega}_{i})\in\Delta_{\zeta}$ is the closest to $(\beta_{i},\omega_{i})$ in that set. From the above formulations, we get that
\begin{align*}
    \|f_{G}-f_{\widetilde{G}}\|_{\infty}&=\sup_{x\in\mathcal{X}}~\Bigg|\sum_{i=1}^{k}\frac{\exp((\omega_{i})^{\top}x+\beta_{i})}{\sum_{j=1}^{k}\exp((\omega_{j})^{\top}x+\beta_{j})}\cdot[\mathcal{E}(x,\eta_{i})-\mathcal{E}(x,\overline{\eta}_{i})]\Bigg|\\
    &\leq \sum_{i=1}^{k}\sup_{x\in\mathcal{X}}~\frac{\exp((\omega_{i})^{\top}x+\beta_{i})}{\sum_{j=1}^{k}\exp((\omega_{j})^{\top}x+\beta_{j})}\cdot|\mathcal{E}(x,\eta_{i})-\mathcal{E}(x,\overline{\eta}_{i})|\\
    &\leq \sum_{i=1}^{k}\sup_{x\in\mathcal{X}}~|\mathcal{E}(x,\eta_{i})-\mathcal{E}(x,\overline{\eta}_{i})|\\
    &\lesssim \sum_{i=1}^{k}\sup_{x\in\mathcal{X}}~\|\eta_i-\overline{\eta}_i\|\\
    &\lesssim\zeta.
\end{align*}
Here, the first inequality is according to the triangle inequality, the second inequality occurs as the softmax weight is bounded by one, and the fourth inequality follows from the fact that the expert function $\mathcal{E}(x,\cdot)$ is a Lipschitz function. Next, we have
\begin{align*}
    \|f_{\widetilde{G}}-f_{\overline{G}}\|_{\infty}&= \sup_{x\in\mathcal{X}}~\Bigg|\sum_{i=1}^{k}\Bigg(\frac{\exp((\omega_{i})^{\top}x+\beta_{i})}{\sum_{j=1}^{k}\exp((\omega_{j})^{\top}x+\beta_{j})}-\frac{\exp((\overline{\omega}_{i})^{\top}x+\overline{\beta}_{i})}{\sum_{j=1}^{k}\exp((\overline{\omega}_{j})^{\top}x+\overline{\beta}_{j})}\Bigg)\cdot \mathcal{E}(x,\overline{\eta}_{i})\Bigg|\\
    &\leq \sum_{i=1}^{k}\sup_{x\in\mathcal{X}}~\Bigg|\frac{\exp((\omega_{i})^{\top}x+\beta_{i})}{\sum_{j=1}^{k}\exp((\omega_{j})^{\top}x+\beta_{j})}-\frac{\exp((\overline{\omega}_{i})^{\top}x+\overline{\beta}_{i})}{\sum_{j=1}^{k}\exp((\overline{\omega}_{j})^{\top}x+\overline{\beta}_{j})}\Bigg|\cdot |\mathcal{E}(x,\overline{\eta}_{i})|\\
    &\leq \sum_{i=1}^{k}\sup_{x\in\mathcal{X}}~\Bigg|\frac{\exp((\omega_{i})^{\top}x+\beta_{i})}{\sum_{j=1}^{k}\exp((\omega_{j})^{\top}x+\beta_{j})}-\frac{\exp((\overline{\omega}_{i})^{\top}x+\overline{\beta}_{i})}{\sum_{j=1}^{k}\exp((\overline{\omega}_{j})^{\top}x+\overline{\beta}_{j})}\Bigg|\\
    &\lesssim \sum_{i=1}^{k}\sup_{x\in\mathcal{X}}~\Big(\|\omega_{i}-\overline{\omega}_{i}\|\cdot\|x\|+|\beta_{i}-\overline{\beta}_{i}|\Big)\\
   &\lesssim \sum_{i=1}^{k}\sup_{x\in\mathcal{X}}~(\zeta\cdot B+\zeta)\\
    &\lesssim\zeta.
\end{align*}
Above, the second inequality occurs as the expert $\mathcal{E}(x,\overline{\eta}_{i})$ is bounded. The third inequality happens since the softmax is a Lipschitz function. According to the triangle inequality, we have
\begin{align*}
    \|f_{G}-f_{\overline{G}}\|_{\infty}\leq \|f_{G}-f_{\widetilde{G}}\|_{\infty}+\|f_{\widetilde{G}}-f_{\overline{G}}\|_{\infty}\lesssim\zeta.
\end{align*}
By definition of the covering number, we deduce that
\begin{align}
    \label{eq:covering_bound}
    {N}(\zeta,\mathcal{F}_k(\Theta),\|\cdot\|_{\infty})\leq |\Delta_{\zeta}|\times|\Omega_{\zeta}|\leq \mathcal{O}(n^{-(d+1)k})\times\mathcal{O}(n^{-qk})\leq\mathcal{O}(n^{-(d+1+q)k}).
\end{align}
Combine equations~\eqref{eq:bracketing_covering} and \eqref{eq:covering_bound}, we achieve that
\begin{align*}
    H_B(2\zeta,\mathcal{F}_{k}(\Theta),\normf{\cdot})\lesssim \log(1/\zeta).
\end{align*}
Let $\zeta=\varepsilon/2$, then we obtain that 
\begin{align*}
    H_B(\varepsilon,\mathcal{F}_k(\Theta),\|.\|_{L^{2}(\mu)}) \lesssim \log(1/\varepsilon).
\end{align*}
Hence, the proof is completed.

\subsection{Proof of Proposition~\ref{theorem:regression_rate_hmoe}}
\label{appendix:regression_rate_hmoe}
\noindent
In this proof, we will adapt notions from the empirical process theory in Appendix~\ref{appendix:regression_rate_softmax} to the HMoE setting and employ some arguments in that appendix. In particular, we denote $\mathcal{F}_{k^*_1k_2}(\Theta)$ as the set of HMoE-based regression functions w.r.t all mixing measures in $\mathcal{G}_{k^*_1k_2}(\Theta)$, that is, $\mathcal{F}_{k^*_1k_2}(\Theta):=\{h_{G}(x):G\in\mathcal{G}_{k^*_1k_2}(\Theta)\}$.
Next, an $L^{2}$ ball centered around the regression function $h_{G_*}(x)$ and intersected with the set $ \mathcal{F}_{k^*_1k_2}(\Theta)$ is given by
\begin{align*}   \mathcal{F}_{k^*_1k_2}(\Theta,\delta):=\left\{h \in \mathcal{F}_{k^*_1k_2}(\Theta): \|h -h_{G_*}\|_{L^2(\mu)} \leq\delta\right\}.
\end{align*}
Lastly, we denote by $H_B(t, \mathcal{F}_{k^*_1k_2}(\Theta,t),\|\cdot\|_{L^2(\mu)})$ the bracketing entropy \cite{vandeGeer-00} of $ \mathcal{F}_{k^*_1k_2}(\Theta,u)$ under the $L^{2}$-norm. 
By arguing in a similar fashion to Appendix~\ref{appendix:regression_rate_softmax}, we can show that in order to reach the desired conclusion, it suffices to demonstrate that
\begin{align}    
H_B(\varepsilon,\mathcal{F}_{k^*_1k_2}(\Theta),\|.\|_{L^{2}(\mu)}) \lesssim \log(1/\varepsilon), \label{eq:bracket_entropy_bound_hmoe}
\end{align}
for any $0 < \varepsilon \leq 1/2$. 
Let $\zeta\leq\varepsilon$ and $\{\pi_1,\ldots,\pi_N\}$ be the $\zeta$-cover under the $L^{\infty}$-norm of the set $\mathcal{F}_{k^*_1k_2}(\Theta)$ where $N:={N}(\zeta,\mathcal{F}_{k^*_1k_2}(\Theta),\|\cdot\|_{\infty})$ is the $\zeta$-covering number of the metric space $(\mathcal{F}_{k^*_1k_2}(\Theta),\|\cdot\|_{\infty})$. By employing the same arguments in Appendix~\ref{appendix:regression_rate_softmax}, we obtain that 
\begin{align}
    \label{eq:bracketing_covering_hmoe}
    H_B(2\zeta,\mathcal{F}_{k^*_1k_2}(\Theta),\normf{\cdot})\leq\log N=\log {N}(\zeta,\mathcal{F}_{k^*_1k_2}(\Theta),\|\cdot\|_{\infty}).
\end{align}
Thus, it is necessary to bound the covering number $N$. For that purpose, let us denote $\Delta:=\{(\beta,\omega)\in\mathbb{R}\times\mathbb{R}^d:(\beta,\omega,\nu,\kappa,\eta)\in\Theta\}$ and $\Omega:=\{(\nu,\kappa,\eta)\in\mathbb{R}\times\mathbb{R}^{d}\times\mathbb{R}^q:(\beta,\omega,\nu,\kappa,\eta)\in\Theta\}$. Recall that the parameter space $\Theta$ is compact, then the sets $\Delta$ and $\Omega$ are also compact. Therefore, there exist $\zeta$-covers for $\Delta$ and $\Omega$, which will be denoted as $\Delta_{\zeta}$ and ${\Omega}_{\zeta}$, respectively. Furthermore, it can be justified that
\begin{align*}
    |\Delta_{\zeta}|\leq \mathcal{O}(\zeta^{-(d+1)k^*_1}), \quad |\Omega_{\zeta}|\leq \mathcal{O}(\zeta^{-(d+q+1)k^*_1k_2}).
\end{align*}
For a mixing measure $G=\sum_{i_1=1}^{k^*_1}\exp(\beta_{i_1})\sum_{i_2=1}^{k_2}\exp({\nu}_{i_2|i_1})\delta_{(\omega_{i_1},{\kappa}_{i_2|i_1},{\eta}_{i_1i_2})}\in\mathcal{G}_{k^*_1k_2}(\Theta)$, we take into account two additional mixing measures defined as
\begin{align*}
    \widetilde{G}&:=\sum_{i_1=1}^{k^*_1}\exp(\beta_{i_1})\sum_{i_2=1}^{k_2}\exp(\bar{\nu}_{i_2|i_1})\delta_{(\omega_{i_1},\bar{\kappa}_{i_2|i_1},\bar{\eta}_{i_1i_2})},\\
    \overline{G}&:=\sum_{i_1=1}^{k^*_1}\exp(\bar{\beta}_{i_1})\sum_{i_2=1}^{k_2}\exp(\bar{\nu}_{i_2|i_1})\delta_{(\bar{\omega}_{i_1},\bar{\kappa}_{i_2|i_1},\bar{\eta}_{i_1i_2})}.
\end{align*}
Above, $(\bar{\nu}_{i_2|i_1},\bar{\kappa}_{i_2|i_1},\bar{\eta}_{i_1i_2})\in{\Omega}_{\zeta}$ such that $(\bar{\nu}_{i_2|i_1},\bar{\kappa}_{i_2|i_1},\bar{\eta}_{i_1i_2})$ is the closest to $({\nu}_{i_2|i_1},{\kappa}_{i_2|i_1},{\eta}_{i_1i_2})$ in that set, while $(\bar{\beta}_{i_1},\bar{\omega}_{i_1})\in\Delta_{\zeta}$ is the closest to $(\beta_{i_1},\omega_{i_1})$ in that set. Additionally, we also denote
\begin{align*}
    h_{i_1}(x)&:=\sum_{i_2=1}^{k_2}\softmax((\kappa_{i_2|i_1})^{\top}x+\nu_{i_2|i_1})\mathcal{E}(x,\eta_{i_1i_2}),\\
    \tilde{h}_{i_1}(x)&:=\sum_{i_2=1}^{k_2}\softmax(({\kappa}_{i_2|i_1})^{\top}x+{\nu}_{i_2|i_1})\mathcal{E}(x,\bar{\eta}_{i_1i_2}),\\
    \bar{h}_{i_1}(x)&:=\sum_{i_2=1}^{k_2}\softmax((\bar{\kappa}_{i_2|i_1})^{\top}x+\bar{\nu}_{i_2|i_1})\mathcal{E}(x,\bar{\eta}_{i_1i_2}),
\end{align*}
for all $i_1\in[k^*_1]$. Now, we start providing an upper bound for the term $\|h_{G}-h_{\widetilde{G}}\|_{\infty}$ as
\begin{align}
    \|h_{G}-h_{\widetilde{G}}\|_{\infty}&=\sum_{i_1=1}^{k^*_1}\softmax((\omega_{i_1})^{\top}x+\beta_{i_1})\cdot\|h_{i_1}-\bar{h}_{i_1}\|_{\infty}\leq\sum_{i_1=1}^{k^*_1}\|h_{i_1}-\bar{h}_{i_1}\|_{\infty}\nonumber\\
    \label{eq:hmoe_bound}
    &\leq \sum_{i_1=1}^{k^*_1}\Big(\|h_{i_1}-\tilde{h}_{i_1}\|_{\infty}+\|\tilde{h}_{i_1}-\bar{h}_{i_1}\|_{\infty}\Big).
\end{align}
The first term in the above right hand side can be bounded as 
\begin{align}
    \|h_{i_1}-\tilde{h}_{i_1}\|_{\infty}&\leq\sum_{i_2=1}^{k_2}\sup_{x\in\mathcal{X}}~\Big|\softmax((\kappa_{i_2|i_1})^{\top}x+\nu_{i_2|i_1})\cdot[\mathcal{E}(x,\eta_{i_1i_2})-\mathcal{E}(x,\bar{\eta}_{i_1i_2})]\Big|\nonumber\\
    &\leq\sum_{i_2=1}^{k_2}\sup_{x\in\mathcal{X}}~\Big|\mathcal{E}(x,\eta_{i_1i_2})-\mathcal{E}(x,\bar{\eta}_{i_1i_2})\Big|\nonumber\\
    &\lesssim\sum_{i_2=1}^{k_2}\sup_{x\in\mathcal{X}}~(\|\eta_{i_1i_2}-\bar{\eta}_{i_1i_2}\|\cdot\|x\|)\nonumber\\
    \label{eq:hmoe_bound_1}
    &\leq\sum_{i_2=1}^{k_2}\sup_{x\in\mathcal{X}}~(\zeta\cdot B)\lesssim\zeta,
\end{align}
where the second last inequality occurs as the input space is bounded, that is, $\|x\|\leq B$ for all $x$ for some constant $B>0$.\\

\noindent
Next, the second term in the right hand side of equation~\eqref{eq:hmoe_bound} is bounded as
\begin{align}
    \|\tilde{h}_{i_1}-\bar{h}_{i_1}\|_{\infty}&\leq\sum_{i_2=1}^{k_2}\sup_{x\in\mathcal{X}}~\Big|\softmax((\kappa_{i_2|i_1})^{\top}x+\nu_{i_2|i_1})-\softmax((\bar{\kappa}_{i_2|i_1})^{\top}x+\bar{\nu}_{i_2|i_1})\Big|\cdot|\mathcal{E}(x,\bar{\eta}_{i_1i_2})|\nonumber\\
    &\lesssim \sum_{i_2=1}^{k_2}\sup_{x\in\mathcal{X}}~\Big|\softmax((\kappa_{i_2|i_1})^{\top}x+\nu_{i_2|i_1})-\softmax((\bar{\kappa}_{i_2|i_1})^{\top}x+\bar{\nu}_{i_2|i_1})\Big|\nonumber\\
    &\lesssim \sum_{i_2=1}^{k_2}\sup_{x\in\mathcal{X}}~\Big(\|\kappa_{i_2|i_1}-\bar{\kappa}_{i_2|i_1}\|\cdot\|x\|+|\nu_{i_2|i_1}-\bar{\nu}_{i_2|i_1}|\Big)\nonumber\\
    \label{eq:hmoe_bound_2}
     &\lesssim \sum_{i_2=1}^{k_2}\sup_{x\in\mathcal{X}}~(\zeta\cdot B+\zeta)\lesssim\zeta.
\end{align}
From equations~\eqref{eq:hmoe_bound}, \eqref{eq:hmoe_bound_1} and \eqref{eq:hmoe_bound_2}, we deduce that $\|h_{G}-h_{\widetilde{G}}\|_{\infty}\lesssim\zeta$. Furthermore, we also have that
\begin{align*}
    \|h_{\widetilde{G}}-h_{\overline{G}}\|_{\infty}&\leq\sum_{i_1=1}^{k^*_1}\sup_{x\in\mathcal{X}}\Big|\softmax((\omega_{i_1})^{\top}x+\beta_{i_1})-\softmax((\bar{\omega}_{i_1})^{\top}x+\bar{\beta}_{i_1})\Big|\cdot|\bar{h}_{i_1}(x)|\\
    &\lesssim\sum_{i_1=1}^{k^*_1}\sup_{x\in\mathcal{X}}\Big|\softmax((\omega_{i_1})^{\top}x+\beta_{i_1})-\softmax((\bar{\omega}_{i_1})^{\top}x+\bar{\beta}_{i_1})\Big|\\
    &\lesssim \sum_{i=1}^{k}\sup_{x\in\mathcal{X}}~\Big(\|\omega_{i_1}-\overline{\omega}_{i_1}\|\cdot\|x\|+|\beta_{i_1}-\overline{\beta}_{i_1}|\Big)\nonumber\\
     &\lesssim \sum_{i=1}^{k}\sup_{x\in\mathcal{X}}~(\zeta\cdot B+\zeta)\lesssim\zeta.
\end{align*}
Then, by the triangle inequality, we have
\begin{align*}
    \|h_{G}-h_{\overline{G}}\|_{\infty}\leq \|h_{G}-h_{\widetilde{G}}\|_{\infty}+\|h_{\widetilde{G}}-h_{\overline{G}}\|_{\infty}\lesssim\zeta.
\end{align*}
By definition of the covering number, we deduce that
\begin{align}
    \label{eq:covering_bound_hmoe}
    {N}(\zeta,\mathcal{F}_{k^*_1k_2}(\Theta),\|\cdot\|_{\infty})\leq |\Delta_{\zeta}|\times|\Omega_{\zeta}|\leq \mathcal{O}(n^{-(d+1)k})\times\mathcal{O}(n^{-(d+q+1)k})\leq\mathcal{O}(n^{-(2d+2+q)k}).
\end{align}
From equations~\eqref{eq:bracketing_covering_hmoe} and \eqref{eq:covering_bound_hmoe}, we achieve that
\begin{align*}
    H_B(2\zeta,\mathcal{F}_{k^*_1k_2}(\Theta),\normf{\cdot})\lesssim \log(1/\zeta).
\end{align*}
By setting $\zeta=\varepsilon/2$, we obtain that 
\begin{align*}
    H_B(\varepsilon,\mathcal{F}_{k^*_1k_2}(\Theta),\|.\|_{L^{2}(\mu)}) \lesssim \log(1/\varepsilon).
\end{align*}
Hence, the proof is completed.
\subsection{Identifiability of the Softmax Gating MoE}
\label{appendix:identifiability}
\begin{proposition}
    \label{prop:general_identifiability}
    Suppose that the equation $f_{G}(x)=f_{G_*}(x)$ holds for almost every $x$, then we obtain $G\equiv G'$.
\end{proposition}
\begin{proof}[Proof of Proposition~\ref{prop:general_identifiability}]
    To start with, let us expand the equation $f_{G}(x)=f_{G_*}(x)$ as
    \begin{align}
        \label{eq:general_identifiable_equation}
        \sum_{i=1}^{k}\softmax\Big((\omega_{i})^{\top}x+\beta_{i}\Big)\cdot \mathcal{E}(x,\eta_{i})=\sum_{i=1}^{k_*}\softmax\Big((\boi)^{\top}x+\bzi\Big)\cdot \mathcal{E}(x,\eta^*_{i}).
    \end{align}
    Since the expert function $x\mapsto\mathcal{E}(x,\eta)$ is strongly identifiable, the set of functions in $x$ given by $\{\mathcal{E}(x,\eta'_i):i\in[k']\}$, where $\eta'_1,\eta'_2,\ldots,\eta'_{k'}$ are distinct parameters for $k'\in\mathbb{N}$, is linearly independent. Thus, if $k$ is different from $k_*$, then we can find $i\in[k]$ such that $\eta_i\neq\eta^*_j$ for any $j\in[k_*]$, implying that $\softmax((\omega_{i})^{\top}x+\beta_{i})=0$, which is a contradiction as the softmax value cannot be zero. Therefore, it must hold that $k=k_*$ and $\Big\{\softmax\Big((\omega_{i})^{\top}x+\beta_{i}\Big):i\in[k]\Big\}=\Big\{\softmax\Big((\boi)^{\top}x+\bzi\Big):i\in[k_*]\Big\}$,
    for almost every $x$. Without loss of generality, we assume that 
    \begin{align}
        \label{eq:general_soft-soft}
        \softmax\Big((\omega_{i})^{\top}x+\beta_{i}\Big)=\softmax\Big((\boi)^{\top}x+\bzi\Big),
    \end{align}
    for almost every $x$ for any $i\in[k_*]$. Due to the invariance to translations of the softmax function, equation~\eqref{eq:general_soft-soft} implies that $\omega_{i}=\boi+u_1$ and $\beta_{i}=\bzi+u_0$ for some $u_1\in\mathbb{R}^d$ and $u_0\in\mathbb{R}$. Recall from the assumption (A.3), since $\omega_{k}=\omega^*_{k}=\zerod$ and $\beta_{k}=\beta^*_{k}=0$, we get $u_1=\zerod$ and $u_0=0$, leading to $\omega_{i}=\boi$ and $\beta_{i}=\bzi$ for all $i\in[k_*]$. Given these results, we can rewrite equation~\eqref{eq:general_identifiable_equation} as
    \begin{align}
        \label{eq:general_new_identifiable_equation}
        \sum_{i=1}^{k_*}\exp(\beta_{i})\exp((\omega_{i})^{\top}x)\mathcal{E}(x,\eta_i)=\sum_{i=1}^{k_*}\exp(\bzi)\exp((\boi)^{\top}x)\mathcal{E}(x,\eta^*_i),
    \end{align}
    for almost every $x$. Subsequently, let $P_1,P_2,\ldots,P_m$ be a partition of the index set $[k_*]$, where $m\leq k$, such that 
    
    \vspace{0.5em}
    (i)
    $\exp(\beta_{i})=\exp(\beta^*_{i'})$ for any $i,i'\in P_j$ and $j\in[k_*]$; 
    
    \vspace{0.5em}
    (ii) $\exp(\beta_{i})\neq\exp(\beta_{i'})$ if $i$ and $i'$ are not in the same set $P_j$.
    
    \vspace{0.5em} 
    \noindent
    Based on the above partition, we rewrite equation~\eqref{eq:general_new_identifiable_equation} as
    \begin{align*}
        \sum_{j=1}^{m}\sum_{i\in{P}_j}\exp(\beta_{i})\exp\Big((\omega_{i})^{\top}x\Big)\mathcal{E}(x,\eta_i)=\sum_{j=1}^{m}\sum_{i\in{P}_j}\exp(\bzi)\exp\Big((\boi)^{\top}x\Big)\mathcal{E}(x,\eta^*_i),
    \end{align*}
    for almost every $x$. Since $\omega_{i}=\boi$ and $\beta_{i}=\bzi$ for al $i\in[k_*]$, we have $\{\eta_i:i\in P_j\}\equiv\{\eta^*_i:i\in P_j\}$,
    for almost every $x$ for any $j\in[m]$. 
    Therefore, it follows that
    \begin{align*}
        G=\sum_{j=1}^{m}\sum_{i\in P_j}\exp(\beta_{i})\delta_{(\omega_{i},\eta_i)}=\sum_{j=1}^{m}\sum_{i\in P_j}\exp(\beta_{i})\delta_{(\boi,\eta^*_i)}=G_*.
    \end{align*}
    Hence, we reach the conclusion of this proposition.
\end{proof}

\bibliography{references}
\bibliographystyle{abbrv}
\end{document}